\documentclass{article}

\usepackage{PRIMEarxiv}

\usepackage[utf8]{inputenc} % allow utf-8 input
\usepackage[T1]{fontenc}    % use 8-bit T1 fonts
\usepackage{hyperref}       % hyperlinks
\usepackage{url}            % simple URL typesetting
\usepackage{booktabs}       % professional-quality tables
\usepackage{amsfonts}       % blackboard math symbols
\usepackage{nicefrac}       % compact symbols for 1/2, etc.
\usepackage{microtype}      % microtypography
\usepackage{lipsum}
\usepackage{fancyhdr}       % header
\usepackage{graphicx}       % graphics
\graphicspath{{media/}}     % organize your images and other figures under media/ folder

\usepackage{algorithm}
\usepackage{algpseudocode}
\usepackage{amsmath}

\usepackage{natbib}
\setcitestyle{authoryear,open={(},close={)}}

\title{SentimentArcs: A Novel Method for Self-Supervised Sentiment Analysis of Time Series Shows SOTA Transformers Can Struggle Finding Narrative Arcs}
\author{
  Jon Chun \\
  Digital Humanities Colab \\
  Integrated Program for Humane Studies \\
  Kenyon College \\
  Gambier, OH 43022\\
  \texttt{chunj@kenyon.edu} \\
}

\begin{document}
\maketitle

\begin{abstract}

SOTA Transformer and DNN short text sentiment classifiers report over 97\% accuracy on narrow domains like IMDB movie reviews. Real-world performance is significantly lower because traditional models overfit benchmarks and generalize poorly to different or more open domain texts. This paper introduces SentimentArcs, a new self-supervised time series sentiment analysis methodology that addresses the two main limitations of traditional supervised sentiment analysis: limited labeled training datasets and poor generalization. A large ensemble of diverse models provides a synthetic ground truth for self-supervised learning. Novel metrics jointly optimize an exhaustive search across every possible corpus:model combination. The joint optimization over both the corpus and model solves the generalization problem. Simple visualizations exploit the temporal structure in narratives so domain experts can quickly spot trends, identify key features, and note anomalies over hundreds of arcs and millions of data points. To our knowledge, this is the first self-supervised method for time series sentiment analysis and the largest survey directly comparing real-world model performance on long-form narratives.

\end{abstract}

% keywords can be removed

\section{Introduction}

Sentiment Analysis (SA), also called Opinion Mining, is a Natural Language Processing task that extracts sentiment or feelings within a text \citep{Liu2012SentimentAA}. It is particularly challenging because of the heavily contextualized, implied and subjective ways emotion is expressed in the written word. In addition, sentiment classifiers rapidly drop in accuracy when moving from binary positive/negative classification (97.5\% Smart RoBERTa Large) \citep{paperswithcodeSST2} to fine-grained five classification (59.1\% RoBERTa Large at 59.1\%) \citep{paperswithcodeSST5}. The inherent difficulty of sentiment analysis is also reflected in surprisingly poor inter-annotation agreement (IAA) between human annotators \citep{Siegert2013InterraterRF}. In contrast to most short text training sets, a large drop in IAA is seen when labeling the sentiment in natural texts \citep{Schmidt2018AnEO}.

According to current NLP leaderboards, most state of the art (SOTA) sentiment analysis models are based on some form of supervised learning \citep{paperswithcodeAll} \citep{NLPProgressSA}. BERT Transformers and variants like RoBERTa have begun displacing CNN, LSTMs and other DNN variants. Simpler lightweight traditional machine learning models like Multinomial Naive Bayes \citep{Schmidt2018AnEO} or more sophisticated ensembles like XGBoost \citep{Tilly2021PredictingMI} are still popular because they offer advantages in terms of speed, deployment, updating and relatively decent performance with suitable training. Finally, the simplest lexical models offer advantages in terms of transparency and explainability but at the cost of further losses in performance \citep{Dhaoui2017SocialMS}. 

Informal rankings of model performance are based on the shaky assumption that performance on domain-specific labeled training sets (e.g. tweets and reviews) generalize to different domains (e.g. news or medical records). Accuracy drops up to 12\% even between training and production datasets within the same domain (e.g. movies and electronics reviews) \citep{ElSahar2019ToAO}. This paper shows short text SOTA Transformer and DNN models are highly unstable, surprisingly incoherent and frequently bested by simpler models on long narrative texts. This surprising finding suggests selecting models based upon narrow benchmarks without regard to domain shifts or the individual characteristics of a given corpus can lead to below-baseline performance. More broadly, this finding suggests even high performance models require both an automated search of all corpus:model combinations along with a subsequent human expert to identify high-performing outliers.

There are three popular options to pick the best model for a given corpus. The most straightforward path is to follow SOTA leaderboards, recent survey papers \citep{Guo2021AnOO} or competitions \citep{Nakov2016SemEval2016T4} and pick the highest performing models that can be replicated or pulled from a pre-trained model zoo like Huggingface \citep{huggingface}. For now, these are generally Transformers or DNNs fine-tuned on popular labeled datasets similar to the intended application domain. A few narrow domains with distinctive linguistic patterns (e.g. financial reporting) can justify the cost of creating custom annotated dataset \citep{Snorkel} closely matching the texts expected in production. Of these three options, only custom labeled datasets can reliably verify real world performance and only when new texts resemble training examples.

The paucity of large high-quality training datasets combined with performance degradation accompanying domain shift are major limitations of most sentiment analysis models today. A growing number of new unsupervised sentiment analysis models have shown promising results using a combination of methods like custom embeddings, data augmentation, retraining, and alignment of latent spaces \citep{Singh2020360DV}. These models overcome the labeled data limitations of supervised methods by mapping new text representations to representations of texts with known sentiments. Usually they provide no general metrics on cross-domain performance or specific metrics for a particular corpus.

In summary, traditional supervised models are limited by few quality labeled training datasets which are very costly to create and generalize poorly. In either case, good performance is often proportional to the narrowness of the domain, the distinctiveness of language and the degree of alignment between training and production text domains. Newer self-supervised models overcome training dataset scarcity/cost but still face the generalization problem. Neither supervised nor unsupervised models offer guarantees on real world performance on different corpora or relative to other models. The traditional 'one model fits all texts' approach ignores the evidence that all models suffer from generalization because all texts exhibit some degree of domain shift relative to training datasets.

The SentimentArcs methodology presented in this paper solves the problem of limited datasets by generating a synthetic ground truth from a large ensemble of diverse models. It solves the ubiquitous domain shift problem with an exhaustive search to optimize joint corpus:model metrics. The Model Stability Metric measures how suitable each model is for a particular corpus. The Corpus Compatibility Metric reveals how challenging a given corpus is for the entire ensemble of models. These two novel metrics allow AutoML approaches to identify and rank the performance of all corpus:model combinations.  Easily digestible visualizations exploit the highly structured nature of time series to efficiently enable human-in-the-loop experts train models, interrogate sentiment arcs and discover new features or trends.

SentimentArcs uses novel time series processing with dimensionality reduction and DTW hierarchical clustering to efficiently process large numbers of corpus:model combinations. Both the model ensemble and reference corpora can be customized and expanded. Most current sentiment analysis focuses on standalone short texts. The few existing time series sentiment analysis libraries have no way to efficiently quantify, compare and optimize performance on more challenging long text narratives.  SentimentArcs is a new time series sentiment analysis methodology consisting of an ensemble of dozens of models, a reference corpus, a novel time series processing pipeline and new metrics for long text narratives. These can include novels, plays, collected works or any temporally ordered compilation of texts (e.g. social media threads, news, and financial reporting).

% \subsection{Contributions}

The main contributions of this paper are as follows:

1) It introduces the first self-supervised methodology for diachronic sentiment analysis. This overcomes the fundamental limitations of (a) supervised sentiment analysis (manually labelling datasets), (b) unsupervised sentiment analysis (manually testing and sifting through large corpus:model space with no guiding performance metrics) and (c) both supervised and unsupervised approaches (generalizing to out-of-domain texts with performance guarantees).

2) It introduces several novel metrics that can quantify, rank and automate optimization over all 'corpus:model' combinations for time series sentiment analysis.

3) It implements an extensive and diverse ensemble of 34 sentiment analysis models analyzing 25 narrative texts. It performs an exhaustive search through all 840 unique corpus:model combinations. It optimizes both cross-domain and cross-model performance to narrow the gap between theoretical benchmarks and real-world performance.

4) It assembles a corpora of 25 diverse long-form narratives carefully curated by our literary scholar collaborator, Katherine Elkins. This corpora can serve as common reference for future researchers to benchmark performance on long narrative texts. 

5) It provides human experts several simplified views of relative model performance to quickly identify, quantify and analyze critical points in any narrative sentiment time series. These critical points include inter-model peaks/valleys, coherence metrics, and anomalies on both the local and global scales. These features provide high-level explanations, insights and discoveries largely absent from previous sentiment analysis techniques. Upcoming collaborative publications with Katherine Elkins further critique the sentiment arcs produce by SentimentArcs.

This paper is organized into the following sections. It begins with a review of prior work in sentiment analysis with subsections on a variety of model architectures. The next two sections describe both (a) the composition of the corpora reference dataset and (b) the ensemble of sentiment analysis models. Next, the SentimentArcs' self-supervised sentiment analysis technique is presented in the Methodology and Metrics sections. In combination with extensive Appendices, the Results section discusses the findings from applying the ensemble to our reference corpora. Future Work discusses immediate follow-up research questions we are pursuing at the Kenyon College Digital Humanities Colab. Finally, the Conclusion summarizes the major ideas, contributions and findings from this research.

\section{Related Work}

\subsection{History and Trends}
\begin{figure}[h]
\centering
\includegraphics[width=12cm, height=8cm]{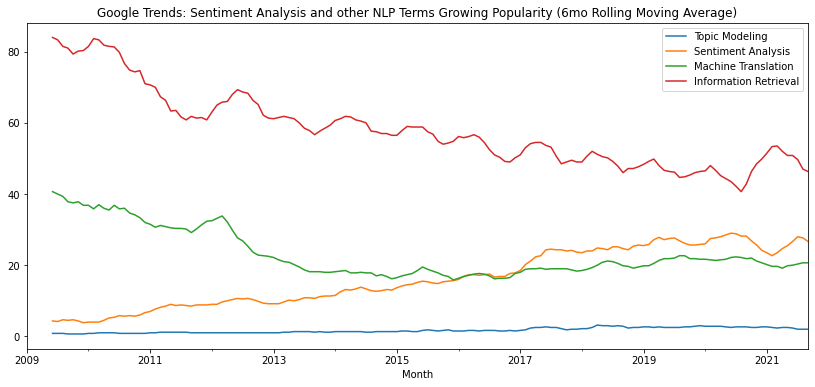}
\caption{The Rising Popularity of Sentiment Analysis}
\label{fig:sa_trend}
\end{figure}

Sentiment analysis, also known as Opinion Mining, is a popular task in the field of Natural Language Processing. The Google Trends plot in Figure \ref{fig:sa_trend} shows the term 'Sentiment Analysis' rose from relative obscurity a decade ago until today when it has suprassed 'Machine Translation'. Over this same period sentiment analysis has evolved from relatively simple models (e.g. lexicons) to encompass a wide variety of architectures described in Section \ref{sec:models} Models.

\graphicspath{ {images/} }

\subsection{Models}

\paragraph{Lexical Models:}

Lexical sentiment models use a dictionary of words bearing emotional meaning. Each word comes with a label indicating directional sentiment (e.g. positive or negative) or sometimes a degree of sentiment (e.g. 1 to 5). Several of the most popular lexical models in the SentimentArcs ensemble include the Hu-Liu Opinion Lexicon \citep{Liu2010SentimentAA} and the NRC Word-Sentiment Association Lexicon \citep{Mohammad2013CROWDSOURCINGAW}.

Lexical models have advantages of not needing training, executing quickly and being easily extended. Their primary limitations are fewer lexicons to choose from and lower performance. Since lexical models do not take context into account, they run afoul of common constructs like negation, intensifiers and more complex semantic issues like polysemy, irony, humor and sarcasm. 

SyuzhetR is an library with 300 stars on github.com that includes four lexicons: AFINN \citep{IMM2011}, Bing \citep{Hu2004MiningAS}, NRC \citep{Mohammad2013CROWDSOURCINGAW} and an extended superset of these called Syuzhet (renamed as Jockers in the SentimentR library). The Loughran-McDonald lexicon \citep{Loughran2010WhenIA} was extracted from financial texts to reduce domain shift effects when analyzing financial texts. The Jockers-Rinker lexicon is an augmented version of the Syuzhet(Jockers) lexicon found in the library SentimentR \citep{SentimentR}.

\paragraph{Heuristic Models:}

To address the most obvious shortcomings of lexical models, heuristic models like VADER and frameworks like SentimentR augment lexicons with rules to detect common contextual patterns that modify sentiment (see Table 2). Swafford \citep{Swafford} lists some common failure modes for lexical models that heuristic rules attempt to correct. Rinker surveyed a variety of corpus types (e.g. novels, debates and reviews) for these 'valence shifting' patterns and found they can affect over 20\% of lexicon words found in the corpus  \citep{SwaffordRinker}.

The R library SentimentR includes eleven sentiment lexicons as well as extensive and programmable rules. It also provides a variety of visualizations including time series plots and words highlighted by degree and direction of sentiment \citep{SentimentR}.

\begin{table}[htbp]
\label{table_heuristics}
 \caption{Lexical + Heuristic Rules}
  \centering
  \begin{tabular}{llll}
    \toprule
    % \multicolumn{}{c}{Model}                   \\
    % \cmidrule(r){4}
    Rule     & VADER     & SentimentR & Example \\
    \midrule
    Negation                & Yes  & Yes  & \textit{not} bad \\
    Contractions            & Yes  & No   & isn't bad to \textit{is not} bad \\
    Punctuation             & Yes  & Yes  & great\textit{!!!!!} vs great. \\
    Word Shape              & Yes  & Yes  & \textit{AMAZING} vs amazing \\
    Degree Modifiers        & Yes  & Yes  & \textit{very} good \\
    Adversative Conjunction & No   & Yes  & sad \textit{yet}  uplifting \\
    Slang                   & Yes  & Yes  & \textit{sux} \\
    Slang Modifiers         & Yes  & Yes  & \textit{uber} \\
    Emoticons               & Yes  & Yes  & \textit{:D}  \\
    Emojis                  & Yes  & Yes  & e.g. smiley emoji \\
    Acronyms                & Yes  & No   & \textit{LOL} \\
    \bottomrule
  \end{tabular}
  \label{tab:table_heuristics}
\end{table}
	
As a stand-alone library, C.J. Hutto's VADER may be the most popular sentiment analysis model with 3.2k stars on github.com. VADER is also embedded in the popular NLTK library (10.1k stars) \citep{Bird2004NLTKTN}. VADER's lexicon consists of over 7.5k words scraped from social media, each assigned a sentiment within the range -4 to 4 based upon the consensus of 10 trained independent human raters. It augments this core lexicon with rules listed in Table \ref{tab:table_heuristics}. VADER claims to outperform humans raters by an f1 metric of 0.96 vs 0.84 in simple classification tests \citep{Hutto2014VADERAP}. VADER does not appear in SOTA leaderboards suggesting these performance metrics do not generalize to out-of-domain datasets.

\paragraph{Classic Machine Learning Models:}

Classic machine learning models in this paper refer to traditional supervised classifiers like Logistic Regression, Support Vector Machines, Decision Trees and ensembles like XGBoost. This group of models is distinct from newer DNN and Transformer models in complexity, the latter having millions to billions of trained weights. The scikit-learn library \citep{Pedregosa2011ScikitlearnML} offers the largest collection of classical ML models in Python. More specialized models are provided by libraries like XGBoost for boosted ensembles \citep{Chen2016XGBoostAS} and sktime \citep{Lning2019sktimeAU} for working with time series data types.

Although classic ML models are rarely found in academic SOTA leaderboards, they are popular in real world applications \citep{Rahman2020ASO} for practical reasons \citep{Paleyes2020ChallengesID}. As a group, they offer a good trade-off between fast training, minimal overhead and good performance that can be enhanced with careful feature engineering and custom training. Since they are relatively lightweight and trainable, ML models are occasionally incorporated into SOTA models. Some examples include custom embeddings fed into a Naive Bayes classifier \citep{thongtan-phienthrakul-2019-sentiment} and stacked ensembles with support vector regression \citep{Akhtar2020HowIA}.

\paragraph{Deep Neural Networks:}

Deep Neural Networks like LSTMs, RNN+CNNs and architectural variants like BiLSTMs were SOTA models until the arrival of large language models based upon the scalable attention mechanism. These sequence-to-sequence DNNs serially process word tokens, accumulate an internal memory state (RNN) and selectively retain the most important information via gates (LSTM/GRU). CNN layers add the ability to selectively focus attention and efficiently form associations over longer spans of text.

One of the reasons DNN models outperform lexical, heuristic, or embedding models is because they consider context around sentiment words. DNN models often use static embeddings like Word2Vec \citep{Mikolov2013EfficientEO} or GLoVE \citep{Pennington2014GloVeGV} as input, which adds statistical semantic meaning. In addition, seq-to-seq DNN models process the context around every token. Since DNNs are large models, they can capture more complex relationships between tokens, but this requires a much larger training dataset to train these additional model parameters.

Although traditional DNNs are being displaced by ever larger Transformer models, DNN architectures are evolving to offer advantages in terms of both training and performance. Some SOTA DNN models include deep pyramid CNN classifiers \citep{johnson-zhang-2017-deep}, suffix BiLSTM \citep{Brahma2018ImprovedSM}, LSTM with dynamic skip connections \citep{Gui2019LongSM} and semi-supervised BiLSTM with mixed objective functions \citep{Sachan2019RevisitingLN}.

\paragraph{Transformers (e.g. BERT and T5):}

The 2017 seminal paper 'Attention is All You Need' by Vaswani et al. \citep{Vaswani2017AttentionIA} introduced the Transformer architecture, which has replaced many DNN architectures as SOTA models in NLP tasks. Using an encoder/decoder model, Transformers replace sequence-to-sequence layers with multi-headed self-attention heads. This reduces complexity and enables parallel training of extremely large language models. Both the very large size and selective attention mechanism of Transformers mean it can learn much longer-range dependencies across text. The first Transformers models like BERT \citep{Devlin2019BERTPO} initially achieved SOTA performance across multiple NLP tasks without even requiring customized architectures.

Although there are concerns with the role of SOTA leaderboards \citep{RogersLeaderboards}, they are a useful first approximation to chart the rise of Transformers. The SuperGLUE leaderboard \citep{SuperGLUE} ranks the performance of SOTA models on a variety of NLP tasks. This includes sentiment analysis performance on the SST-2 dataset \citep{socher-etal-2013-recursive}. As of October 2021, every identifiable model in the top 50 models incorporates some form of Transformer architecture (excluding Snorkel MeTaL \citep{Ratner2018SnorkelMW} which requires human raters). Unverified and self-reported performance sentiment analysis leaderboards also show Transformer and DNNs dominate sentiment analysis tasks \citep{PapersiwthcodeSentiment} \citep{RuderNLPProgress}.

\paragraph{Unsupervised Sentiment Analysis Models (e.g. x):}

Unsupervised learning approaches to Sentiment Analysis usually depend upon generating synthetic labels (either directly or indirectly) from existing labeled datasets or from supervised models trained on labeled datasets. Common cross-domain models statistically align distributions or latent space representations of text from trained supervised models with representations of new texts. New representations can then infer sentiment labels from the old representations to perform unsupervised learning. For example, Du and Sun used target domain masked language training and adversarial training of BERT models to distill target embeddings and identify the invariant features shared between the old and new representations \citep{Du2020AdversarialAD}.

Karouzos et al. (2021) provide an outline of the three major approaches to the general problem of Unsupervised Domain Adaption (UDA) and introduce a more robust and sample efficient approach, UDALM \citep{Karouzos2021UDALMUD}. This multitask learning approach fine-tunes a BERT model by simultaneously minimizing both task-specific loss on source data and language modeling loss on the target data. Fuzzy rules-based unsupervised learning \citep{Vashishtha2019FuzzyRB}, like statistical UDA approaches, are yet another example of addressing the limited labeled dataset problem using existing trained supervised models to inform and enable unsupervised learning.  
\paragraph{Ensemble and AutoML Sentiment Analysis:}

Ensembles combine the predictions from a variety of different models under the idea that a collection of weaker models will perform and generalize better than any single strong model. AutoML tests various model and hyperparameter configurations to find the best performing combination. Ensembles require the additional step of arriving at a consensus among the models, which may involve majority voting, Bagging, or Boosting \citep{Bauer2004AnEC}.

Generic ensemble approaches using DNNs are an active area of research \citep{Cliche2017BB} \citep{Ganaie2021EnsembleDL}. Some examples include stacking classic ML models with DNN models \citep{rozental-fleischer-2017-amobee}, training ensembles with boosting \citep{Rane2018SentimentCS}, using additive noise \citep{Blachnik2019EnsemblesOI} and applying genetic algorithms (Lopez et al. 2019). SentiXGBoost uses six different engineered features and a stacked ensemble of six weaker base classifiers with a stronger XGBoost classifier on top to outperform traditional models on six of the most popular benchmarks \citep{HamaAziz2021SentiXGboostES}.

Some ensembles have been designed specifically for sentiment analysis. Goncalves et al. used an ensemble of eight lexical models on social survey texts that ranged widely in coverage (4-95\%) and agreement (33-80\%) \citep{Gonalves2013ComparingAC} 
Singh used enhanced feature extraction (POS and bigrams) to train an ensemble of 9 classic supervised ML classifiers from sklearn on the IMDB dataset \citep{Singh2018SentimentAU}. This feature ensemble approach was improved in 2019 by defining an expanded fuzzy feature set (lexical, POS, semantic, position and polarity), word2vec embeddings and a CNN multiclassifier. \citep{Phan2020ImprovingTP}. SenticNet 6 creatively incorporates both a bottom-up traditional BiLSTM/BERT model and a top-down symbolic logical reasoning system in an ambitious attempt to fuse these very different paradigms \citep{Cambria2020SenticNet6E}.

\subsection{Narrative Sentiment Arcs in the Digital Humanities}

Shortly after WWII, Kurt Vonnegut proposed studying the emotional shape of stories for his graduate thesis in anthropology at the University of Chicago \citep{LaphamsQuarterlyVonnegut} \citep{VonnegutYoutTube}. He posited that popular stories contain a few basic emotional shapes. In 2016, Reagan et al. extracted six basic story shapes from 1,327 stories to support this theory \citep{Reagan2016TheEA}.

In 2011, J. Max Wilson created diachronic sentiment analysis visualizations of the Book of Mormon \citep{MaxWilson} based upon commercial sentiment analysis software \citep{BookOfMormon}. The Helen Wills Neuroscience Institute summarized a variety of narrative sentiment arc visualizations \citep{Bilenko2013VisualizationON}.  Researchers at Educational Testing Services interested in automatically grading student essays \citep{Somasundaran2018TowardsEN} investigated how sentiment arcs and features relate to essay quality \citep{Somasundaran2020EmotionAO}.

In June 2014, Matthew Jockers proposed using diachronic sentiment analysis to detect plot in novels \citep{JockersSyuzhet}. In dialog with other Digital Humanities scholars like Anne Swafford \citep{Swafford}, Ted Underwood \citep{UnderwoodSyuzhet}, Andrew Piper \citep{pipersa}, and Ben Schmidt \citep{SchmidtSyuzhet}, Jockers refined his library SyuzhetR \citep{SyuzhetR} to generate smooth diachronic sentiment arcs from long-form texts. As a literary scholar, Jockers confirmed that the peaks/valleys of extracted sentiment arc often align with the pivotal scenes in novels like James Joyce's \textit{Portrait of the Artist as a Young Man} \citep{JockersBlog}.

\section{Corpora}

SentimentArcs' corpora consists of 25 narratives selected to create a diverse set of well recognized novels that can serve as a benchmark for future studies. The composition of the corpora was limited by the effect of copyright laws as well as historical imbalances. Most works were obtained from US and Australian Gutenberg Projects \citep{GutenbergUS} \citep{GutenbergAU}. The corpora is expected to grow in size and diversity over time. We welcome collaboration with domain experts who can provide analysis and annotations on long text narratives (e.g. individual novels, collected works, plays, long social media threads, news and social media compilations, etc).

\begin{table}[htbp]
 \caption{Corpus Novels}
  \centering
  \begin{tabular}{llll}
    \toprule
    % \multicolumn{}{c}{Model}                   \\
    % \cmidrule(r){4}
    Author           & Title & Date     & Lines \\
    \midrule
    Charles Dickens             & A Christmas Carol                             & 1843  & 1,399 \\
    Charles Dickens             & Great Expectations                            & 1861  & 7.230 \\
    Daniel Defoe                & Robinson Crusoe                               & 1719  & 2,280 \\
    E.M. Forester               & Howards End                                  & 1910  & 8,999 \\
    Frank Baum                  & The Wonderful Wizard of Oz                    & 1850  & 2,238 \\
    Frederick Douglass          & Narrative of the life of Frederick Douglass, an American Slave  & 1845  & 1,688 \\
    F. Scott Fitzgerald          & The Great Gatsby                              & 1925  & 2,950 \\
    George Eliot                & Middlemarch                                   & 1871  & 10,373 \\
    Henry James                 & The Portrait of a Lady                        & 1881  & 13,258 \\
    Homer (trans Emily Wilson)  & The Odyssey                                   & c700BCE/2018  & 6,814 \\
    Ian McEwan                  & Machines Like Me                              & 2019  & 6,448 \\
    Jane Austen                 & Pride and Prejudice                           & 1813  & 5,891 \\
    Joseph Conrad               & Heart of Darkness                             & 1902  & 1,619 \\
    James Joyce                 & A Portrait of the Artist as a Young Man       & 2016  & 5,584 \\
    J.K. Rowling                & Harry Potter and the Sorcerer's Stone         & 1997  & 5,488 \\
    Marcel Proust               & In Search of Lost Time, Vol 3: The Guermantes Way   & 1920  & 8,388 \\
    Mary Shelly                 & Frankenstein                                  & 1818  & 3,282 \\
    Mark Twain                  & Huckleberry Finn                              & 1884  & 5,775 \\
    St. Augustine               & Confessions (Books 1-9)                       & c400  & 3,673 \\
    Toni Morrison               & Beloved                                       & 1987  & 7,102 \\
    Viktor Nabokov              & Pale Fire                                      & 1962  & 2,984 \\
    Virginia Woolf              & Mrs. Dalloway                                 & 1925  & 3,647 \\
    Virginia Woolf              & Orlando                                       & 1928  & 2,992 \\
    Virginia Woolf              & The Waves                                     & 1931  & 3,919 \\
    Virginia Woolf              & To The Lighthouse                             & 1927  & 3,403 \\
    \bottomrule
  \end{tabular}
  \label{tab:table_corpora}
\end{table}

Each novel in the corpora was parsed into lines as described in the Text Preprocessing section below. The minimum corpus length is 1,399 lines, the maximum length is 13,258 lines, and the mean length is 4,856 lines. There is significant variation in lengths with a standard deviation of 2,899 lines and a 25-75\% quartile range between 2,984 to 6,448. There is a right distributional skew of novel lengths as seen in Figure \ref{fig:hist_corpora_len} below.

\begin{figure}[h!]
\centering
\includegraphics[width=10cm, height=8cm]{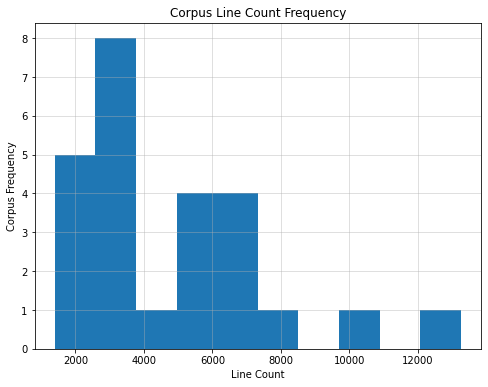}
\caption{Histogram of Corpora Lengths}
\label{fig:hist_corpora_len}
\end{figure}

Several dimensions of diversity were considered for inclusion including popularity, period, genre, topic, style and author diversity. The first version of our corpus includes only English, although Proust and Homer are included in translation. SentimentArcs has processed a larger set of novels, including some in foreign languages. The initial reference corpus is in English since performance across all ensemble models was uneven in less resourced languages \citep{Dashtipour2016MultilingualSA}. 

SentimentArcs' corpora spans approximately 2300 years from Homer's \textit{Odyssey} to the 2019 \textit{Machines like Me} by award-winning author, Ian McEwan. Early 20th century modernists are emphasized with authors like Marcel Proust and Virginia Woolf because this is the speciality of our collaborating literary scholar, Katherine Elkins. In upcoming publications, Elkins provides the ground truth domain expertise with a close and intermediate reading of the 840 sentiment arcs produced by the ensemble's 34 models processing each of the 25 works.

In sum, the corpora includes (1) the two most popular novels on Gutenberg.org \citep{GutenbergTop},  (2) eight of the fifteen most assigned novels at top US universities \citep{EABTopUnivBooks}, and (3) three works that have sold over 20 million copies \citep{WikiBestsellers}. There are eight works by women, two by African-Americans and five works by two LGBTQ authors. Britain leads with 15 authors followed by 6 Americans and one each from France, Russia, North Africa and Ancient Greece.

One of the most important criteria for inclusion in the corpora is familiarity. Unlike traditional supervised sentiment analysis, which provides performance metrics based on pre-labeled reference datasets, SentimentArcs is a self-supervised technique without labels. Instead, it relies on the combination of a synthetic ground-truth, new metrics and (optional) human-in-the-loop domain experts to guide model selection, training and assessment. Well known novels allow human raters to apply domain expertise to more accurately assess model output and gain confidence in generated narrative arcs without pre-labeled datasets.
	
Some traditional literary scholars have expressed skepticism over Digital Humanities (DH) computational approaches like sentiment analysis. A recent paper set out to prove that all major computational DH projects have only produced results that are either false or trivially self-evident \citep{Da}. Another article that resonated with some traditional scholars argued that Digital Humanities can't contribute anything significant to the scholarship because literature is categorically distinct from data \citep{MarcheLAReview}. This inherent cultural skepticism makes diachronic sentiment analysis of literary texts more challenging due to higher expectations for transparency and explainability.

\section{Models}
\label{sec:models}

The self-supervised nature of SentimentArcs relies upon generating a synthetic ground truth for each corpus in the form of a consensus ensemble sentiment arc. This is the median of all normalized sentiment arcs produced by every model in the ensemble. This ensemble median is unique to each corpus and is the fundamental reference for calculating performance metrics. The ensemble currently consists of the following models grouped into the following families (8) Transformer Models, (5) DNNs, (8) Statistical ML Models, (9) Lexical-Heuristic Models and (4) Lexical Models. 

Lightweight models were grouped into the same Jupyter notebook while long-running, more resource intensive models were broken out into separate notebooks (e.g. DNN and Transformers). All notebooks were written in Python 3.7 and executed on Google Colab Pro over the summer and fall of 2021. During this period Google introduced Colab Pro+ and service quality on Colab Pro dropped with unattended long-running jobs blocked. As a result, execution times varied tremendously and no reliable training statistics could be collected on this cloud platform. These will be gathered for future runs on a more stable platform.

An effort was made to keep the code base entirely in Python, but since SyuzhetR and SentimentR are the two top diachronic sentiment analysis libraries, they warranted an exception. This required feeding preprocessed texts to R scripts and importing the results back into the Python Jupyter notebook workflow.

\begin{table}[htbp]
 \caption{Ensemble Models}
  \centering
  \begin{tabular}{lll}
    \toprule
    % \multicolumn{}{c}{Model}                   \\
    % \cmidrule(r){4}
    Name     & label             & Notes \\
    \midrule
    \multicolumn{3}{|c|}{Lexical Models} \\
    \midrule
    SyuzhetR/AFINN         & afinn           & 2,477 tokens \\
    SyuzhetR/Bing          & bing            & 5,469 tokens \\
    SyuzhetR/NRC           & bing            & 5,468 tokens \\
    SyuzhetRLexicon        & SyuzhetR        & 10,748 tokens \\
    SentimentR Lexicon     & SentimentR      & 11,710 tokens \\
    AJ POS Lexicon         & pattern         & 2,918 tokens  \\
    \midrule
    \multicolumn{3}{|c|}{Lexical + Heuristic Models} \\
    \midrule
    SenttimentR/Bing             & huliu          & 5,469 tokens \\
    SenttimentR/NRC              & huliu          & 5,469 tokens \\
    SenttimentR/SentiWord        & sentiword      & 20,093 tokens \\
    SenttimentR/SenticNet        & senticnet      & 23,626 tokens \\
    SentimentR/Loughran-McDonald & lmcd           & 4,150 tokens \\
    SentimentR/Jockers           & jockers        & 10,748 tokens \\
    SentimentR/Jockers-Rinker    & jockers-rinker & 11,710 tokens \\
    VADER                        & vader          & 7,520 tokens  \\
    \midrule
    \multicolumn{3}{|c|}{Traditional Machine Learning Models} \\
    \midrule
    Logistic Regression         & logreg        & scikit-learn \\
    Logistic Regression CV6     & logreg{\_}cv  & scikit-learn \\
    Multinomial Naive Bayes     & multinb       & scikit-learn \\
    Multinomial Naive Bayes/POS & textblob      & scikit-learn \\
    Random Forest               & rf            & scikit-learn \\
    XGBoost                     & xgb           & xgboost \\
    FLAML AutoML                & flaml         & MS FLAML \\
    AutoGluon Text AutoML       & autogluon     & AWS AutoGluon Text \\
    \midrule
    \multicolumn{3}{|c|}{Deep Neural Networks} \\
    \midrule
    Fully Connect Network    & fcn     & 6,287,671 params \\
    Long Short Term Memory   & lstm    & 7,109,089 params \\
    Convolutional Network    & cnn     & 1,315,937 params \\
    CNN AutoML               & Stanza  & Multilingual CNN \\
    CNN AutoML               & Flair   & PyTorch HyperOpt \\
    \midrule
    \multicolumn{3}{|c|}{Transformers} \\
    \midrule
    Distilled BERT            & huggingface      & Huggingface Default \\
    T5 IMDB                   & t5imdb50k        & T5 \\
    BERT Dual Coding          & hinglish         & BERT \\
    BERT Yelp                 & yelp             & BERT \\
    BERT 2way IMDB            & imdb2way         & BERT \\
    BERT Multilingual         & NLPTown          & BERT \\
    XML RoBERTa 8 Languages   & robertaxml8lang  & RoBERTa \\
    Large RoBERTa 15 Datasets & roberta15lg      & RoBERTa \\
    \bottomrule
  \end{tabular}
  \label{tab:table_models}
\end{table}

\paragraph{Lexicons}

Lexicon models are based upon a variety of sentiment dictionaries from standalone libraries (e.g. AFINN, Pattern), the SyuzhetR library or from the SentimentR framework. CLIP's Pattern NLP Framework is a type of multilingual lexical model augmented with synsets focusing on adjective POS. The SyuzhetR(Jockers) lexicon is an augmented superset of previous lexicons like Bing and NRC. The Jockers-Rinker lexicon is an extension of the SyuzhetR lexicon. The overlap between some lexicons provides more weight to the simple, transparent and more familiar lexical models and strengthens the ensemble baseline coherence.

\paragraph{Heuristics:}

Heuristic models include VADER and seven lexical models from the SentimentR framework. Negation and amplification are the most common patterns that modify lexicon tokens (e.g. not happy and very mad) according to a survey of various types of texts \citep{WhySentimentR}. Both VADER and SentimentR have rulesets that catch these and other common syntax patterns that modify the sentiment of lexicon words. Several of these heuristic models overlap with lexical models, and the performance difference lends insight into the practical value of these rules on real-word texts.

\paragraph{Classic ML:}

Some surveys \citep{CorpMLSurvey} have shown classic ML models are more common in industry than academia, where DNN, Transformers and other larger SOTA models are the focus of most research. Therefore, several models were picked from this family based upon on popularity (e.g. Multinomial Naive Bayes and Logistic Regression), explainability (e.g. Decision Trees) and performance (e.g. XGBoost). Many of these models have the fast inference performance of simple lexicon models but with the ability to fine-tune for higher performance. A diversity of optimization strategies were represented including: default settings, boosted ensembles (e.g. XBBoost) and AutoML grid search (AutoGluon and FLAML). AutoGluon was a favorite given previous research showing it outperformed other AutoML libraries \citep{Blohm2021LeveragingAM}.

\paragraph{DNN:}

The ensemble includes very basic FCN, LSTM and CNN sentiment classifier models along with two AutoML CNNs. These models only provide a baseline as no extensive hyperparameter search, network architectural search or advanced architectures were included in the DNN family of models.

Two AutoML packages based on CNNs were included. Stanford's Stanza package \citep{Qi2020StanzaAP} uses a multilingual neural network pretrained as a sentiment classifier. Flair \citep{Akbik2019FLAIRAE} offers many embeddings and uses HyperOpt to optimize an IMDB pre-trained sentiment classifier. Each model was trained on the IMDB labeled dataset.

\paragraph{Transformers:}

One motivating hypothesis of this research was to confirm our expectation that SOTA Transformer models would dominate this new unsupervised narrative sentiment task much like they do traditional supervised simple sentiment analysis SOTA benchmarks. To confirm this, eight large language Transformer models from three basic Transformer architecture variants (BERT, RoBERTa and T5) were included from Huggingface's model zoo \citep{HuggingfaceZoo}. These models used a variety of training and fine-tuning strategies that include a distilled model (DistillBERT \citep{HuggingfaceDistillBERT}), a bilingual code switching model (Hingish \citep{HuggingfaceHinglish}), several multilingual BERT models (NLPTown \citep{HuggingfaceMultilingualNLPTown}, RoBERTa8lang) and models fine-tuned on specific sentiment datasets (yelp \citep{HuggingfaceYelp}, roberta15lg \citep{HuggingfaceRoberta15data}, t5i mdb50 \citep{HuggingfaceT5IMDB}, imdb2way \citep{HuggingfaceIMDBbinary}). The roberta15lg model was favored to outperform because it was the highest performing SOTA Transformer model and specifically fine-tuned on over 15 labeled sentiment datasets.

\section{SentimentArcs Methodology}
\label{sec:methodology}

The methodology outlined here is based upon years of working with a variety of corpora in the Digital Humanities Colab at Kenyon College \citep{KenyonDHColab}. This includes close analysis of novels, screenplays, plays, lyrics, poems, financial documents, social media, news and legal opinions. SentimentArcs was created out of the frustration that existing time series sentiment analysis approaches produce unpredictable results that at times agreed with expert human judgement while at other times were wildly inaccurate. It was difficult to discern reliable patterns between a given corpus:model combination and how coherent sentiment arcs were. Traditional techniques offered no metrics to guide corpus:model selection, and analyzing each combination individually was a grossly inefficient use of the domain expert's time.

SentimentArcs minimizes the demands on a human-in-the-loop evaluator by optimizing joint corpus:model performance to rank and guide selection. SentimentArcs depends upon the generally coherent nature of long-form narrative as well as the temporal interdependence between individual sentiment measurements. This latent structure is used to generate high-level abstract time series plots representing 10,000s of datapoints in the form of narrative arcs easily digestible by human experts. 

Traditional sentiment analysis performance relies upon a few summary metrics (accuracy, f1) or abstract AUC curves to summarize how well thousands of individual sentence sentiment values agree with a labeled short text dataset. SentimentArcs compares time series sentiment plots of both individual models and the ensemble median, which serves as a synthetic ground truth. This process can be automated or a human expert can intervene to quickly verify the results. A human-in-the-loop expert can also efficiently see key features (e.g. start, end, peaks and valleys), understand the overall shape, detect relative sentiment fluctuations and identify local/global anomalies for each corpus:model combination. Instead of assuming performance summary statistics based on narrow benchmarks generalize well, SentimentArc's synthetic ensemble reference baseline, novel metrics and visualizations provide a systematic way to efficiently find the optimal corpus:model combination that can be quantified, ranked and automated.
	
\begin{figure}[h!]
\centering
\includegraphics[width=\textwidth]{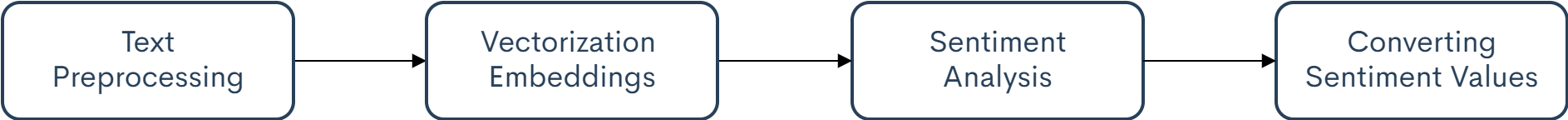}
\caption{Pipeline A: Sentiment Analysis}
\label{fig:pipeline_sentiment}
\end{figure}

\begin{figure}[h!]
\centering
\includegraphics[width=\textwidth]{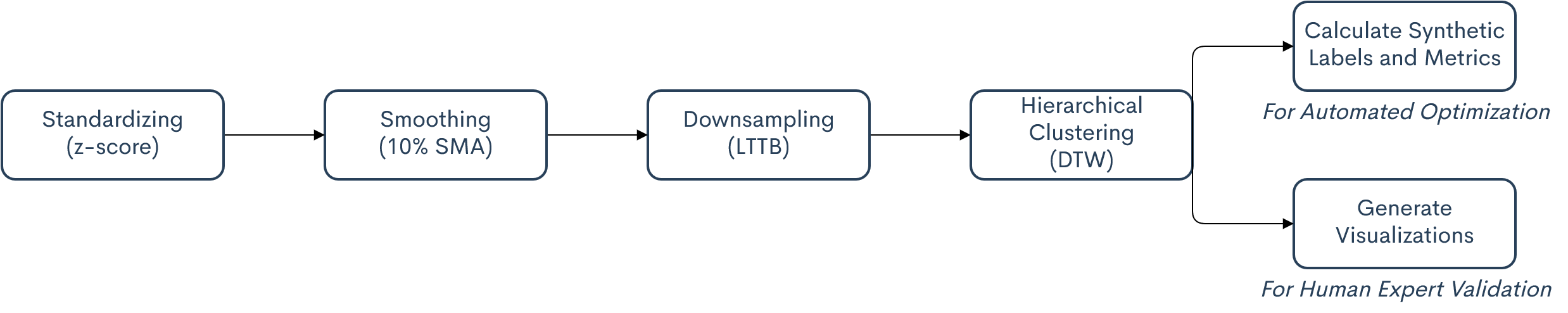}
\caption{Pipeline B: Time Series Clustering}
\label{fig:pipeline_timeseries}
\end{figure}

SentimentArc's processing pipeline is composed of two parts. The first half of the overall sentiment analysis pipeline is seen in Figure \ref{fig:pipeline_sentiment}. It involves the sequence of (a) text preprocessing, (b) vectorization and embeddings (where indicated), (c) calculating sentiment analysis on each line in the corpus, and (d) calibrating sentiment probabilities where possible. The second half of the pipeline clusters all sentiment time series as seen in Figure \ref{fig:pipeline_timeseries}. This involves (a) z-score standardization (b) smoothing with a simple moving average window, (c) downsampling the series, (d) calculating time series distances, (e) performing agglomerative hierarchical clustering, (f) calculating novel metrics for automation, and (g) generating visualizations for human experts.

\subsection{Text Prepreprocessing}

The most difficult preprocessing involved novels that were still under copyright and usually required extensive work to OCR, detect and correct OCR-related errors. Each novel had its own encoding, layout, organizational structure and other peculiarities that needed to be translated into a standardized form: plain text stripped of all headers, footers, and chapter/section headers with plain text grouped into paragraphs separated by two newlines.
	
The next stage of preprocessing depended upon the representation requirements of downstream models. All models except the large language Transformer models shared the following common text processing steps to eliminate low-sentiment bearing tokens and reduce dimensionality. 

\subparagraph{Common Text Processing Steps}
\begin{itemize}
    \item Filter out non-printable characters
    \item Convert to lowercase
    \item Expand contractions
    \item Rejoin end of line hyphenations
    \item Tokenize
    \item Filter punctuation
    \item Remove numbers
    \item Stem (for non-lexical)
\end{itemize}

\subsection{Vectorization and Embeddings}

Text destined for classic ML and DNN models (except AutoGluon Text) was vectorized using TF-IDF 1-3 ngrams and was represented by the top 5000 features. This vectorization was chosen over CountVectorizer(), WordLevel and CharLevel TfidfVectorizer() and HashingVectorizer() methods in the sklearn.feature{\_}extraction.text library \citep{scikitlearnVectorization}. TF-IDF 1-3 ngrams was chosen based upon a slight performance edge seen in randomly sampling both models and corpora with different vectorizations.

The two AutoML approaches for classical ML models offer a variety of pretrained embeddings, and defaults were used. Flair defaults to FastText embeddings trained over Wikipedia \citep{FlairEmbeddings} and Stanza uses word2vec embeddings \citep{StanzaEmbeddings}. Unlike the other models, the low-code AutoGluon Text model automatically incorporates embedding into its sentiment analysis pipeline object TextPredictor without identifying the embedding scheme or hyperparameters it chooses \citep{AutoGluonTextEmbeddings}.

Huggingface Transformer models require tokenization, encoding and padding of raw text into tensors to match each model's expected input. This step was integrated into the classifier method of some models while other models required explicitly calling Transformer.AutoTokenizer. \citep{GoogleSentencePiece}.

\subsection{Sentiment Analysis}

Classical ML and DNN sentiment classifier models both required training before analyzing our corpora. IMDb \citep{maas-etal-2011-learning} and the Stanford Sentiment Treebank (SST) \citep{socher-etal-2013-recursive} are both derived from movie reviews and are two of the largest, most popular labeled sentiment datasets. Google Colab Pro was unable to process the 215k SST dataset for DNNs, so the 50k balanced (negative/positive) IMDb dataset was selected as the training dataset. The list below shows model hyperparameter configuration (in parentheses) and the accuracy achieved after training with a 90/10 split of the training dataset.

The Microsoft AutoML library FLAML was given a time budget of 60 minutes to optimize three high performance gradient boosting tree algorithms (LightGBM, CatBoost, and XGBoost), 2 decision tree models (Decision Trees and ExtraTrees) and a simple Lasso Logistic Regression classifier. Logs show FLAML training reached a plateau well within this time budget. Over several runs, ExtraTree produced the best performance with an accuracy of 0.87 and these hyperparameter settings: 4 estimators, 0.71 max features, 19 max leaves and using the 'gini' criterion.

Amazon Lab's AutoML library AutoGluon Text was given a similar time budge to optimize classification accuracy on the IMDB dataset. Unlike FLAML, AutoGluon only provides the best performance metric without identifying the models or hyperparameter settings. The AutoGluon Text model achieved a reported accuracy of 0.89. 

\begin{itemize}
    \item \textbf{Multinomial Naive Bayes:} (defaults) acc = 0.85
    \item \textbf{Logistic Regression:} (solver="lbfgs",multi{\_}class="auto",max{\_}iter=4000) acc = 0.89
    \item \textbf{Logistic Regression with Cross Validation:} (cv=6, max{\_}iter=500) acc = 0.89
    \item \textbf{Random Forest:} (defaults) acc = 0.84
    \item \textbf{XGBoost(xgb):} (defaults) acc = 0.82
    \item \textbf{FCN:} (emb/flat/dense/dense, 6.3M params) acc = 0.87
    \item \textbf{RNN:} (emb/conv/BiLSTM, 7.1M params) acc = 0.90
    \item \textbf{CNN:} (emb/conv/conv/dense, 1.3M params) acc = 0.87
    \item \textbf{FLAML:} (ExtraTree: n{\_}estimators=4, max{\_}features=0.71, max{\_}leaves=19, criterion='gini') acc = 0.87
    \item \textbf{AutoGluon Text:} (defaults) acc = 0.89
\end{itemize}

\subsection{Converting Sentiment Values}

Where possible, the sentiment scores output by models were converted to calibrated probabilities. This was not possible for every model and resulted in some model sentiment arcs occasionally appearing as discontinuous step functions. Ultimately, most of these issues were resolved downstream using simple moving average smoothing.

\subsection{Time Series Processing}

\subsection{Standardization and Smoothing}
\label{sec:stdsmooth}

\paragraph{Standardization}

Sentiment time series were subsequently standardized using the z-score method to have a mean of 0 and standard deviation of 1 to afford more direct comparisons. This was done using the preprocessing.StandardScaler() method from the python library scikit-learn.

\begin{center}
\textbf{Standardized Normal:} $Z \sim N(0, 1)$ where $Z = \frac{X-\mu}{\sigma}$\\
\end{center}

\paragraph{Smoothing}
After applying the z-score transformation, each time series was smoothed with a 10\% simple moving average (SMA) using the Pandas DataFrame.rolling() method. The min{\_}periods=1 argument allowed the sliding window to dynamically adjust window width down to 1 period to eliminate 5\% clipping at both ends.

A standard window size for SMA was chosen to facilitate comparison with previous and future works. An informal search of literature of diachronic sentiment analysis plots suggested 5 or 10 percent were the most common values. Ten percent was chosen as the standard window size to produce smoother plots and make key features more prominent.

\begin{figure}[!ht]
\centering
\includegraphics[width=.9\linewidth]{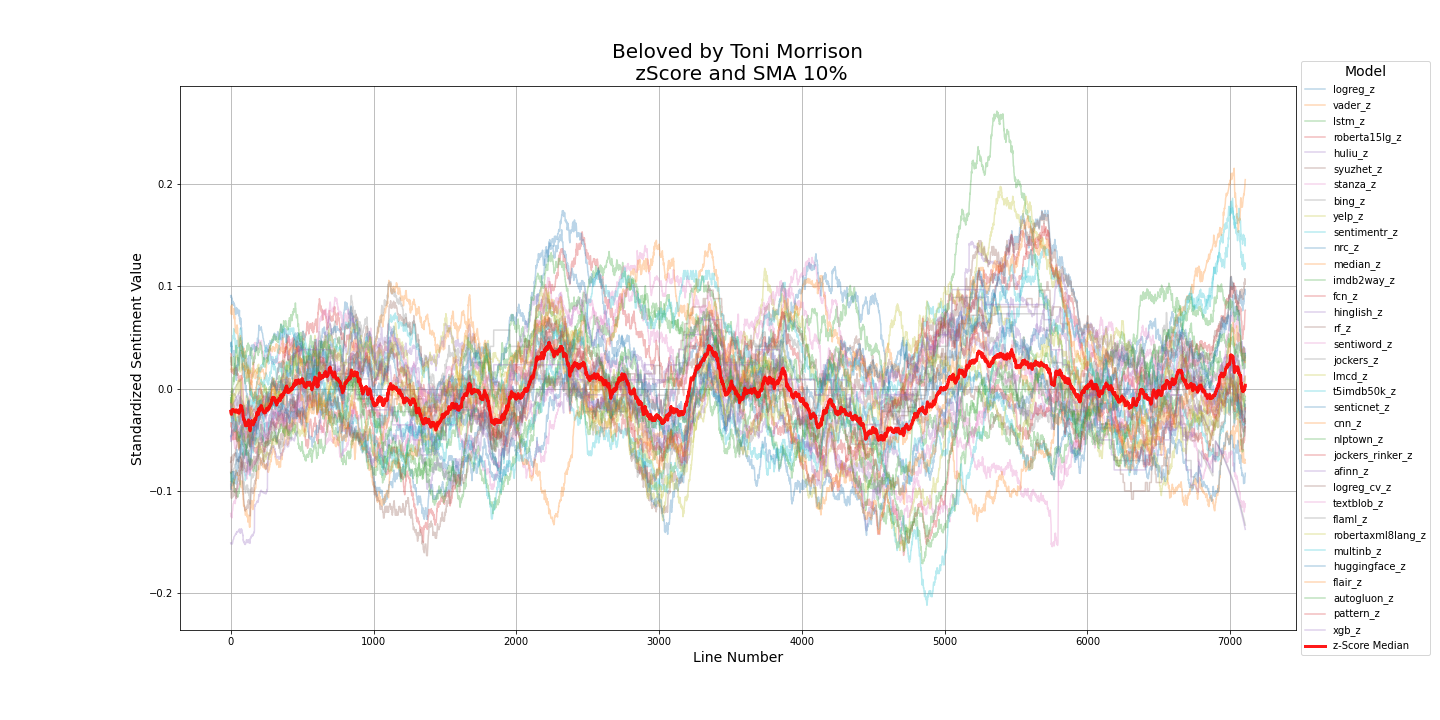}
\caption{Less Cohesive Ensemble Sentiment Plots}
\label{fig:fig_sma_beloved}
\end{figure}

\begin{figure}[!ht]
\centering
\includegraphics[width=.9\linewidth]{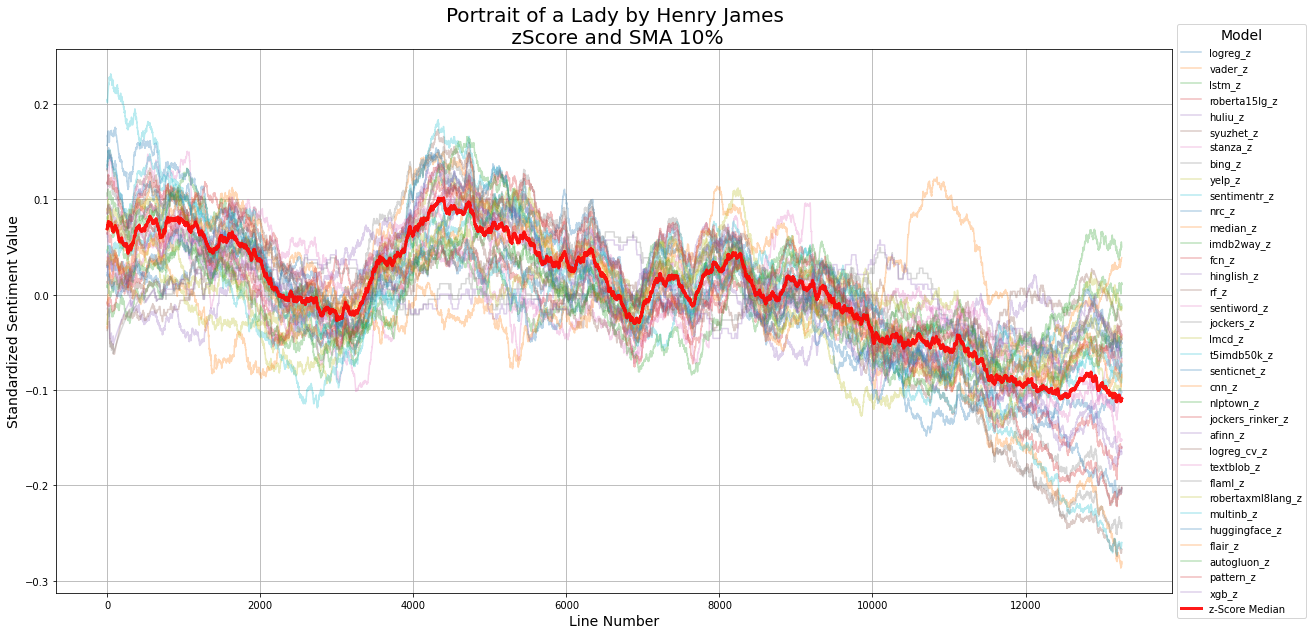}
\caption{More Cohesive Ensemble Sentiment Plots Except at Ends}
\label{fig:fig_sma_portraitlady}
\end{figure}

\paragraph{Downsampling}

Downsampling each time series was done to maximize the processing speed while minimizing the loss of accuracy of the next stage: DTW time series clustering. Dynamic Time Warping has a time complexity of O($n^2$) so dimensionality reduction is critical in this stage. The Largest Triangle Three Buckets (LTTB) algorithm \cite{Steinarsson2013DownsamplingTS} was chosen since, together with Dynamic Time Warping, it fostered meeting the usually contradictory goals of downsampling while preserving key features essential for accurate clustering.

\begin{figure}[h!]
\centering
\includegraphics[width=10cm]{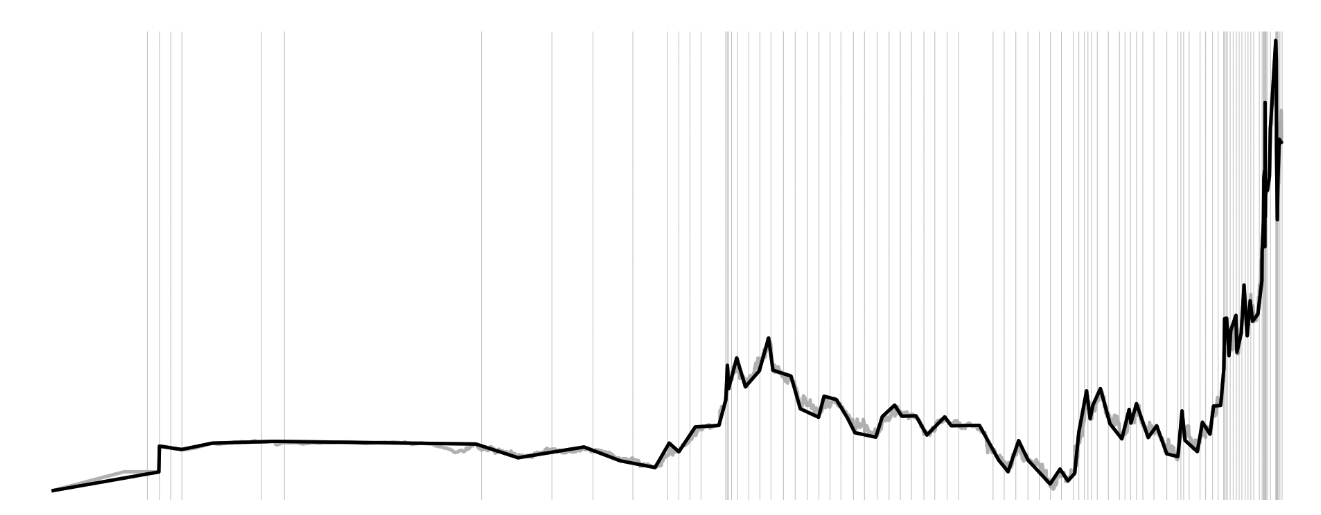}
\caption{Largest-Triangle-Three-Buckets Irregular Approximation \citep{Steinarsson2013DownsamplingTS} }
\label{fig:lttb_approx}
\end{figure}

The Largest Triangle Three Buckets (LTTB) algorithm is an efficient algorithm for dimensionality reduction of time series that normalizes all time series to the same length n while preserving key features like peaks/valleys and starting/ending points. Pseudocode for the LTTB is listed in Algorithm \ref{alg:lttb} below since it is the lesser-known algorithms used in the pipeline.

\begin{algorithm}
\caption{Largest-Triangle-Three-Buckets}
\label{alg:lttb}
\begin{algorithmic}[1]
\Require \textit{data} (original time series data points)
\Require \textit{threshold} (Number of data points to be returned)
\State Split the \textit{data} into equal number of \textit{buckets} as the \textit{threshold} but have the first \textit{bucket} only containing the first data point and the last bucket containing only the last data point
\State Select the point in the first \textit{bucket}
\For {each \textit{bucket} except the first and last}
  \State Rank every point in the \textit{bucket} by calculating the area of a triangle it forms with the selected point in the last bucket and the average point in the next
  \State Select the point with the highest rank within the \textit{bucket}
bucket
\EndFor
\State Select the point in the last \textit{bucket} (There is only one)
\end{algorithmic}
\end{algorithm}

\begin{figure}[h!]
\centering
\includegraphics[width=10cm]{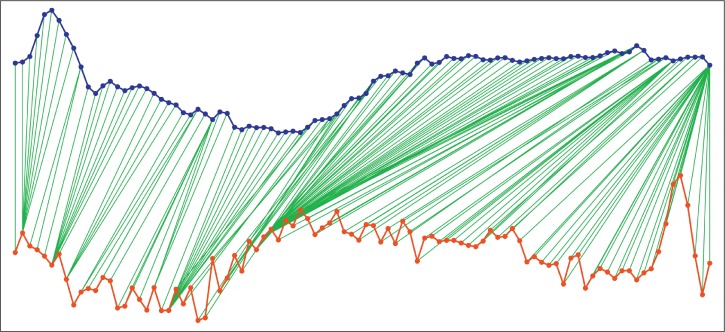}
\caption{Dynamic Time Warping Alignment \citep{Zhang2017DynamicTW} }
\label{fig:dtw_alignment}
\end{figure}

Since LTTB produces downsampled irregular time series as shown in Figure \ref{fig:lttb_approx}, Dynamic Time Warping (DTW)\citep{Berndt1994UsingDT} was a natural fit to calculate time series distances. DTW also accomodates temporal shifts and distortions that frequently occur between similar narrative arcs or story shapes as shown in Figure \ref{fig:dtw_alignment}. The python library dtaidistance \citep{dtaidistance} was chosen since it calculates the DTW distance matrix and provides visual agglomerative hierarchical clustering dendrograms. This simple, flexible and popular combination of data-driven methods extracts key features for calculating distances, is highly scalable, and imposes minimal additional assumptions that could potentially distort the accuracy of clustering.  

The sentiment time series for the corpora range from 1,399 to 13,258 data points, one for each line in a novel.  Experimentation with DTW showed that all corpora could be drastically compressed while preserving distinctive shape and major features. With as little as 25 data points DTW produced clustering of all novels similar to clustering in higher dimensions. Hierarchical clustering dendrograms produced for every novel are included in Appendix \ref{app_c}.

\begin{figure}[!ht]
\centering
\includegraphics[width=.6\linewidth]{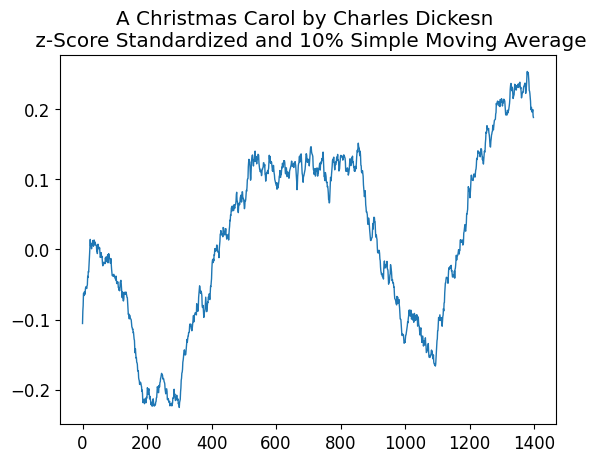}
\caption{Smoothed Plot of \textit{A Christmas Carol} by Charles Dickens (1,339 data points)}
\label{fig:fig_sma_plain_christmascarol}
\end{figure}

\begin{figure}[!ht]
\centering
\includegraphics[width=.6\linewidth]{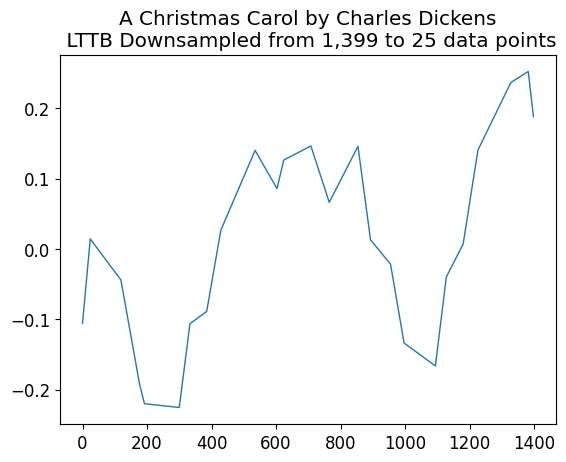}
\caption{Downsampled Plot of \textit{A Christmas Carol} by Charles Dickens (25 data points)}
\label{fig:fig_lttb_plain_christmascarol}
\end{figure}

\clearpage
\section{Metrics}
\label{sec:metrics}

The key insight that enables this self-supervised approach is the automated generation of a synthetic ground truth in lieu of manually annotated training sets. This synthetic time series baseline provides two main benefits. First, exhaustive search over the entire ensemble of models provides confidence in model selection and performance for any particular narrative text. Second, the latent temporal structure inherent in every cohesive narrative allows human-in-the-loop experts to quickly identify, query and validate the critical features among 10,000s of distinct sample points, dozens of competing models and hundreds of unique corpus:model combinations.

The ensemble median is used to calculate three new novel metrics below. The ensemble is weighted toward simple explainable lexicon models to establish a clear baseline. Models with sentiment arcs significantly different from the ensemble median can often be quickly filtered out by manually verifying a few key features like start/end points or peaks/valleys. Occasionally, divergent features reveal important crux points or interesting anomalies missed by the ensemble median. Following the G. Box adage that "All models are wrong, but some are useful," SentimentArcs gives detailed metrics and rankings of every corpus:model combination and harvests useful information from each, instead of discarding all but the ensemble consensus.

\subsection{Metric 1: Model-Corpus Compatibility (MCC) (see Appendix B)}
\label{sec:metrics_mcc}

'\textbf{Model Corpus Incompatibility (MCI)}' measures the inverse of the length-normalized Euclidean distances between a given model arc and the median ensemble sentiment time series. Visually, this is inversely proportional to the area between these two curves. MCI indicates how close a particular model's sentiment arc is to the ensemble median consensus. A lower MCC indicates that a model diverges from the ensemble median with a lower signal-to-noise ratio, but in some cases it may warrant exploration. MCI values are only meaningful relative to each other, not as stand-alone values. Fortunately, divergent plots with significant peak and valley differences are relatively few and can be quickly investigated by examining the corresponding line and surrounding context. Special attention is warranted where two or more divergent models agree on unusual features or other local or global trends.

\textbf{Model-Corpus Compatibility Metric for Model i:}
\begin{align}
\newline
\textbf{MCC}_{i} = \frac{len(TS_{i})}{ \lvert TS_{i} - TS_{corpus{\_}median} \rvert }
\end{align}

\subsection{Metric 2: Ensemble-Corpus Compatibility (ECC) (see Appendix B)}
\label{sec:metrics_ecc}

'\textbf{Ensemble Corpus Compatibility (ECC)}' provides a metric that indicates how well the entire ensemble performs on a particular narrative. It is a floating point number proportional to the inverse of the sum of the Euclidian-normed distance between each model and the ensemble median. The Ensemble Corpus Compatibility metric is (a) a machine-friendly floating point value to automate rankings and (b) a metric for generating human-friendly visualizations to quickly estimate the coherence of the ensemble and individual models on a given corpus. A high ECC metric is associated with more coherent plots and closer agreement among ensemble models. When ECC is small, the plots are less coherent. This suggests less confidence in the stability and applicability of the ensemble for the given corpora. A small ECC indicates more human intervention may be necessary to identify the most accurate models and significant features.

\textbf{Ensemble-Corpus Compatibility Metric for each Model i over every corpus j:}
\begin{equation*}
\newline
\textbf{ECC}_{i} = \frac{1}{\sum_{model=i,corpus=j}{\frac{\lvert TS_{i j} - TS_{corpus{\_}median_{j}} \rvert }{length(TS_{i})}}}
\end{equation*}

This metric measures the stability of the ensemble median for each narrative in the corpora. For each novel, the length-normalized Euclidean distance between a particular model's sentiment arc and the ensemble median (MCI) is calculated. ECC is the inverse of the sum of all such model distances for the novel. When plotted for every novel, as in Figure \ref{fig:metric_ecc_vbars}, ECC gives a view into how consistently a model performs across each novel in the corpora.

Appendix B shows one bar chart per corpus, ranking all models within the ensemble for each corpus. A larger MCC value in these bar charts indicates a model is in closer agreement with the ensemble median. This metric can be interpreted in several ways. First, if a given corpus plot shows the majority of the ensemble models are largely in agreement with the group median, this lends more confidence that the median has picked up on a strong sentiment signal in the narrative text. Varying the composition of the corpora (e.g. mixing novels, plays and social media thread) should result in smaller ECC values that indicate less coherence.

Secondly, if a given corpus plot shows models largely conflict with each other and the ensemble median, it indicates more human analysis may be required to reconcile these disparities. Analysts can focus on distinctive model plots to find those in greatest agreement with human experts regarding key features. Hopefully, a few models will be able to capture key features, but it may require a human-in-the-loop to aggregate the best insights from across several models to piece together a complete interpretation of the narrative arc.

\subsection{Metric 3:  Model-Family Coherence (MFC) (see Appendix C)}
\label{sec:metrics_mfc}

\textbf{Model Family Coherence (per individual corpus or over entire corpora):} One motivating factor behind this paper was to confirm that SOTA models from traditional supervised sentiment analysis of short-texts would similarly dominate this new task of self-supervised analysis of long narratives. Specifically, language Transformers and DNNs were expected to consistently outperform other model types.

\textbf{Famliy-Model Coherence Metric over Entire Corpus:}
\begin{align*}
\newline
\textbf{FMC}_\text{family=F} &= \frac{1}{\sum_{\text{model=i} \in \text{F}}{ Model{\_}Corpus{\_}Compatibility_\text{i}}}
\end{align*}

% MFC = sum over family ( |TS_{amodel} - TS_{ensemble_median}| / len(TS_{corpus}) ) ^ -1

MFC calculates the aggregate Model-Corpus Compatibility (MCC) metric for a particular model family (e.g. lexical vs Transformer). Comparing MFC for different model families provides guidance on selecting the best models to uncover more subtle features for a particular corpora.

\section{Results}

Surprisingly, the eight large language Transformer models in the ensemble exhibited consistent disagreement between each other. The most promising Transformer model, Large RoBERTa fine-tuned on 15 sentiment datasets, often failed to detect key peak/valley features when manually checked by our literary expert, Katherine Elkins. However, it was the best performing Transformer and detected important features outside the baseline more than any other model. Importantly, it also did the best on the most challenging novel for the entire ensemble (lowest ECC): Toni Morrison's \textit{Beloved}.

As a family, the Transformer group was often the least cohesive. This can be seen visually in the ensemble plots in Appendix \ref{app_a} and the hierarchical clustering dendrograms in Appendix \ref{app_c}. In contrast to lexical models, which are usually clustered together, Transformer models seem to diverge from the mean and each other in the ensemble plots. Similarly, Transformer models are unpredictably distributed in the cluster dendrograms compared to lexical models.

Other sentiment model families like classical machine learning models and DNN models also showed varying degrees of incoherence. With these ensemble models and corpora, grouping models into families does not offer much predictive value on performance. As such, Model Family Coherence is more an observational metric that disproves our assumption that Transformers would universally outperform all other models and families.

\begin{figure}[!ht]
\centering
\includegraphics[width=.9\linewidth]{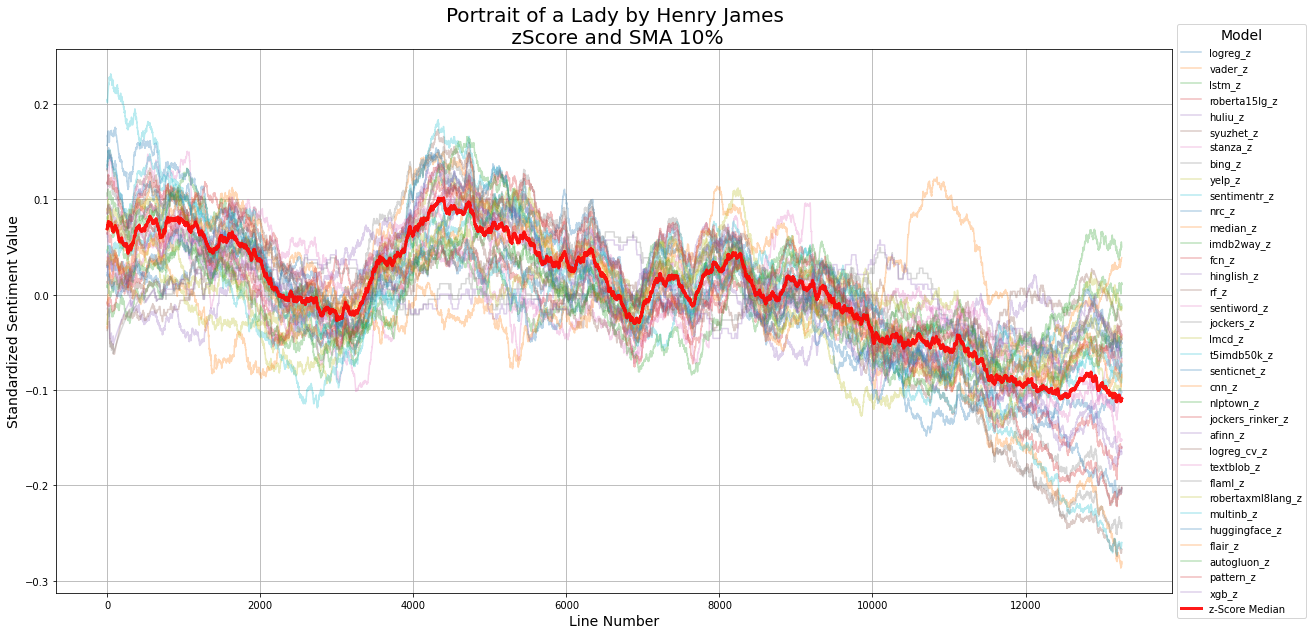}
\caption{Ensemble Sentiment Plots with Incoherent Ends}
\label{fig:fig_sma_portraitlady_results}
\end{figure}

\begin{figure}[!ht]
\centering
\includegraphics[width=.9\linewidth]{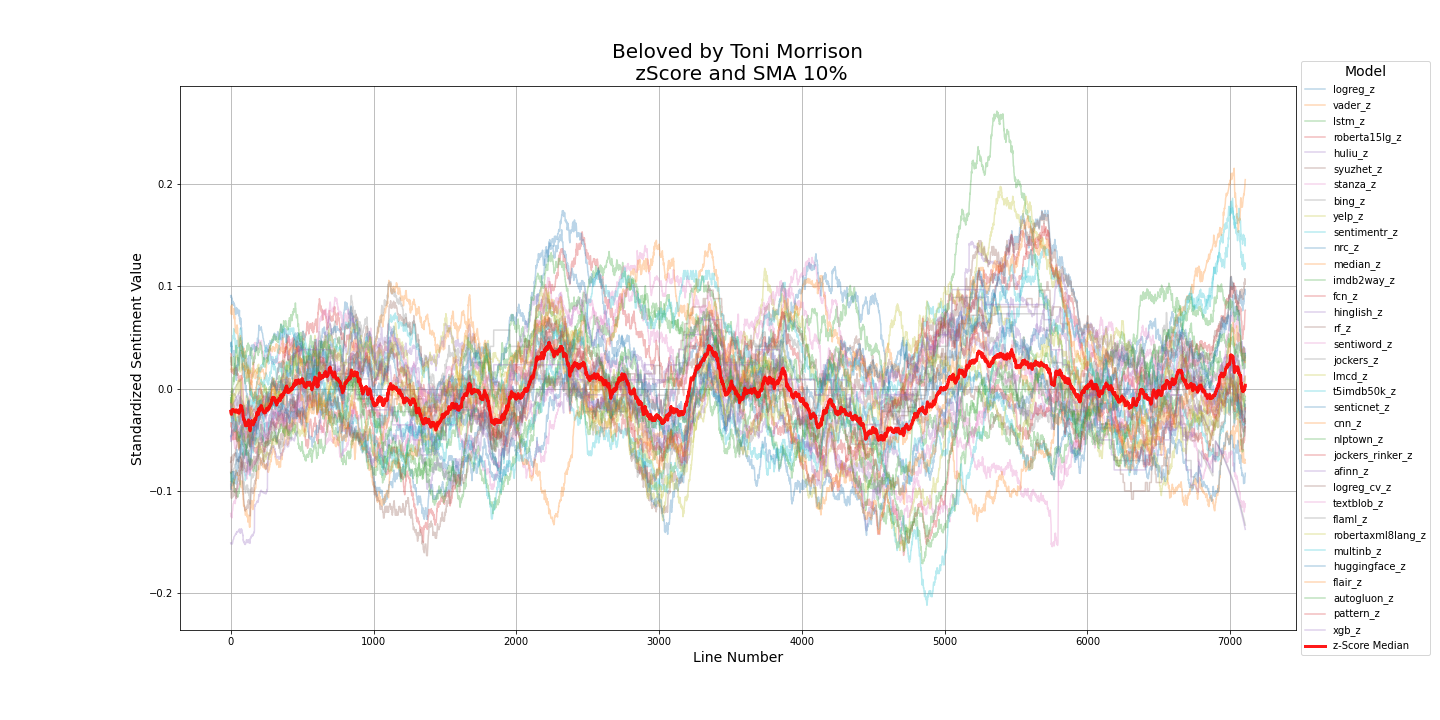}
\caption{Ensemble Sentiment Plots of Variable Incoherence}
\label{fig:fig_sma_beloved_results}
\end{figure}

\clearpage
\subsection{Model Corpus Compatibility (MCC) Across Entire Corpora}

\begin{figure}[!ht]
\centering
\includegraphics[width=.9\linewidth]{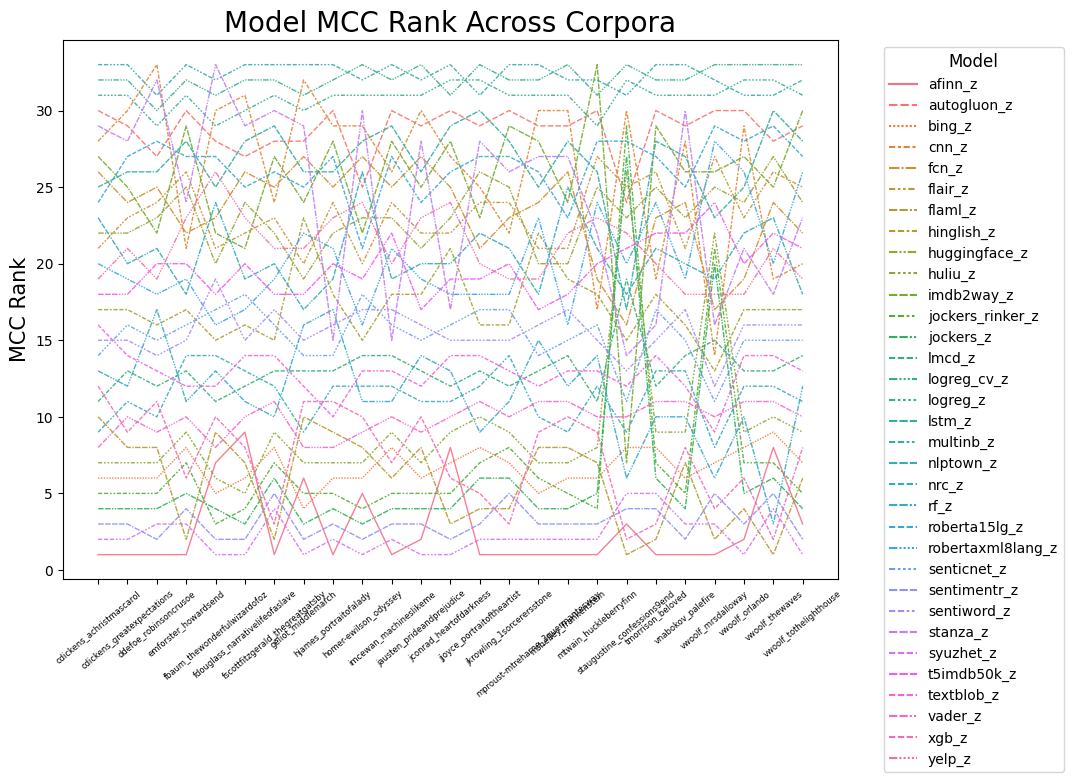}
\caption{Model-Corpus Compatibility Across All Models and Corpora}
\label{fig:metric_mcc_lines}
\end{figure}

Figure \ref{fig:metric_mcc_lines} is a detailed visualization showing how each model performs on every novel. These plots are more legible in Juypter notebooks and provide a summary view of all corpus:model sentiment arcs. A higher MCC value indicates a model's sentiment arc more closely matches the ensemble median. In these plots, models are ranked by their MCC metric.

The models are ranked based upon agreement with the ensemble median. The models with the highest agreement with the ensemble median are ordered from the bottom up. For example, the lexical models AFINN and SyuzhetR appear as the bottom two line plots for nearly every the novel in the corpora. This means they align best with the ensemble median. On the other end of the spectrum, the Multinomial Naive Bayes model and six-fold cross-validated Logistic Regression ML models in the top of Figure \ref{fig:metric_mcc_lines} are ranked as most incoherent relative to the ensemble median.

Note that the lexical AFINN model is the top ranked model for approximately half the novels and those novels tend to have straightforward narratives including an autobiography, childrens' adventure and epic adventures. AFINN falters with modern novels using more complex language and with more ambiguous plots \citep{Elkins2019CanSA}. Therefore, the suggested approach is to start analysis with the ensemble baseline to seek a baseline sentiment arc. Then, to find more subtle features and anomalies, progressively study narrative arcs further from this consensus. 

\begin{figure}[!ht]
\centering
\includegraphics[width=.9\linewidth]{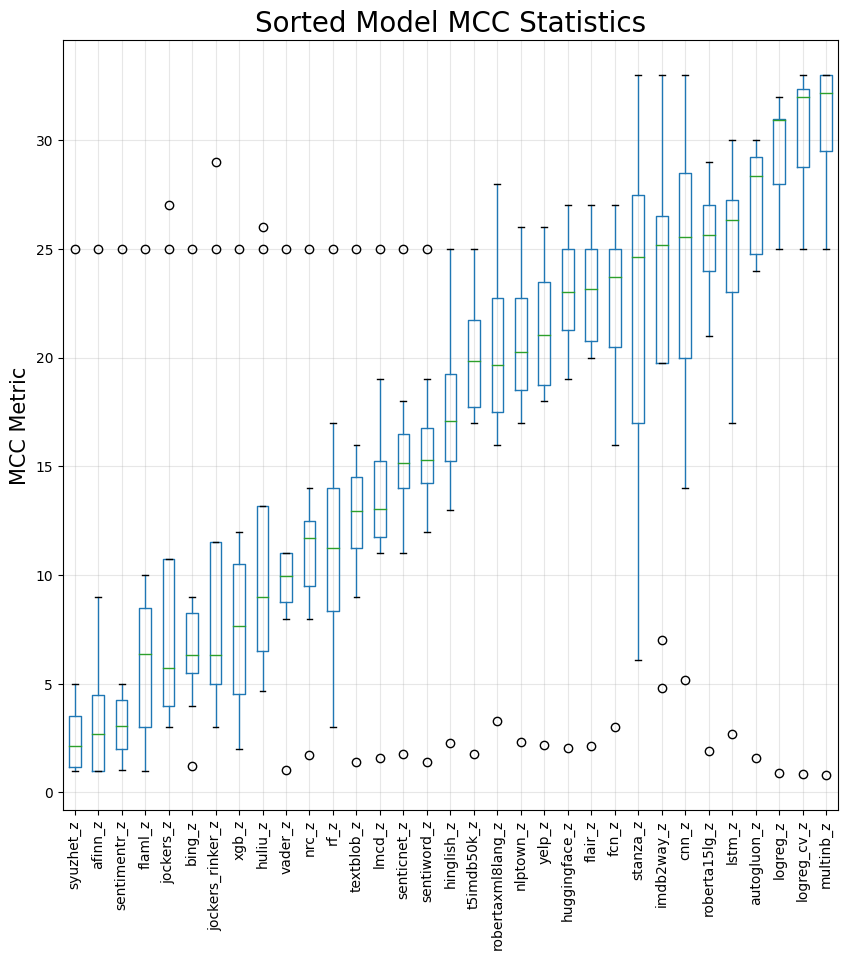}
\caption{Sample Model-Compatibility Statistics}
\label{fig:metric_mcc_box_stats}
\end{figure}

Figure \ref{fig:metric_mcc_box_stats} is a less detailed box plot visualization comparing the performance of each model across the entire corpora. Here, models are ranked from left to right according to how much they agree with the ensemble median as measured by the MCC metric. Again, the MMC values for the lexical models SyuzhetR and AFINN show the greatest agreement with the ensemble mean (although AFINN has higher variance and more distant outliers). 

In contrast, the classical ML Multinomial Naive Bayes and Logistic Regression models are the three models furthest from the ensemble median. These are rightmost bars in Figure \ref{fig:metric_mcc_box_stats}. These three models are unlikely to be discovering features missed by the Transformers. The more likely explanation is that they are simply unable to consistently perform well on this task and these particular corpora.

Stanza, and several other AutoML and ensemble approaches, stand out for having more variable MCC metrics. This is also reflected in the dendrograms of Appendix \ref{app_c} that show these models in scattered and unpredictable clustering patterns.

Most sophisticated models (Transformers and DNNs) are clustered together on the right side of Figure \ref{fig:metric_mcc_box_stats}. This indicaties they tend to disagree with the ensemble consensus. As the RoBERTa15lg model showed with Toni Morrison's \textit{Beloved}, sometimes this is because these models can pick up on sentiment expressed in more complex ways than the ensemble median. 

In Figure \ref{fig:metric_mcc_box_stats} the AutoML library FLAML and the XGBoost ensemble are the only traditionally high-performing models that are also on the left side, which shows agreement with the ensemble median. The validity of the ensemble consensus is enhanced when more advanced models appear shifted left. It would be interesting to see how many such high-performance models shift left as the corpora becomes narrower and language more standardized (e.g. court opinions, financial documents or research papers).

\clearpage
\subsection{Model Corpus Compatibility (MCC) On a Single Novel}

\begin{figure}[!ht]
\centering
\includegraphics[width=0.8\linewidth]{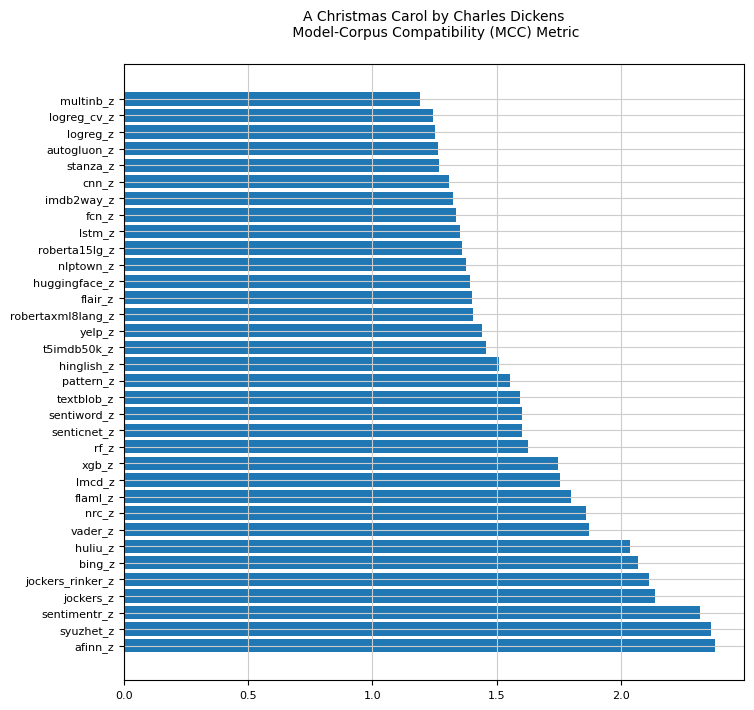}
\caption{Model-Corpus Compatibility on \textit{A Christmas Carol} by Charles Dickens}
\label{fig:metric_mcc_cdickens_acc}
\end{figure}

Figure \ref{fig:metric_mcc_cdickens_acc} ranks models by MCC values on a sample corpus, Charles Dickens' \textit{A Christmas Carol}. Greater MCC values correspond to more agreement with the ensemble median. Models with highest agreement with the ensemble mean are at the bottom and ordered by decreasing MCC upward. The MCC ranking of models for this novel correspond to the ordering seen in Figure \ref{fig:metric_mcc_box_stats}. In both figures the sentiment arcs produced by AFINN and SyuzhetR are the most coherent with the ensemble median while the three most divergent arcs are from classic ML models.

Note the clustering of performance seen within the ranking by MCC values. The highest three models are all related lexical models: AFINN, SyuzhetR and SentimentR. The next cluster contains four models: three lexical-heuristic from SentimentR (jockers, jockers-rinker and huliu) and one purely lexical model (bing). The third cluster is less distinct and more varied in model type composition. It contains (2) lexical-heuristic models (VADER, lmcd), (1) lexical model (NRC), and (2) AutoML/ensemble classical ML models (FLAML, XGBoost).  

These clustering patterns give insight to the overall coherence of the ensemble. This pattern also suggests which model(s) to explore for detecting features and anomalies missed by the ensemble median. If the the plots in Appendix \ref{app_a} show a smooth, continuous difference between models (e.g. Wilson's translation of Homer's \textit{Odyssey}) then the ensemble has a strong baseline sentiment, but it may be harder to find features and anomalies outside this baseline.  If there are discontinuous jumps between clusters of models (e.g. Joyce's \textit{Portrait of the Artist as a Young Man}), the ensemble may be less suited to the corpus and the true sentiment signal may be more distributed among high MCC clusters rather than concentrated in the ensemble median.

\clearpage
\subsection{Ensemble Corpus Compatibility (ECC) Metrics}

\begin{figure}[h!]
\centering
\includegraphics[width=.6\linewidth]{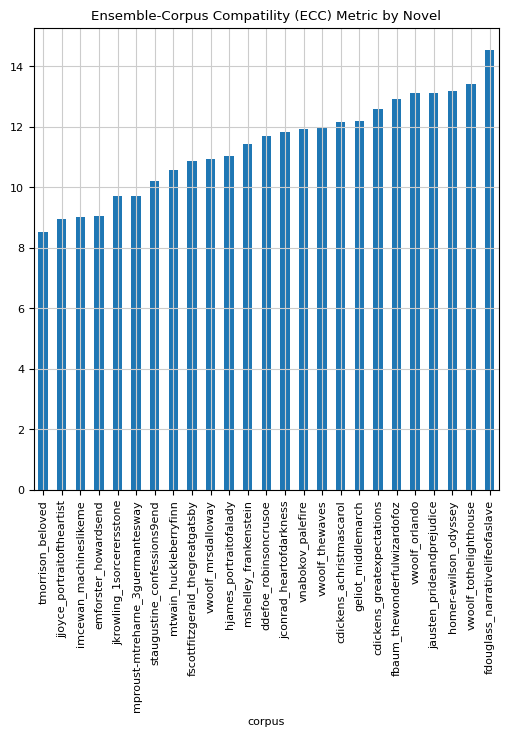}
\caption{Ensemble-Corpus Compatibility for Reference Corpus}
\label{fig:metric_ecc_vbars}
\end{figure}

A global view of ECC metrics across every novel is visualized in Figure \ref{fig:metric_ecc_vbars}. The ECC metric gives a general sense of how coherent the entire model ensemble performs for each specific novel. The lower ECC novels on the left comport with our literary expert's detailed analysis. Toni Morrison's \textit{Beloved} stands out as a very challenging novel for the ensemble with greater disagreement among the sentiment arcs in Appendix \ref{app_a}. Similarly, Joyce's \textit{Portrait of the Artist as a Young Man} uses symbolism and allusions while McEwan's contemporary novel is complicated by an AI protagonist with great emotional ambiguity.

What is more surprising is that the modernist novel \textit{To the Lighthouse} by Woolf, sometimes described as relatively 'plotless', is seen as among the most coherent novels in the corpora. In contrast, the other top ranked ECC novels use more direct writing styles: autobiographical (Douglass' \textit{Narrative of the Life of Frederick Douglass, an American Slave}), an epic adventure (Wilson's translation of Homer's \textit{Odyssey}) and a children's adventure (Baum's \textit{The Wonderful Wizard of Oz}). The two Dickens' novels are also highly ranked.

\clearpage
\subsection{Model Family Coherence Metrics (MFC)}

\begin{figure}[h!]
\centering
\includegraphics[width=.6\linewidth]{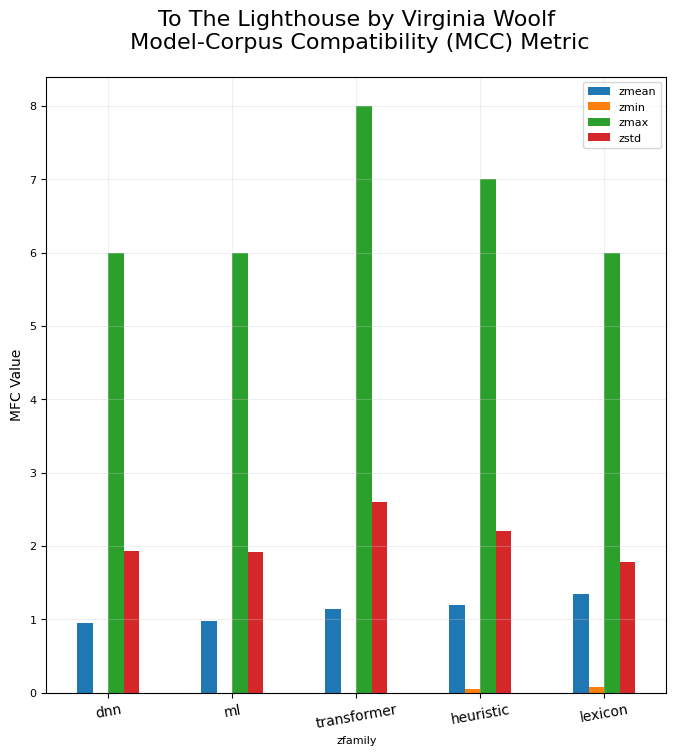}
\caption{Model-Family Coherence for a Novel in Corpus}
\label{fig:metric_mfc_vbars}
\end{figure}

For additional insight into comparative model performances, Figure \ref{fig:metric_mfc_vbars} compares the aggregate model performance of model families according to the classification in Table \ref{tab:table_models}. In particular, we wanted to compare the relative performance between baseline models (Lexical), popular industry models (Heuristic, ML) and SOTA research models (Transformer, DNN).

Two results stand out. First, SOTA Transformer models have the most variable performance as measured by MCC metrics. Although this often indicates noise, for Transformers it often suggests they've detected more subtle sentiment arc features (e.g. peaks/valleys) missed by the ensemble consensus. For example, the large RoBERTa model trained on fifteen popular sentiment analysis datasets was exceptional in finding key features in novels like Toni Morrison's \textit{Beloved}, the most challenging for the ensemble as measured by the ECC metric.

Second, the distinction between these simplified performance metrics is not as clear as the performance difference seen on traditional stand-alone, short-text sentiment analysis. This suggests time series sentiment analysis on more natural long-form narratives is an NLP task distinct from previous sentiment analysis tasks. The novel SentimentArcs framework, processing pipelines and metrics highlight the greater challenges, variability and unpredictability of this task.

\clearpage
\subsection{Sentiment Arc DTW Hierarchical Clustering}

\begin{figure}[h!]
\centering
\includegraphics[width=.9\linewidth]{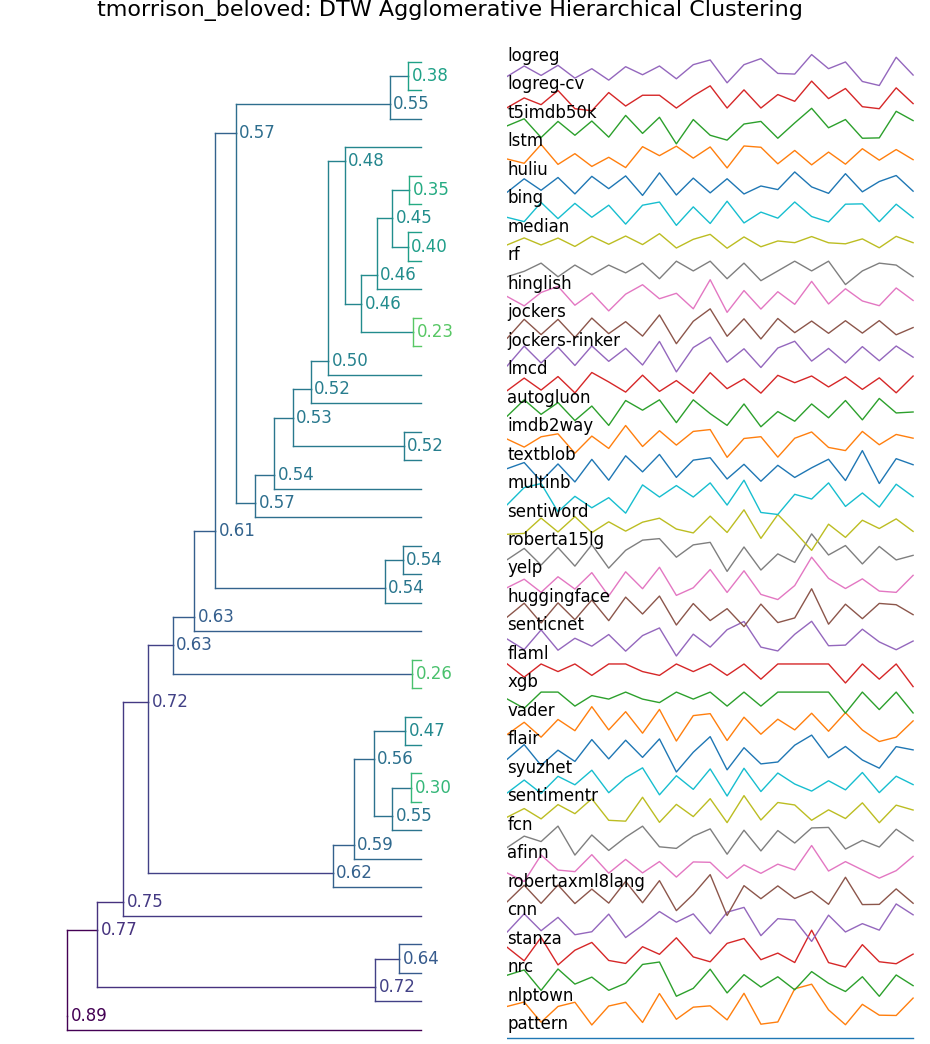}
\caption{ \textit{Beloved} by Toni Morrison}
\label{fig:fig_hclust_beloved}
\end{figure}

\begin{figure}[h!]
\centering
\includegraphics[width=.9\linewidth]{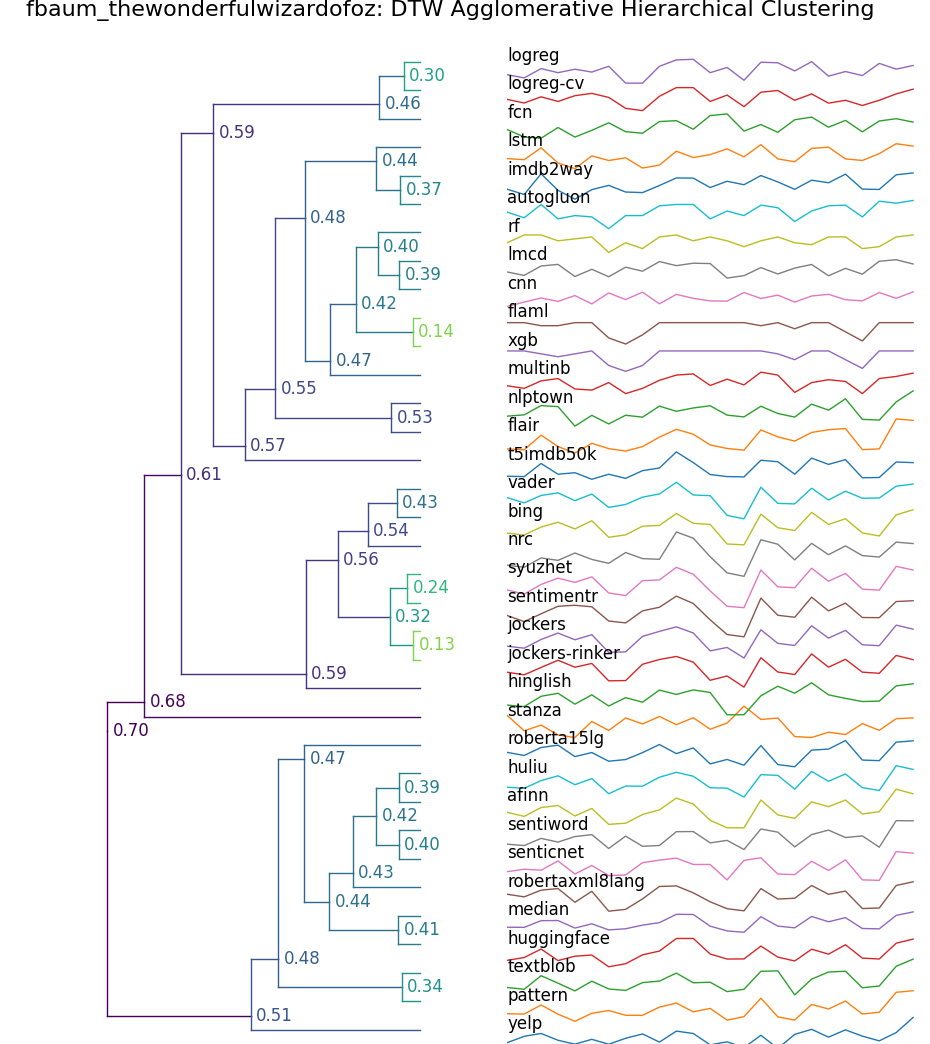}
\caption{ \textit{The Wonderful World of Oz} by Frank Baum}
\label{fig:fig_hclust_wizardoz}
\end{figure}

Figure \ref{fig:fig_hclust_beloved} shows an agglomerative hierarchical dendrogram for the simplified sentiment arcs generated from Morrison's \textit{Beloved}. Figure \ref{fig:fig_hclust_wizardoz} shows the dendrogram for Baum's \textit{The Wonderful Wizard of Oz}. The right half of all dendrograms show all simplified sentiment arcs labeled with the model name that generated it. The left halves show an agglomerative hierarchical tree that clusters sentiment arcs according to the distance matrix generated by the Dynamic Time Warping algorithm. The number at each junction give a numeric measure of distance between the time series and/or cluster of time series. 

These dendrograms show how different models and families of models perform differently on each corpus. Baum's \textit{The Wonderful Wizard of Oz} has more distinct group clustering from top to bottom: (a) outliers (logreg, logreg-cv, fcn), (b) SOTAs (Transformers, DNNs and ensembles), (c) lexicons and (d) mixed SOTA (Mostly Transformers and Lexicons). In contrast, Morrison's \textit{Beloved} has four clusters from top to bottom (a) outliers (logreg, logreg-cv, t5imdb50k), (b) Large Mixed SOTA, (c) isolated Transformers (Yelp, huggingface), (d) mixed lexicons (lexicons, FCN, robertaxml8lang, flair), (e) SOTA (NLPTown, CNN, Stanza, NRC). Morrison's dendrogram less distinct clusters show greater mixing of model families. This reinforces the findings in Section \ref{sec:metrics_mfc} that \textit{Beloved} is the most challenging novel for the ensemble as measured by MFC in contrast to \textit{The Wizard of Oz}'s more coherent sentiment arcs.   

\clearpage
\section{Future Work}

SentimentArcs is being expanded with more models and corpora. Future additional lexicons include SO-CAL, Harvard General Inquirer and LIWC. Classic ML models will include SVM. FastText provides a popular subword embedding model to the ensemble. If resources permit, more DNN training details and specialized architectures like TextCNN and SentiXBBoost's stacked ensembles would be valuable additions. New AutoML libraries are being explored including TPOT and AutoKeras. Huggingface's rapidly growing hub of pretrained Transformer models offers the possibility to incorporate more diverse architectures and non-English models.

New corpora in the form of additional novels or long-form narratives from entirely different domains like finance, social media threads/groups or temporal compilations of news articles are high on the priority list. Again, we welcome collaboration with domain experts from diverse disciplines to help expand our existing corpora or create entirely new reference corpora for other domains.

This paper introduces SentimentArcs to process, quantify, rank and cluster narrative arcs. An upcoming complimentary paper will explore identifying, localizing, ranking and interpreting features within narrative arcs. Scholars have used computational approaches on corpora of dozens to thousands of novels to detect general trends in literature in a technique called distant reading \citep{moretti2013distant} \citep{Reagan2016TheEA}. This upcoming paper introduces a new form of computational intermediate reading with new metrics and methodologies to detect, quantify and contextualize features within a single novel's narrative arc.

\clearpage
\subsection{DTW Hierarchical Clustering Dendrograms}

\begin{figure}[h!]
\centering
\includegraphics[width=.5\linewidth]{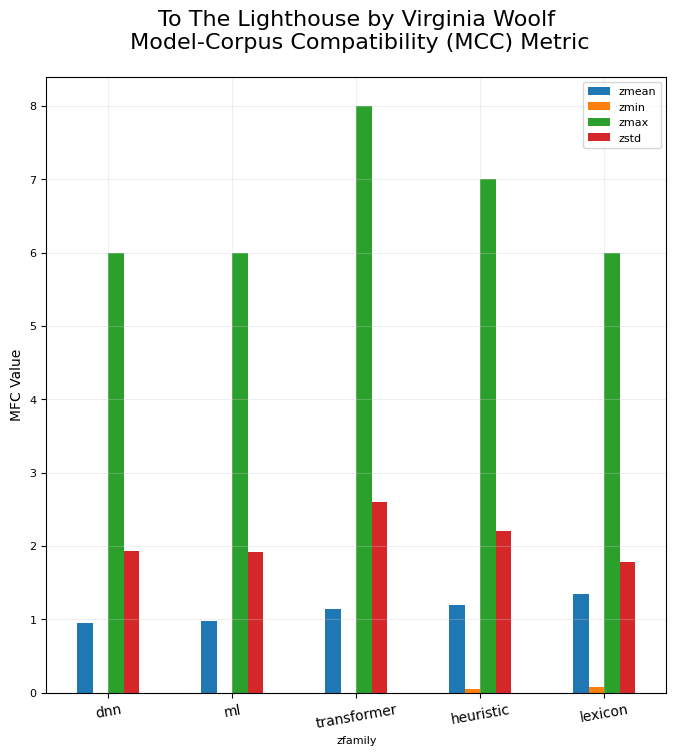}
\caption{Sample DTW Hierarchical Clustering Dendrogram}
\label{fig:metric_dtw_dendrogram}
\end{figure}

\section{Limitations}

SentimentArcs is designed to address the most difficult aspects of text sentiment analysis, but it comes with limitations inherent to the task and application domain. Coherent long narratives offer a latent structure in the form of auto-regressive time dependence that can be exploited using visual narrative arcs. These narratives often involve complex, parallel, contradictory and surprising subplots, conflicting character motivations, and ambiguous or ambivalent psychological states. Unlike typical short text sentiment classifiers, language in narrative is often intentionally manipulated for artistic effect, and this makes the semantic meaning more indirect, diffuse, or even misleading. The domain expert human-in-the-loop is an essential arbitrator of ground truth for these types of texts.

Literary scholars are averse to the very concept of a single ‘ground truth,’ and this is why SentimentArcs is designed to search the problem space of all corpus:model combinations exhaustively, to provide selection metrics, and to incorporate human-in-the-loop experts efficiently. High-stakes analysis of financial or intelligence OSINT narratives also merit careful analysis of results and a critique of findings. For the most challenging texts and applications, SentimentArcs can drastically reduce resource demands on human experts in order to confirm existing beliefs and uncover new features of complex narratives. It is not a turn-key solution, however.

\section{Conclusions}

This paper introduced SentimentArcs and a novel methodology for self-supervised time series sentiment analysis of long-form narratives.  It includes the largest collection of sentiment analysis time series analysis (840) we know of from over two dozen literary novels and almost three dozen implemented models representing most architectures. The large ensemble of models provides a synthetic ground truth, and a unique time series processing pipeline is fast and flexible, producing new metrics for quantifying, ranking and automating joint corpus:model performance optimization over all possible combinations. Simple human-digestible visualizations enable rapid human-in-the-loop explanation, verification and feature detection.

\section{Impact Statement}

NLP sentiment analysis can provide valuable insights into the desires, beliefs, and personalities of humans on a large scale. Adding a temporal dimension to the sentiment analysis of an individual, an organization or a topic enables the construction of much more personalized, detailed, and predictive models for both individual and group psychology \citep{Li2021MicromacroDO}. SentimentArcs advances sentiment analysis of long text narratives with applications far beyond literary analysis: improving health care \citep{Sud2021TimeSB} \citep{Sabra2018PredictionOV} \citep{Sanglerdsinlapachai2021ImprovingSA}, making markets more efficient \citep{Anand2020WhoMT} \citep{Sharpe2020ThePO}, and detecting criminal or terrorist threats on social media \citep{Gaikwad2021OnlineED} \citep{Mansour2018SocialMA} \citep{Ahmad2019DetectionAC}.   

Sentiment analysis, when integrated into a stimulus-response loop as in manipulated search results \citep{Epstein2015TheSE} or social newsfeeds \citep{fbgoel2014nyt}, can also provide feedback to help shape the beliefs and actions of humans on a large scale \citep{Mihltz2015BeyondSS}. When combined with multimodal data like facial recognition or online behavioral data, sentiment analysis can enable the manipulation of humans to provoke anger and increase engagement \citep{Fan2014AngerIM}. Without human-in-the-loop supervision, these automated systems have been shown to induce depression, low self-esteem, and even suicide in social media users, especially in the young \citep{Rousseau2017TheRA} \citep{Steers2014SeeingEE}. By providing a semi-supervised, scalable, and efficient human-in-the-loop solution to narrative sentiment analysis, we hope SentimentArcs offers a new and better methodology for responsible parties to deploy these models with more oversight and better outcomes for all.

%Bibliography

% \usepackage{natbib}
% \bibliographystyle{unsrt} % plainat} % unsrtnat}
% \title{Bibliography management: \texttt{natbib} package}
% \author{Jon Chun}
% \date {Oct 2021}

% \bibliographystyle{abbrvnat}  

\bibliography{references}  

\medskip

% \begin{appendices}
% \appendix

\clearpage
\appendix
\section{Appendix A. Ensemble Sentiment Arcs}
\label{app_a}

For every novel in the corpora, a visualization of the sentiment arcs generated by all ensemble models is presented here. As described in Section \ref{sec:stdsmooth}, these plots are the result of starting with the time series sentiment values produced by each model and then (a) converting to floating point values where necessary/possible, (b) standardizing using z-scores and (c) smoothing with a 10\% simple moving average.

These visualization provide (a) a general overview of how coherent the ensemble models perform on a given corpus, which model(s) potentially detect more subtle features or anomalies, important features like peaks/valleys and sections of the narrative where individual or groups of models have higher or lower confidence in detecting a strong sentiment signal.

\begin{figure}[!ht]
\centering
\includegraphics[width=.9\linewidth]{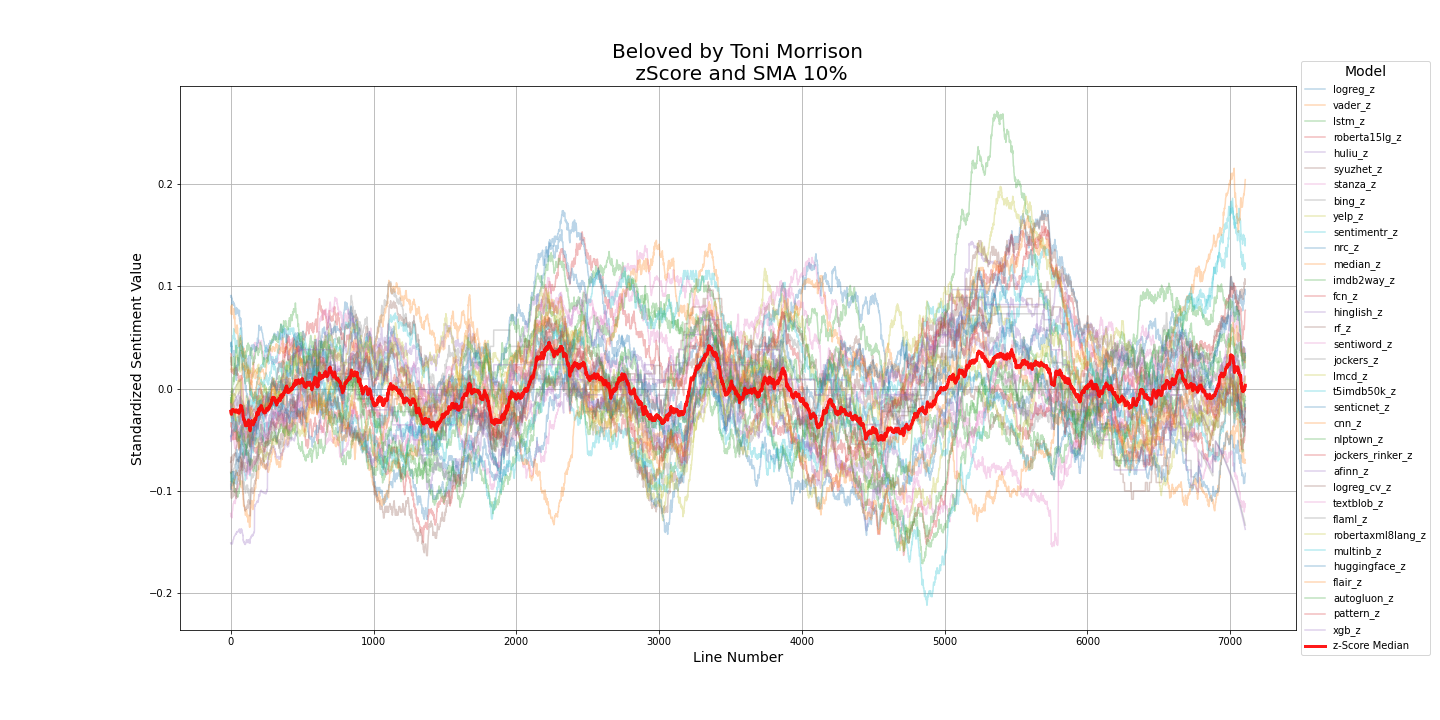}
\caption{Less Cohesive Ensemble Sentiment Plots}
\label{appfig:fig_sma_beloved}
\end{figure}

\begin{figure}[!ht]
\centering
\includegraphics[width=.9\linewidth]{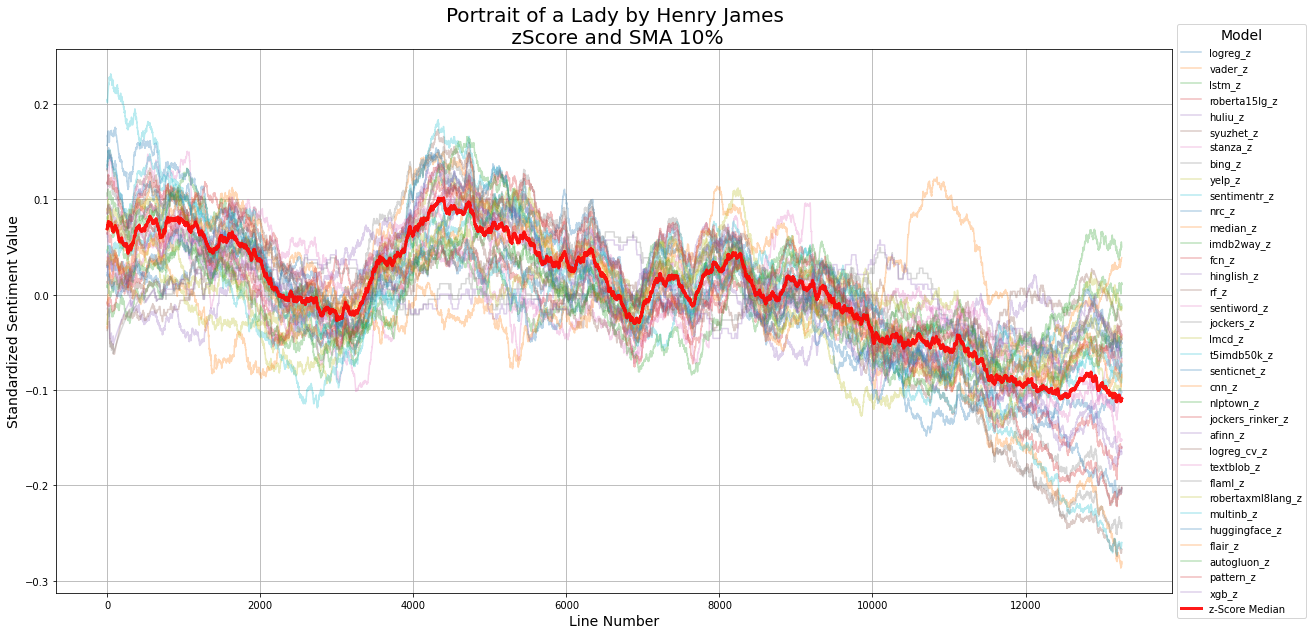}
\caption{More Cohesive Ensemble Sentiment Plots Except at Ends}
\label{appfig:fig_sma_portraitlady}
\end{figure}

\begin{figure}[!ht]
\centering
\includegraphics[width=0.9\linewidth]{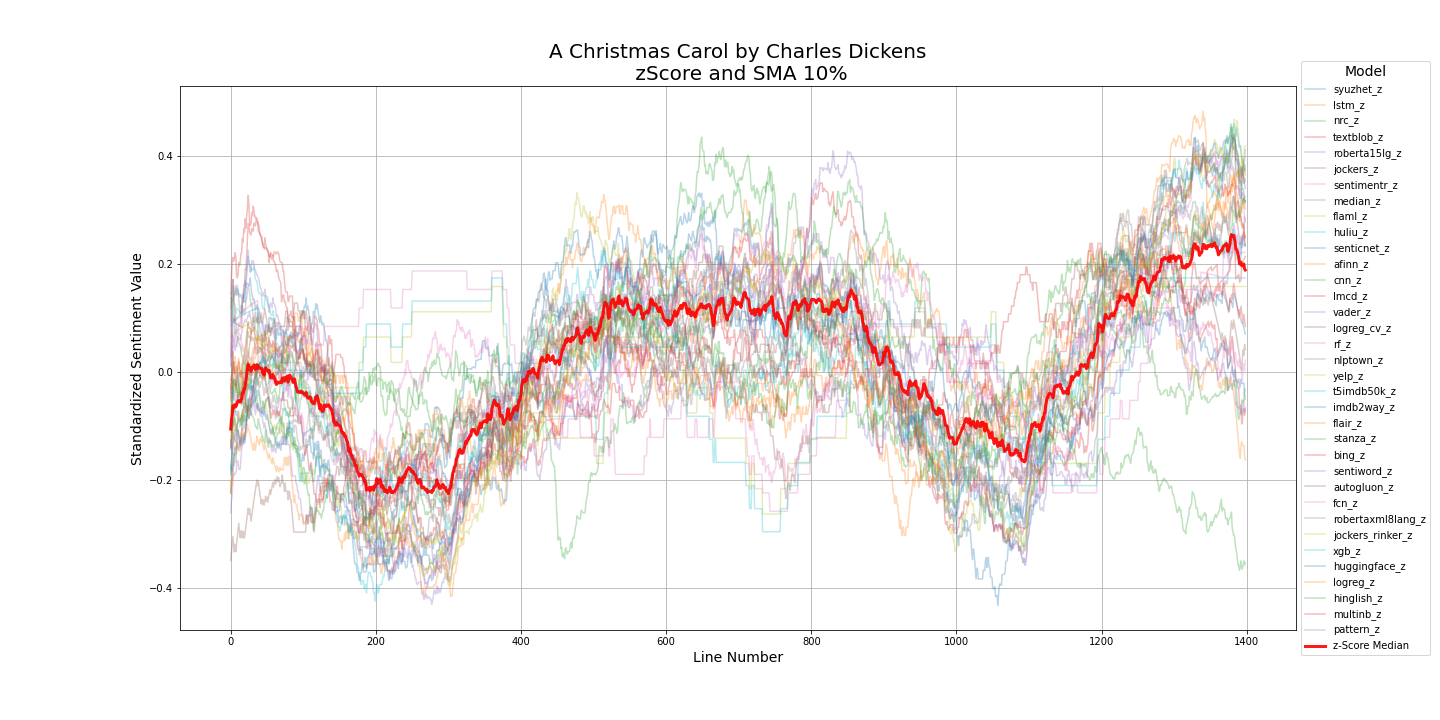}
\caption{ \textit{A Christmas Carol} by Charles Dickens}
\label{appfig:metric_esp_cdickens_acc}
\end{figure}

\begin{figure}[!ht]
\centering
\includegraphics[width=0.9\linewidth]{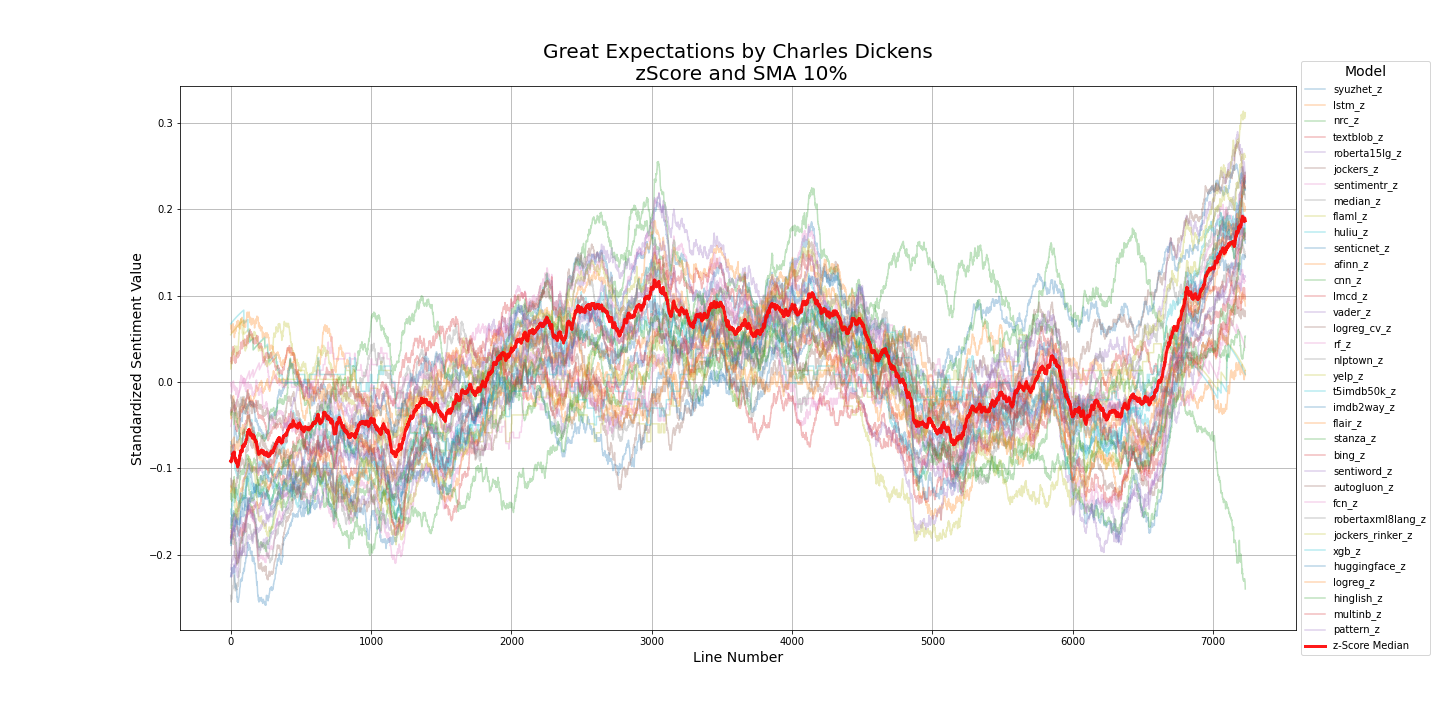}
\caption{ \textit{Great Expectations} by Charles Dickens}
\label{appfig:metric_esp_cdickens_ge}
\end{figure}

\begin{figure}[!ht]
\centering
\includegraphics[width=0.9\linewidth]{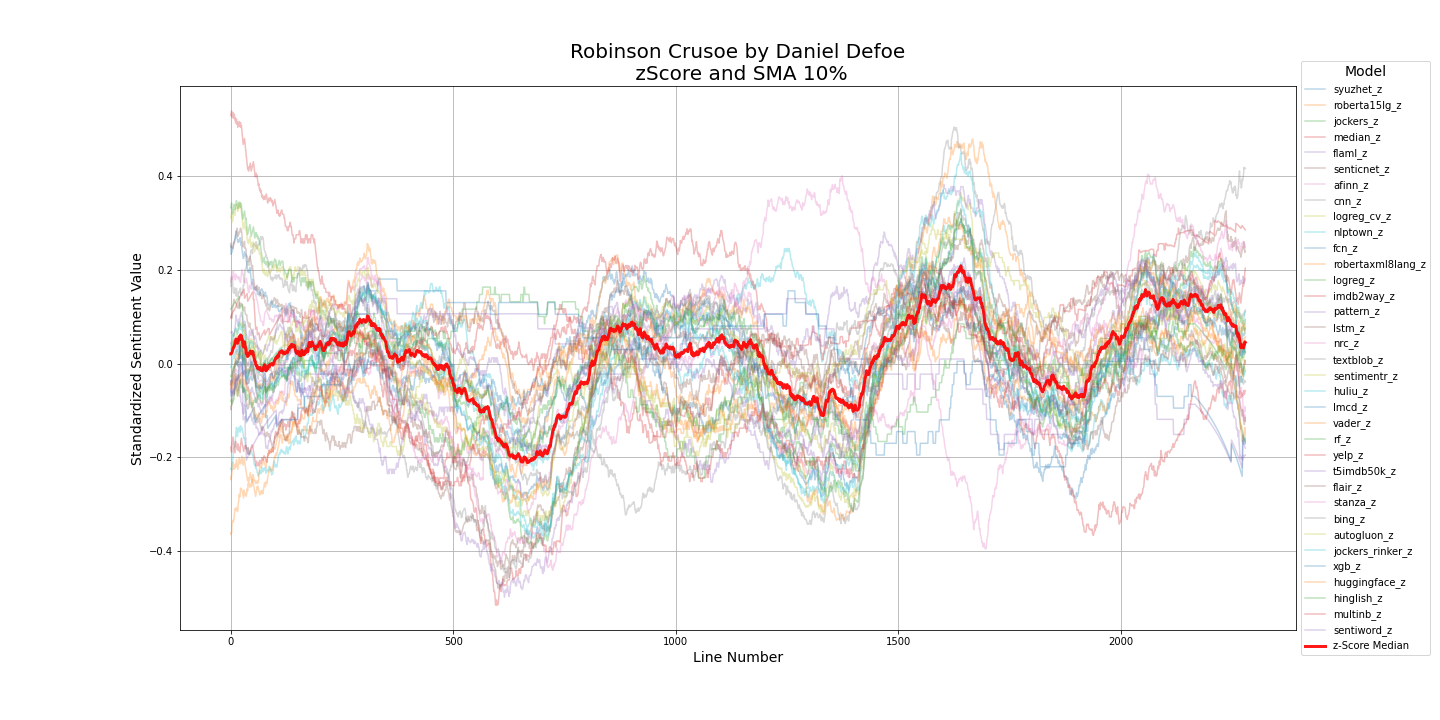}
\caption{ \textit{Robinson Crusoe} by Daniel Defoe}
\label{appfig:metric_esp_ddefoe_rc}
\end{figure}

\begin{figure}[!ht]
\centering
\includegraphics[width=0.9\linewidth]{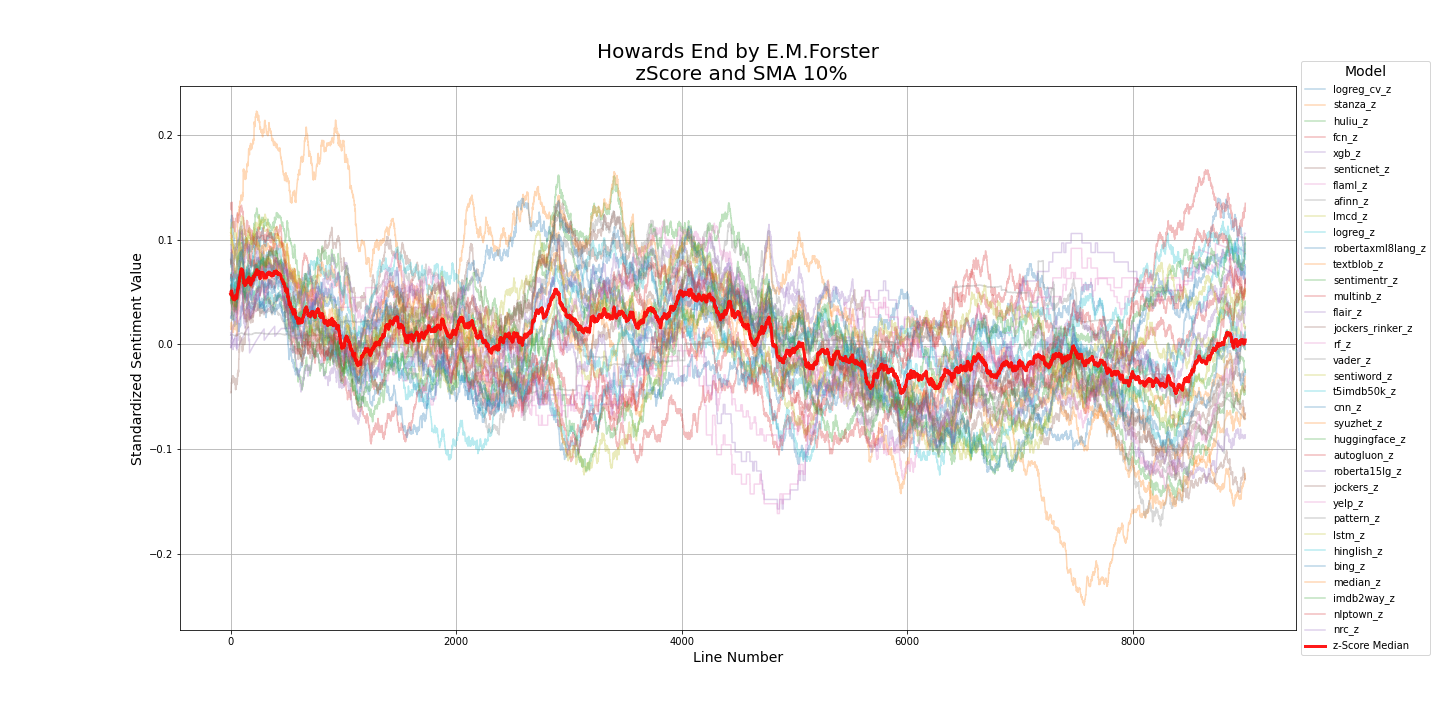}
\caption{ \textit{Howards End} by E. M. Forster}
\label{appfig:metric_esp_emforster_he}
\end{figure}

\begin{figure}[!ht]
\centering
\includegraphics[width=0.9\linewidth]{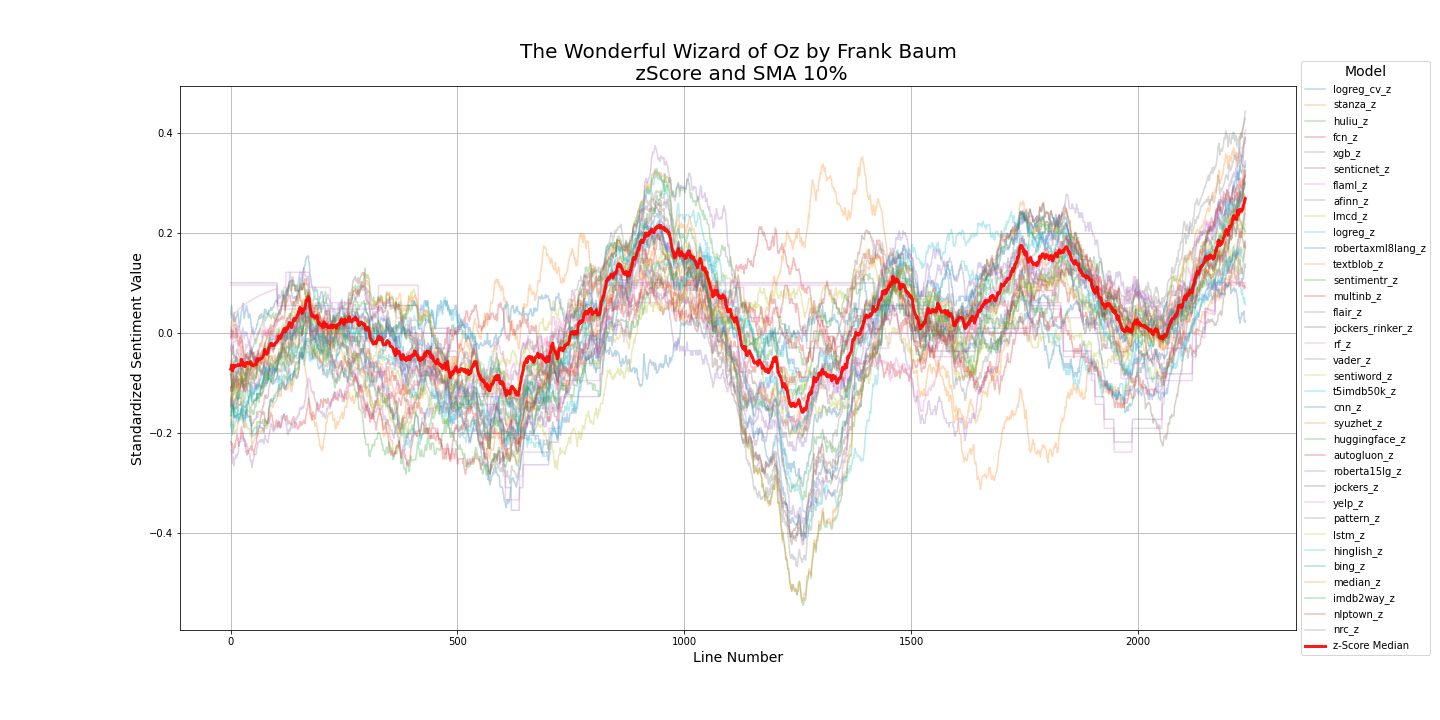}
\caption{ \textit{The Wonderful Wizard of Oz} by Frank Baum}
\label{appfig:metric_esp_fbaum_twwoo}
\end{figure}

\begin{figure}[!ht]
\centering
\includegraphics[width=0.9\linewidth]{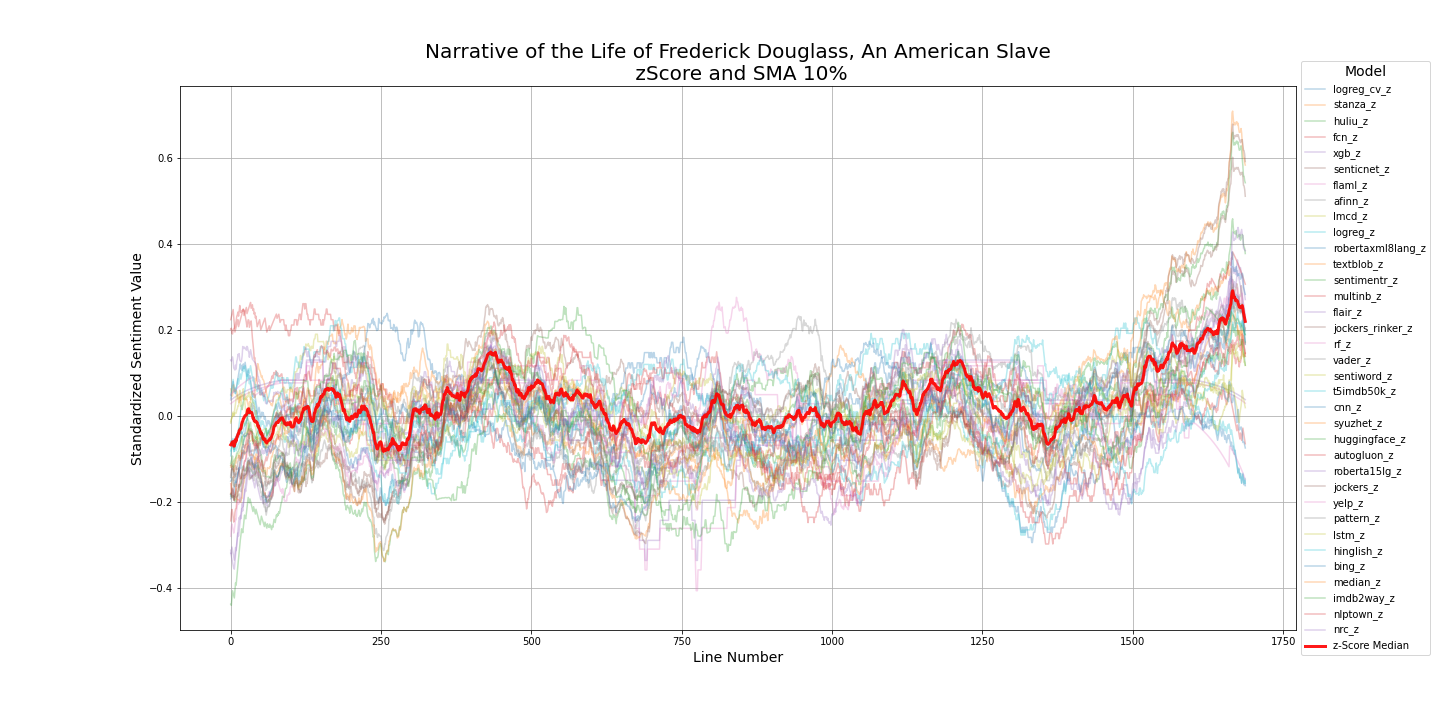}
\caption{ \textit{The Narrative of the Life of Frederick Douglass, an American Slave} by Frederick Douglass}
\label{appfig:metric_esp_fdouglass_tnotlofdaas}
\end{figure}

\begin{figure}[!ht]
\centering
\includegraphics[width=0.9\linewidth]{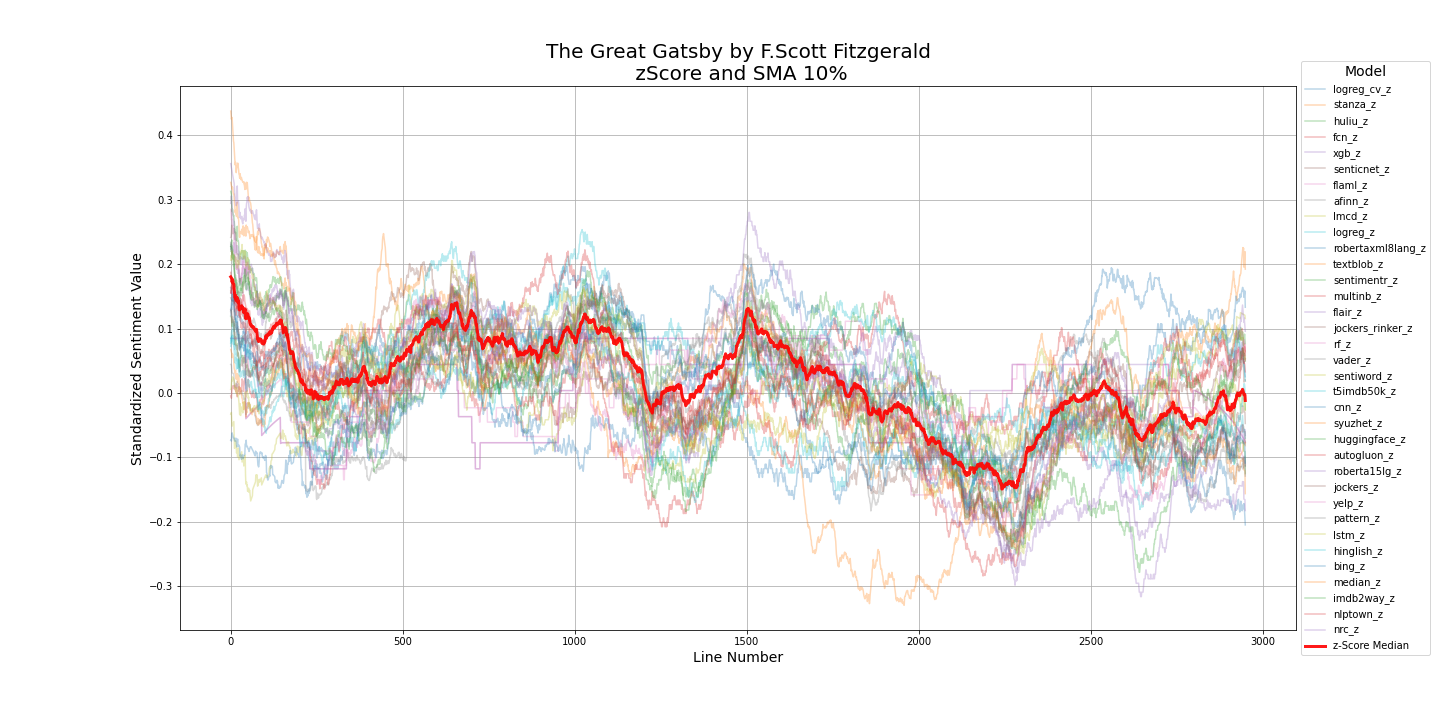}
\caption{ \textit{The Great Gatsby} by F. Scott Fitzgerald}
\label{appfig:metric_esp_fsfitzgerald_gg}
\end{figure}

\begin{figure}[!ht]
\centering
\includegraphics[width=0.9\linewidth]{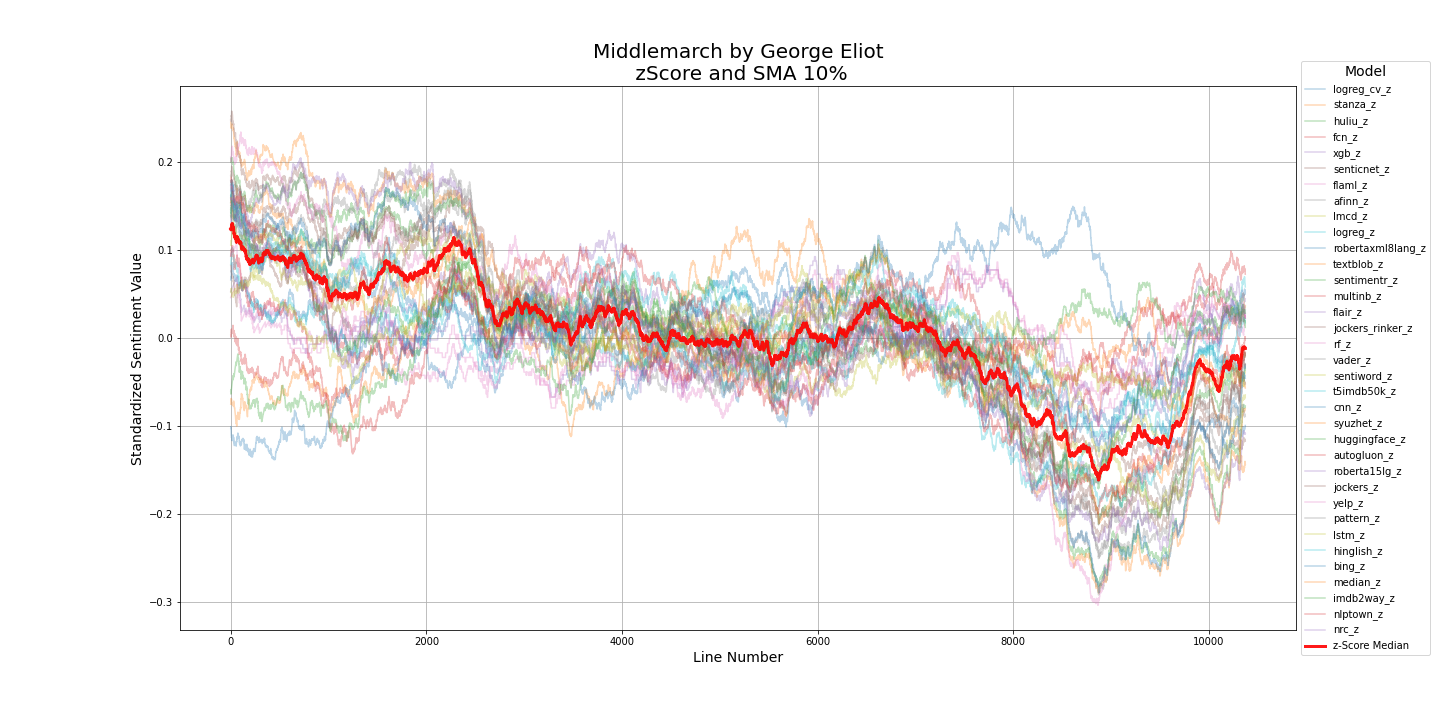}
\caption{ \textit{Middlemarch} by George Eliot}
\label{appfig:metric_esp_geliot_mm}
\end{figure}

\begin{figure}[!ht]
\centering
\includegraphics[width=0.9\linewidth]{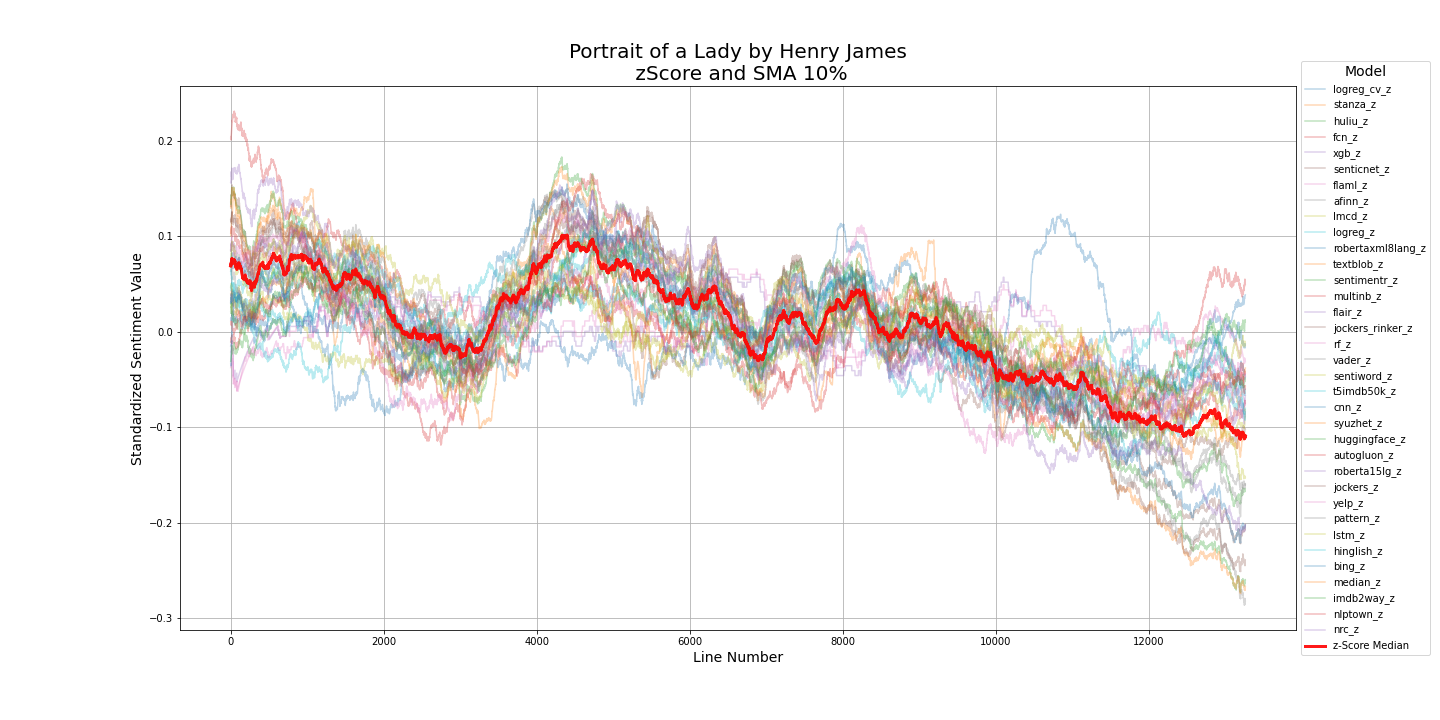}
\caption{ \textit{Portrait of a Lady} by Henry James}
\label{appfig:metric_esp_hjames_poal}
\end{figure}

\begin{figure}[!ht]
\centering
\includegraphics[width=0.9\linewidth]{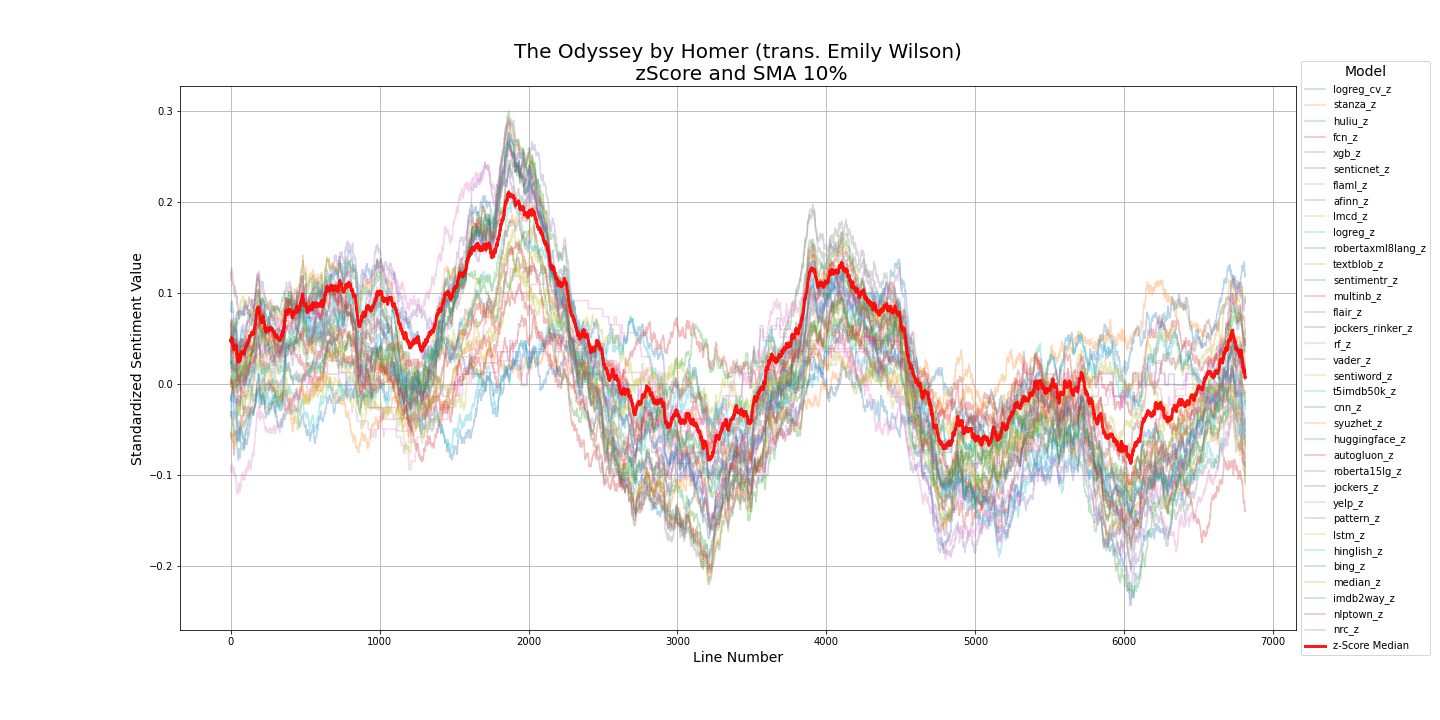}
\caption{ \textit{Odyssey} by Homer (trans. Emily Wilson) }
\label{appfig:metric_esp_homerwilson_o}
\end{figure}

\begin{figure}[!ht]
\centering
\includegraphics[width=0.9\linewidth]{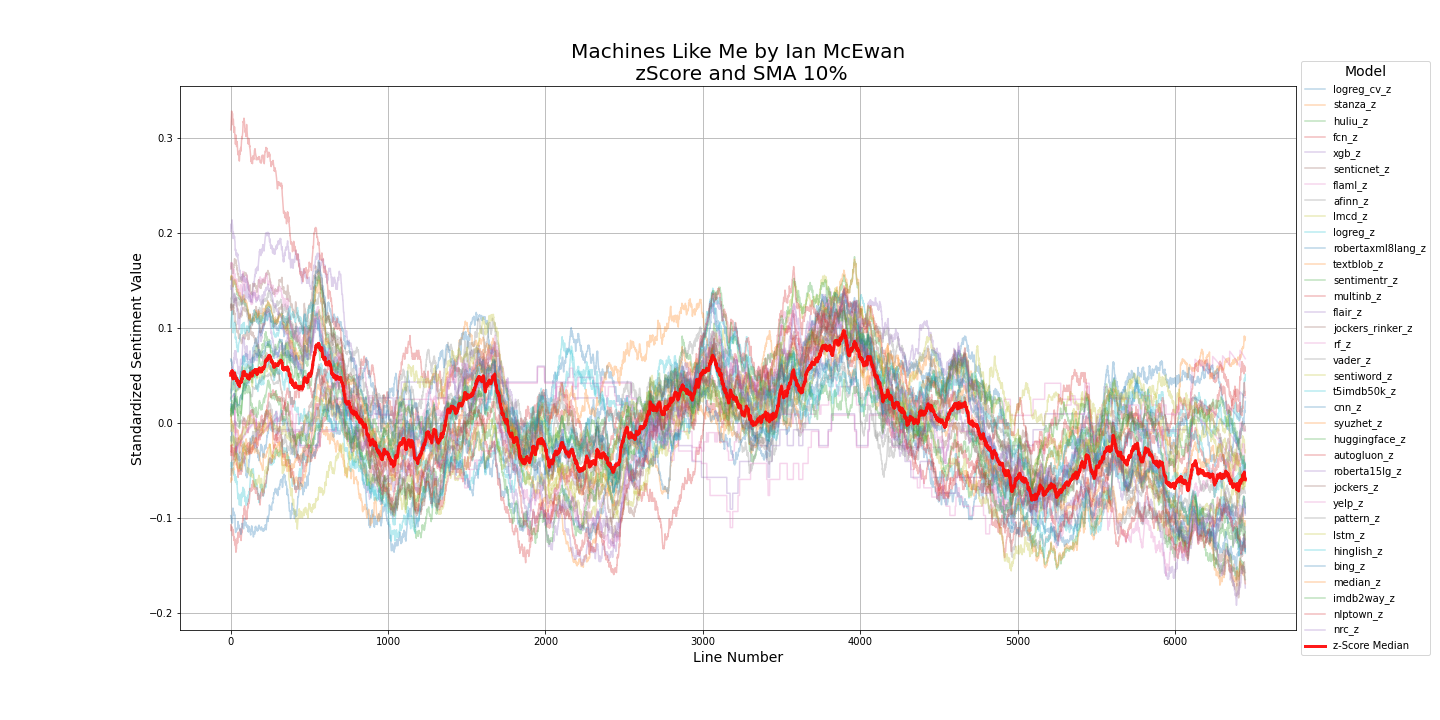}
\caption{ \textit{Machines Like Me} by Ian McEwan}
\label{appfig:metric_esp_imcewan_mlm}
\end{figure}

\begin{figure}[!ht]
\centering
\includegraphics[width=0.9\linewidth]{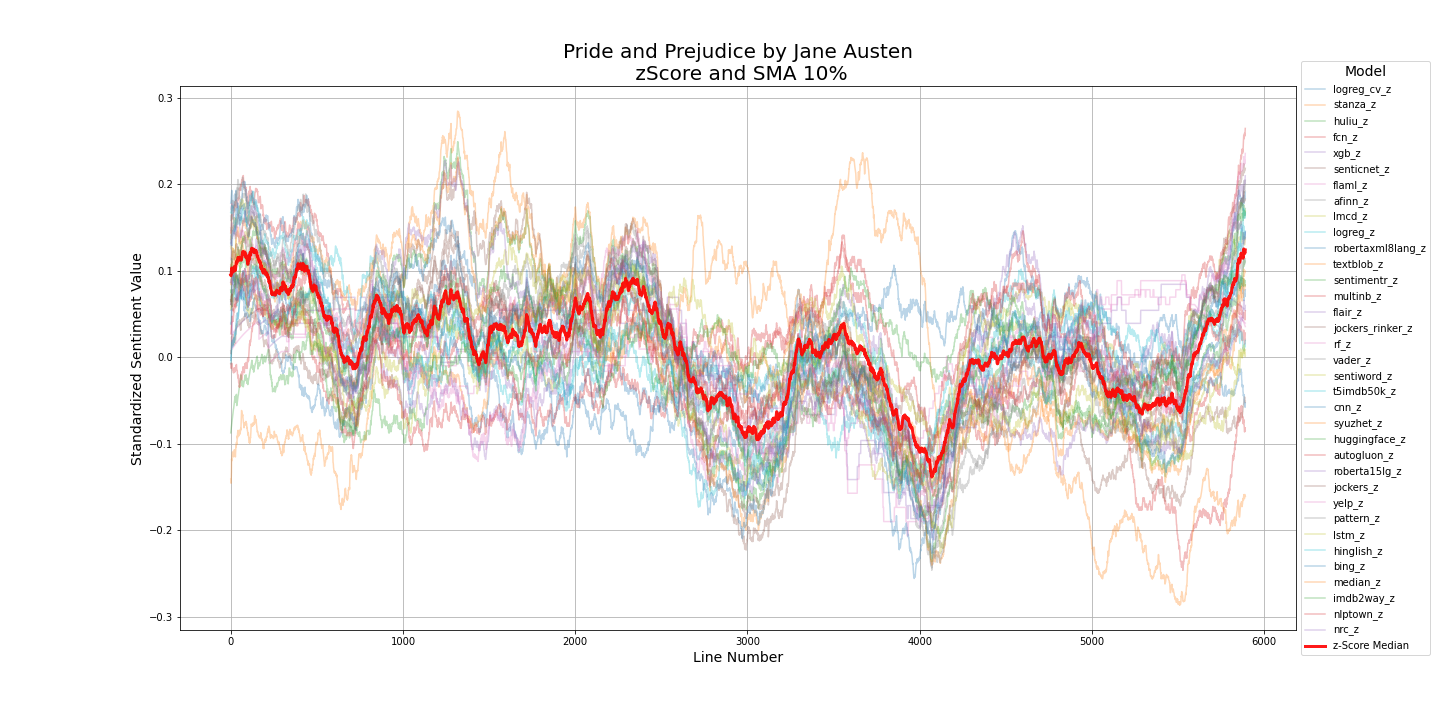}
\caption{ \textit{Pride and Prejudice} by Jane Austen }
\label{appfig:metric_esp_jausten_pap}
\end{figure}

\begin{figure}[!ht]
\centering
\includegraphics[width=0.9\linewidth]{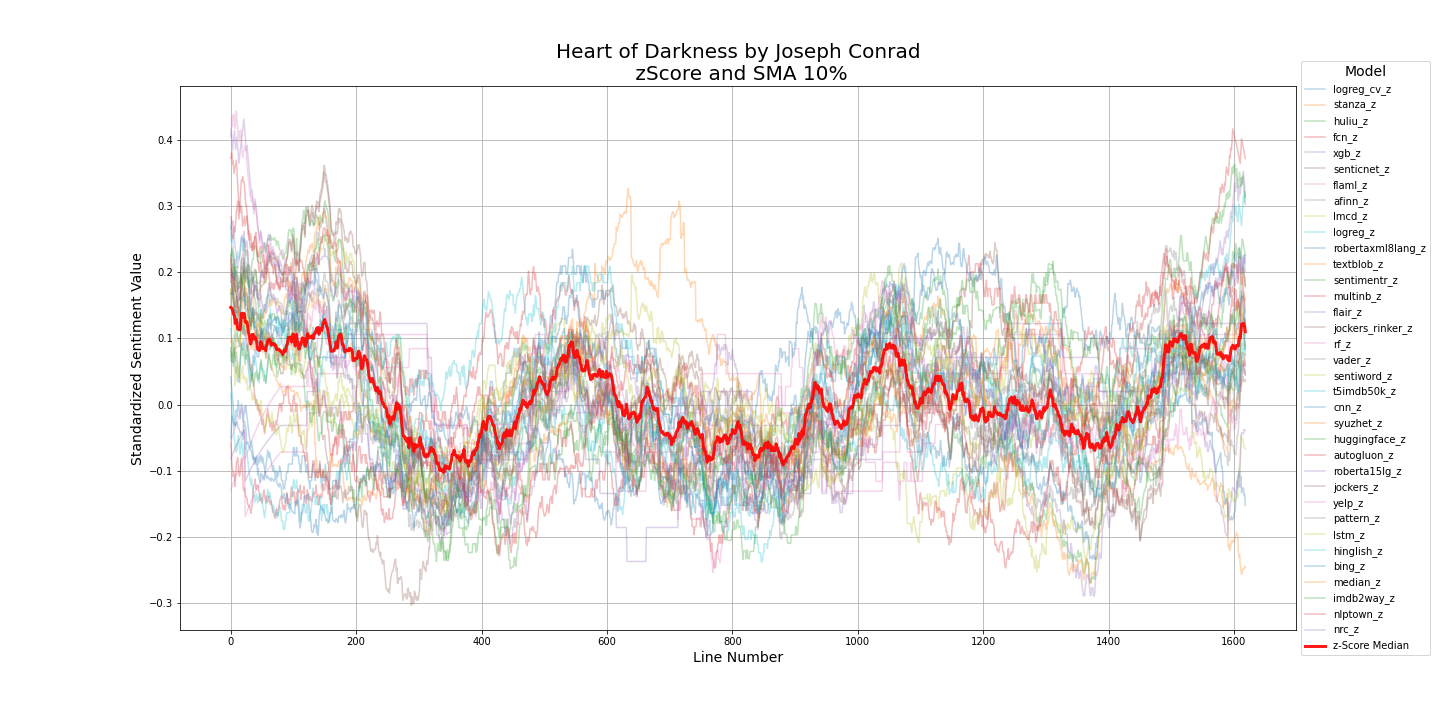}
\caption{ \textit{Heart of Darkness} by Joseph Conrad}
\label{appfig:metric_esp_jconrad_hod}
\end{figure}

\begin{figure}[!ht]
\centering
\includegraphics[width=0.9\linewidth]{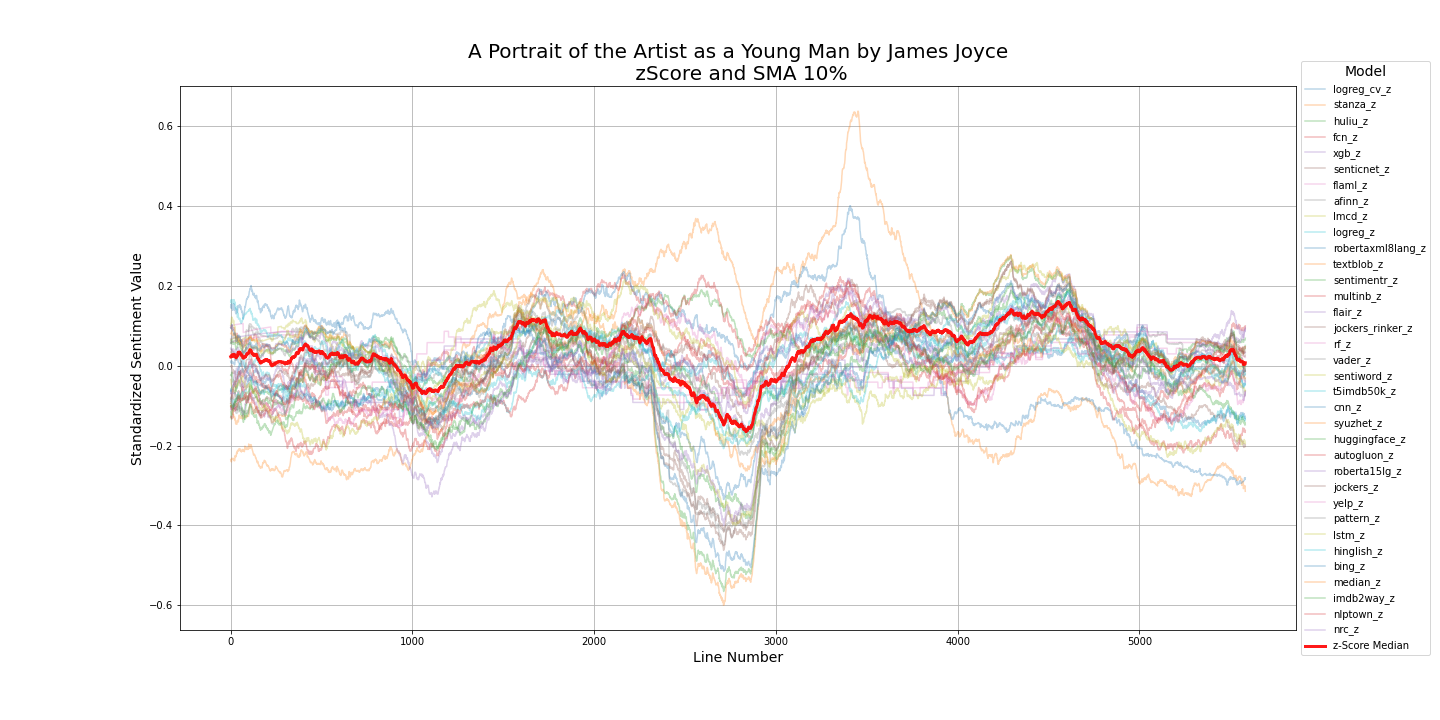}
\caption{ \textit{A Portrait of the Artists as a Young Man} by James Joyce }
\label{appfig:metric_esp_jjoyce_apotaaaym}
\end{figure}

\begin{figure}[!ht]
\centering
\includegraphics[width=0.9\linewidth]{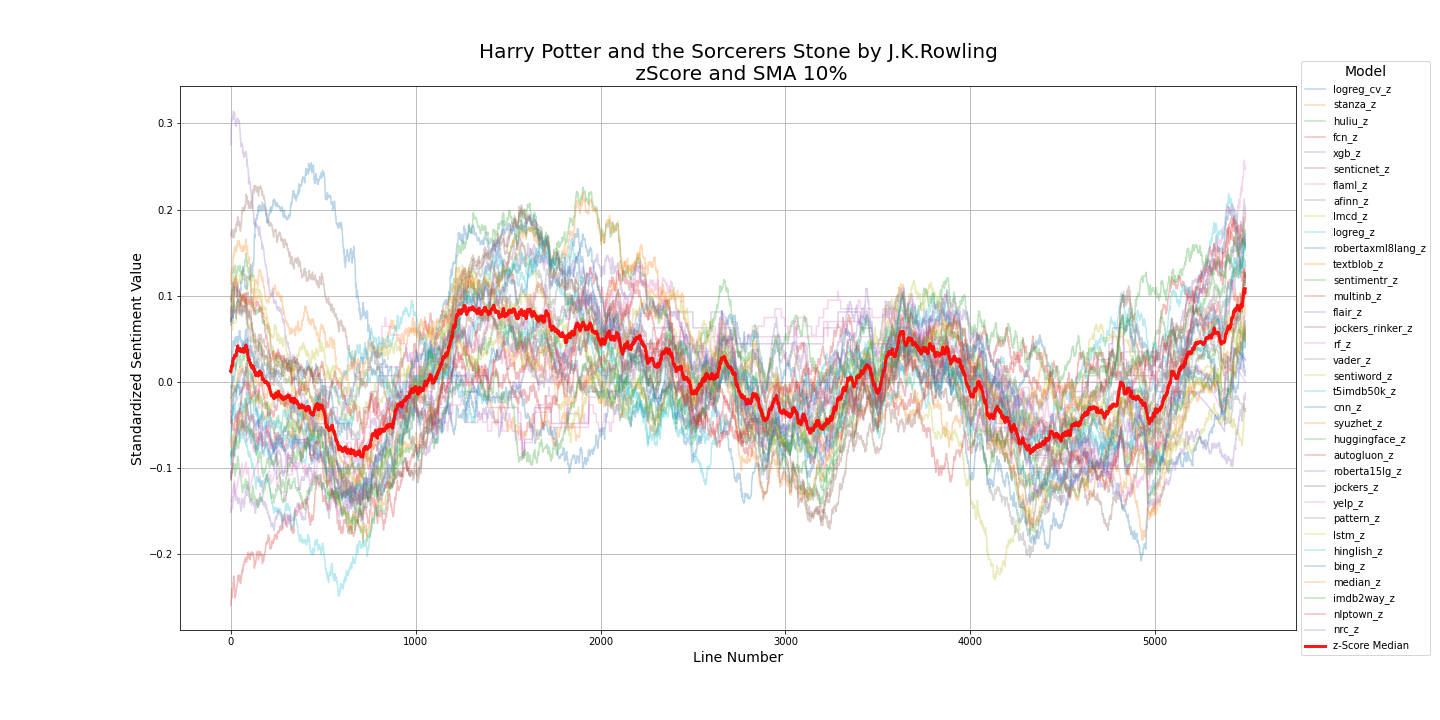}
\caption{ \textit{Harry Potter and the Sorcerer's Stone} by J. K. Rowling}
\label{appfig:metric_esp_jkrowling_hpatss}
\end{figure}

\begin{figure}[!ht]
\centering
\includegraphics[width=0.9\linewidth]{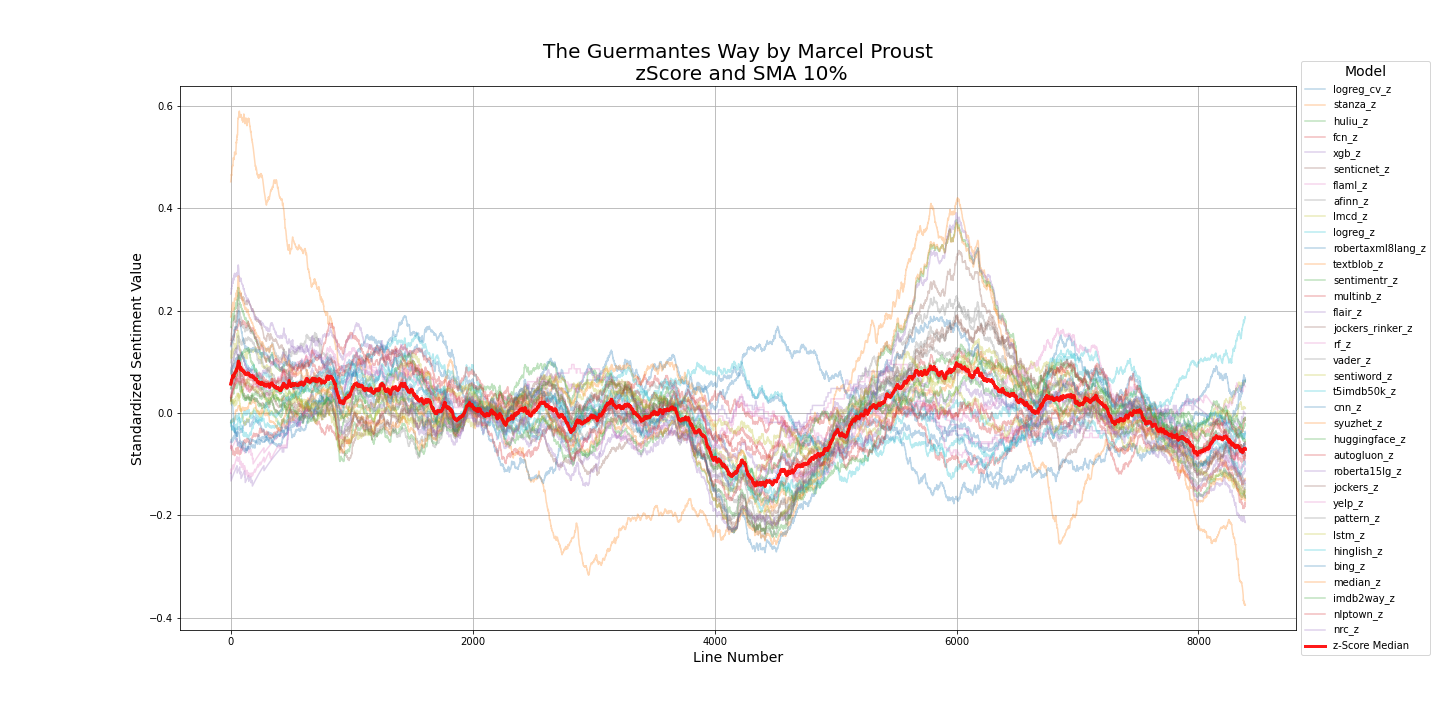}
\caption{ \textit{The Guermantes Way} by Marcel Proust (trans. Mark Treharne) }
\label{appfig:metric_esp_mprousttreharne_tgw}
\end{figure}

\begin{figure}[!ht]
\centering
\includegraphics[width=0.9\linewidth]{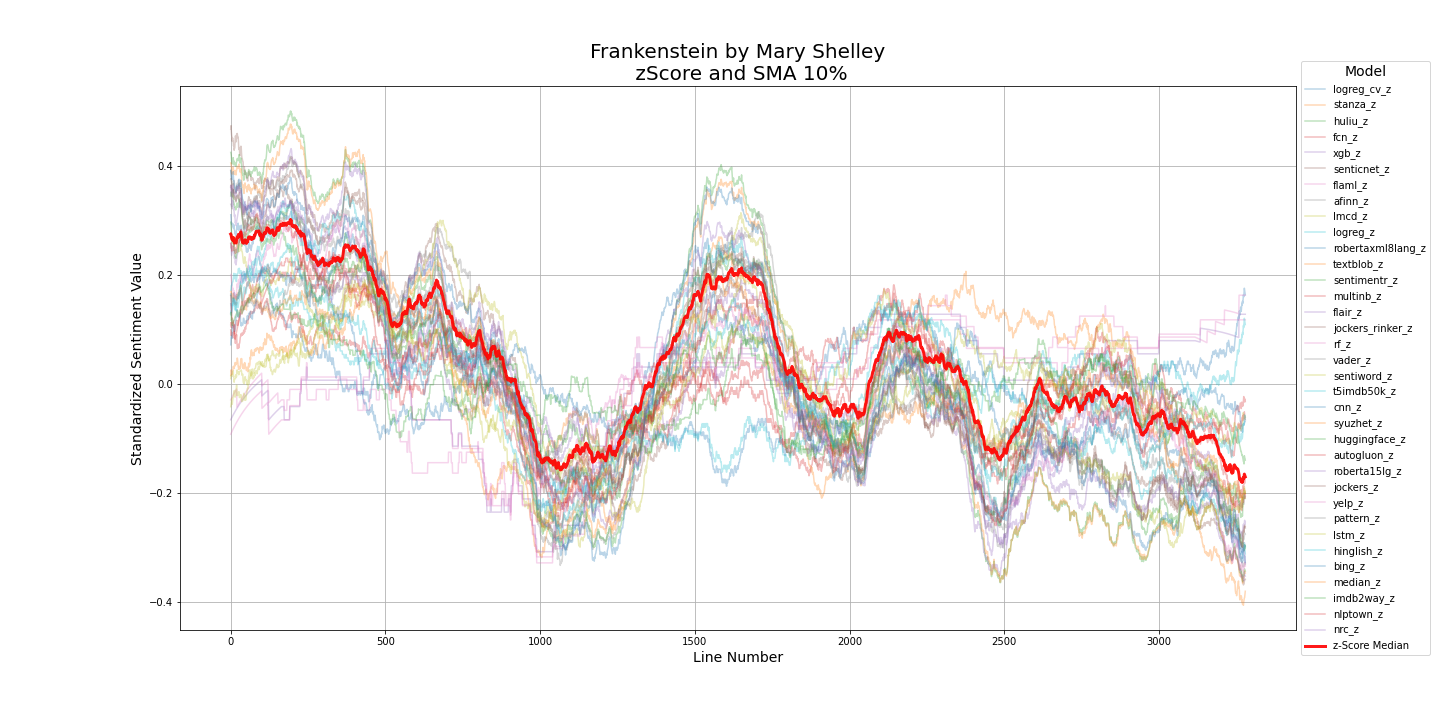}
\caption{ \textit{Frankenstein} by Mary Shelley}
\label{appfig:metric_esp_mshelley_f}
\end{figure}

\begin{figure}[!ht]
\centering
\includegraphics[width=0.9\linewidth]{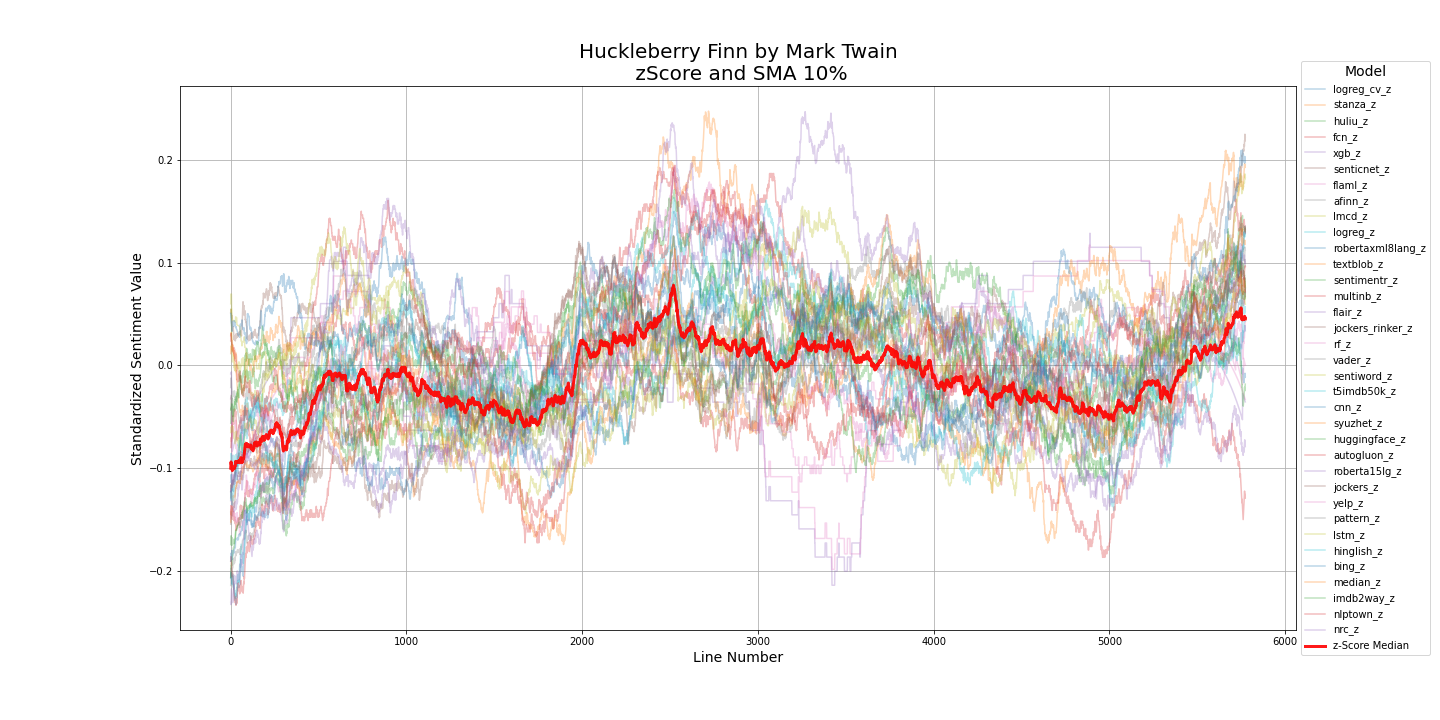}
\caption{ \textit{Huckleberry Finn} by Mark Twain}
\label{appfig:metric_esp_mtwain_hf}
\end{figure}

\begin{figure}[!ht]
\centering
\includegraphics[width=0.9\linewidth]{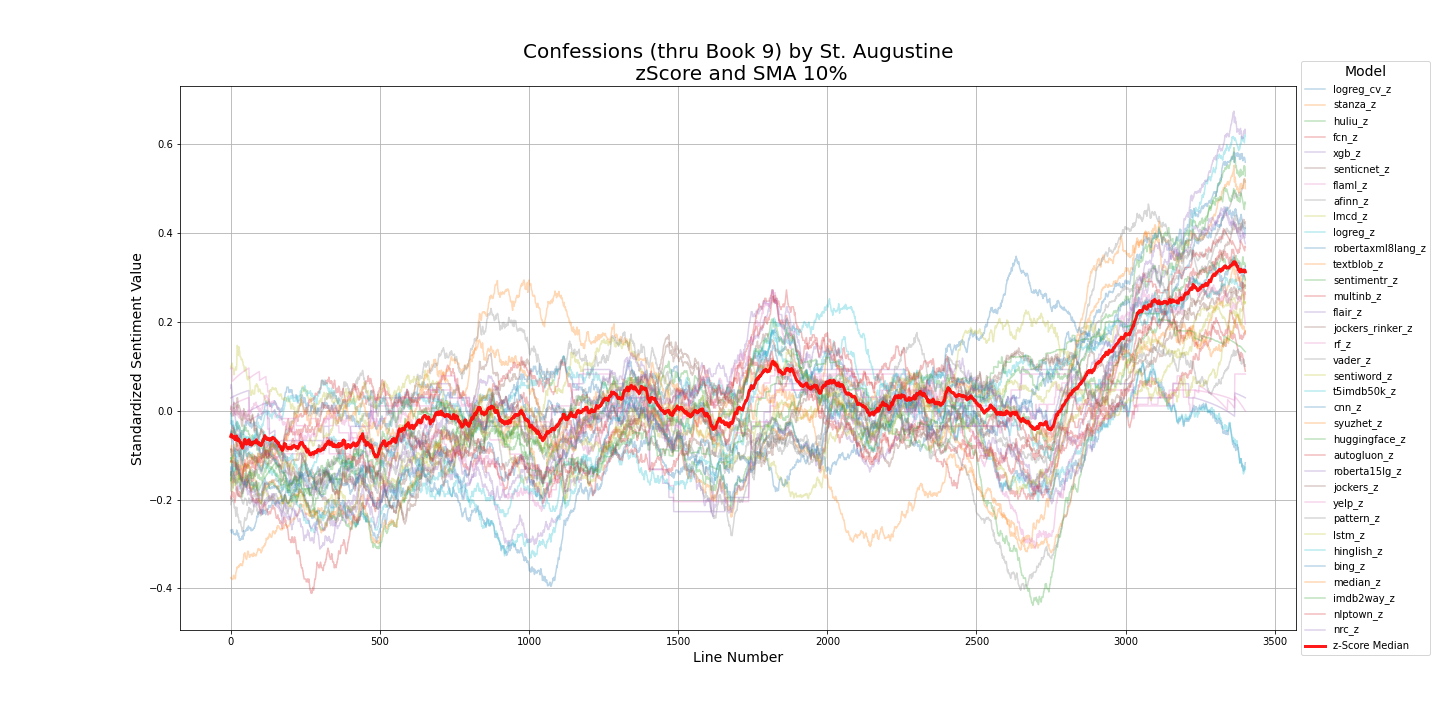}
\caption{ \textit{Confessions (thru Book  9)} by St. Augustine}
\label{appfig:metric_esp_staugustine_c}
\end{figure}

\begin{figure}[!ht]
\centering
\includegraphics[width=0.9\linewidth]{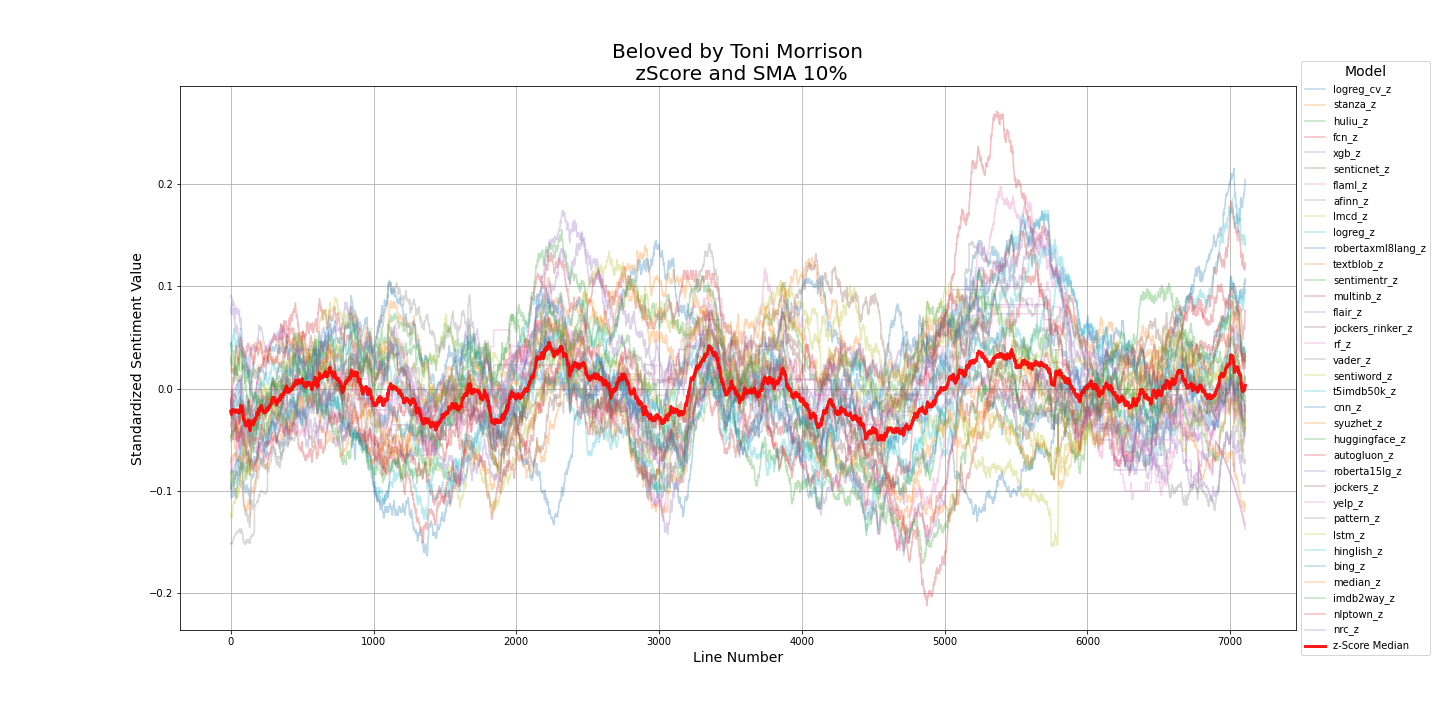}
\caption{ \textit{Beloved} by Toni Morrison}
\label{appfig:metric_esp_tmorrison_b}
\end{figure}

\begin{figure}[!ht]
\centering
\includegraphics[width=0.9\linewidth]{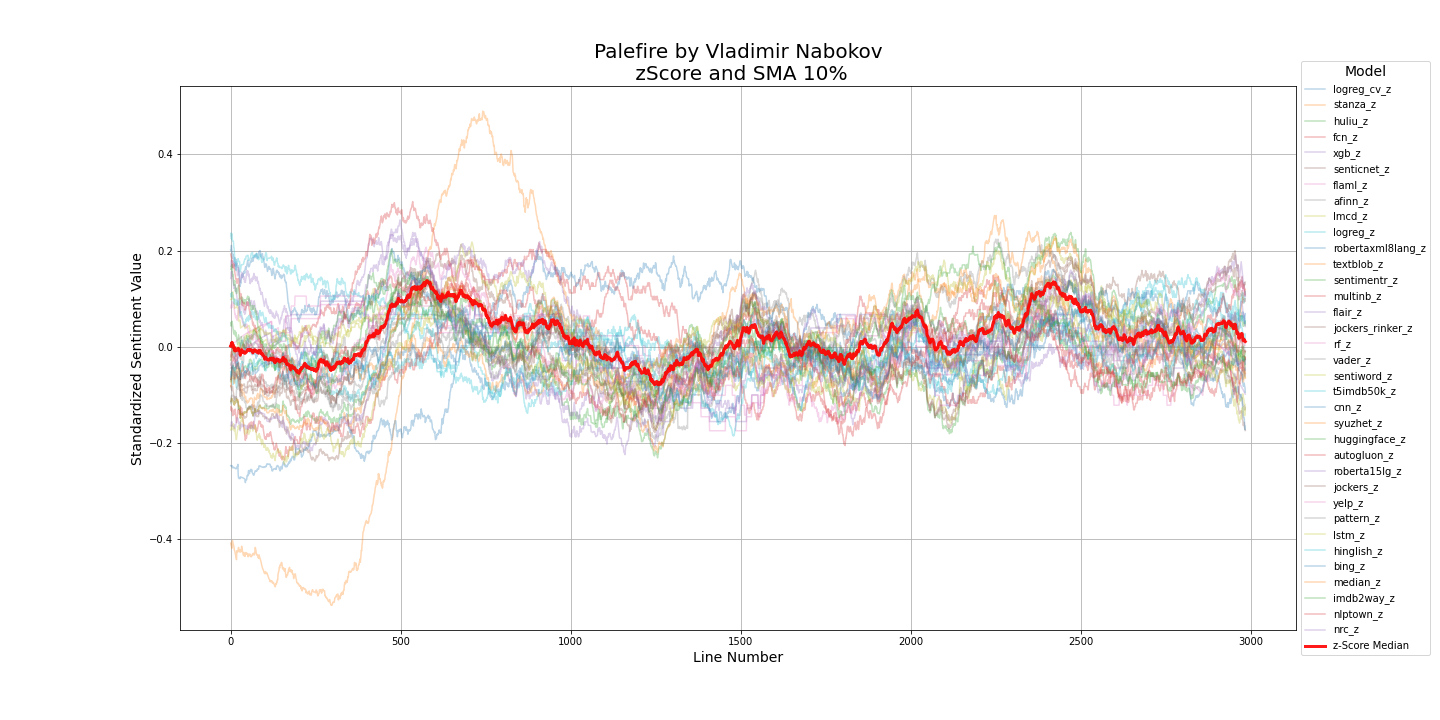}
\caption{ \textit{Pale Fire} by Vladimir Nabokov}
\label{appfig:metric_esp_vnabokov_pf}
\end{figure}

\begin{figure}[!ht]
\centering
\includegraphics[width=0.9\linewidth]{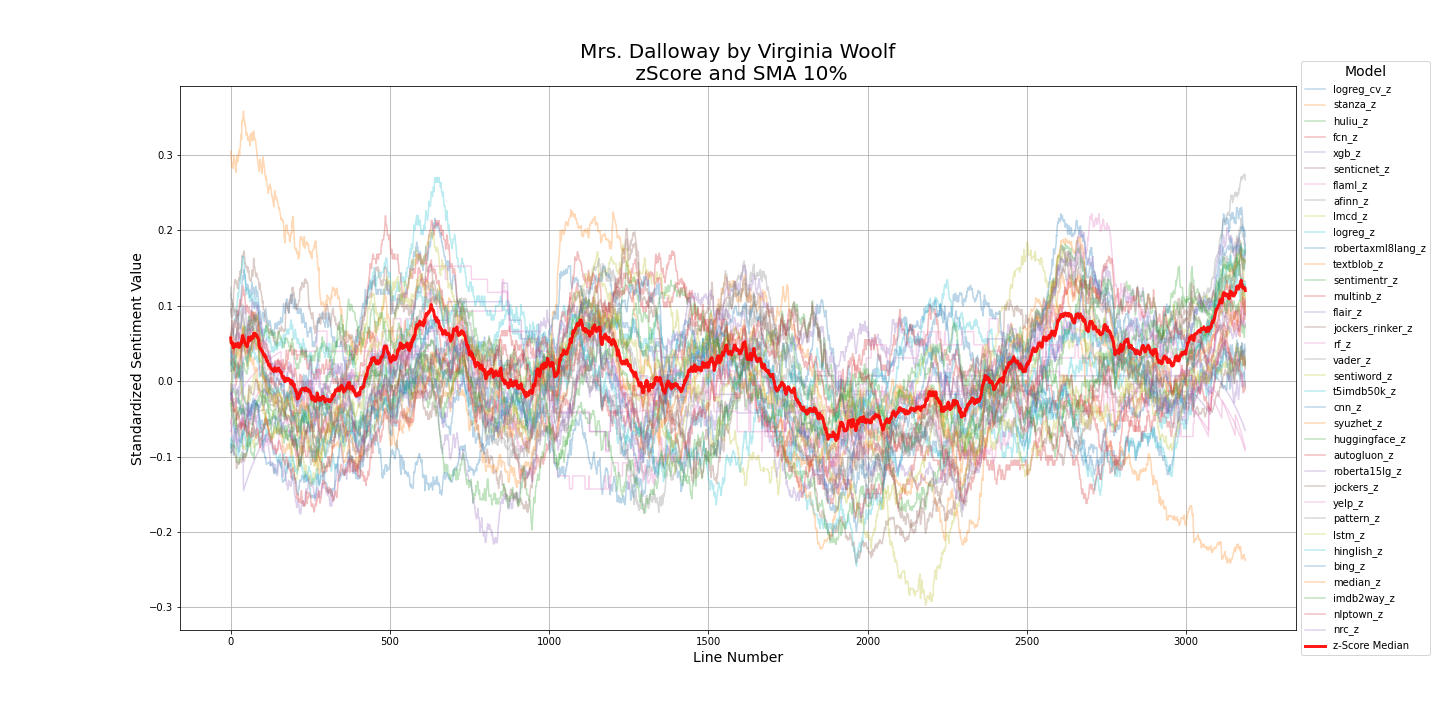}
\caption{ \textit{Mrs. Dalloway} by Virginia Woolf}
\label{appfig:metric_esp_vwoolf_md}
\end{figure}

\begin{figure}[!ht]
\centering
\includegraphics[width=0.9\linewidth]{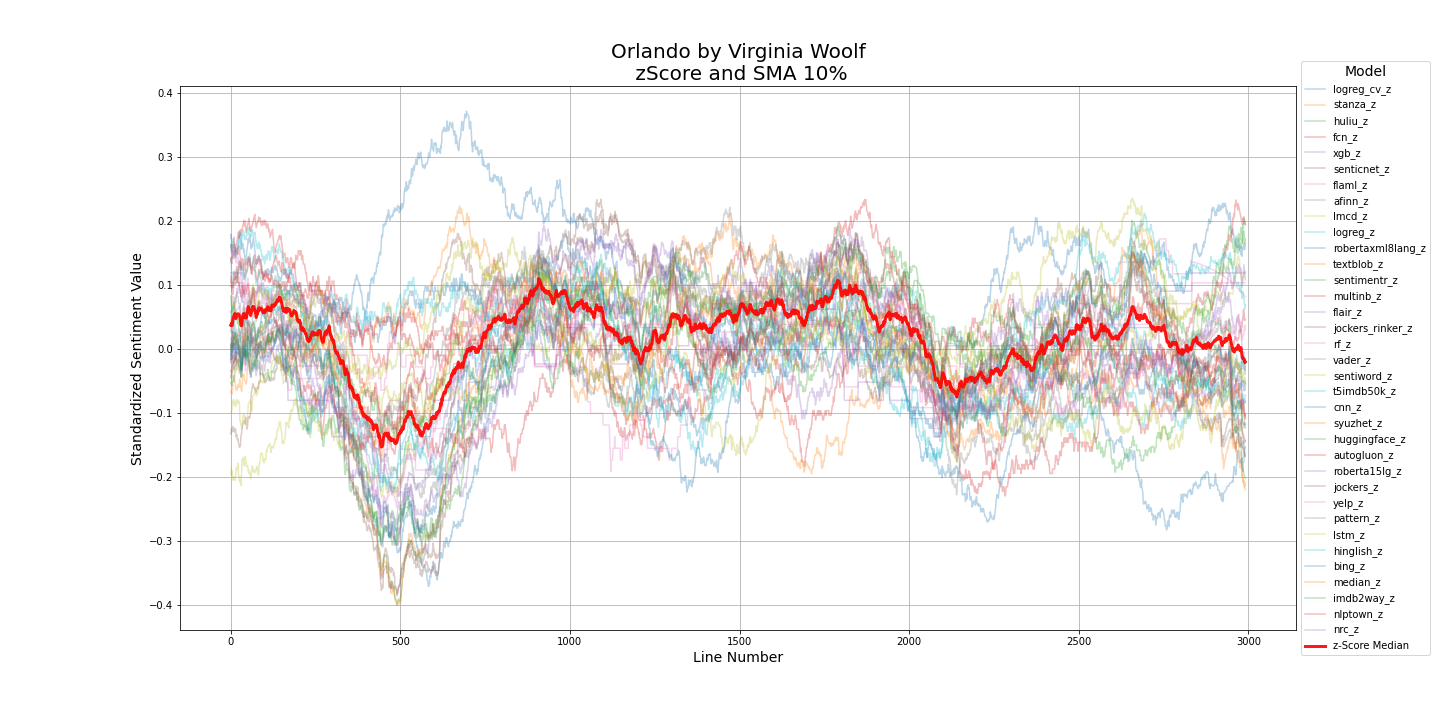}
\caption{ \textit{Orlando} by Virginia Woolf}
\label{appfig:metric_esp_vwoolf_o}
\end{figure}

\begin{figure}[!ht]
\centering
\includegraphics[width=0.9\linewidth]{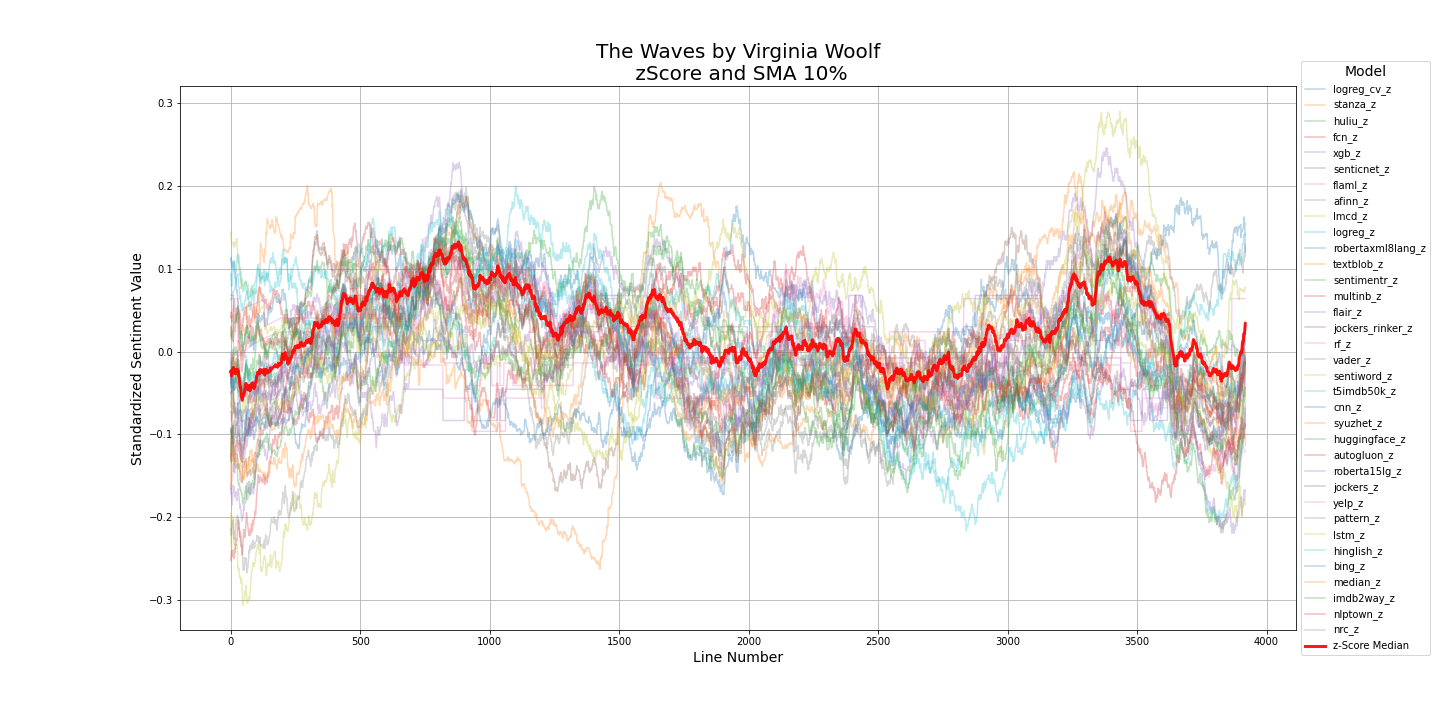}
\caption{ \textit{The Waves} by Virginia Woolf}
\label{appfig:metric_esp_vwoolf_tw}
\end{figure}

\begin{figure}[!ht]
\centering
\includegraphics[width=0.9\linewidth]{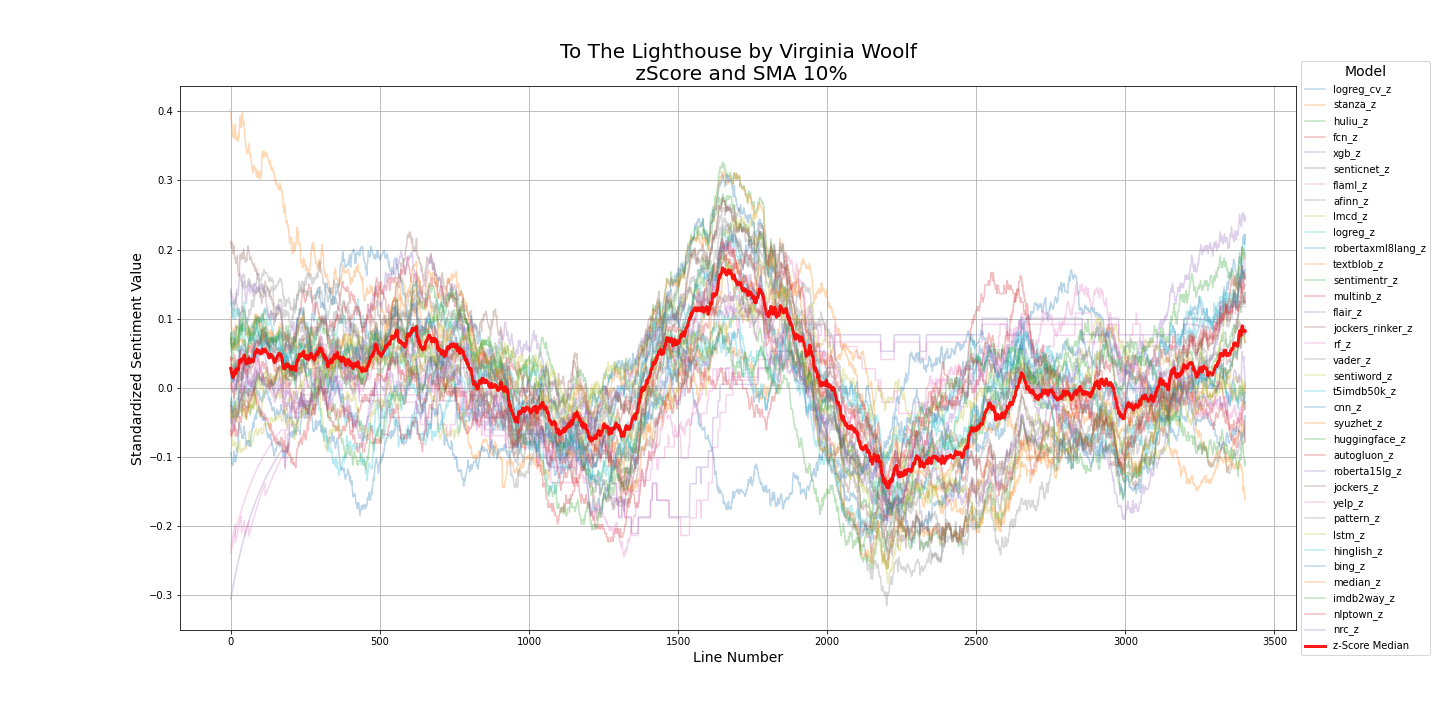}
\caption{ \textit{To The Lighthouse} by Virginia Woolf}
\label{appfig:metric_esp_vwoolf_ttl}
\end{figure}

\clearpage
\section{Appendix B. Model Corpus Compatibility}
All Model Sentiment Arcs by Corpus:
\label{app_b}

The following 25 plots rank how every model in the ensemble performed on each novel in the corpora. Larger values and longer bars correspond to better performance according to the MCC metric described in the Section \ref{sec:metrics_mcc} Model Corpus Compatibility. 

The following plots are presented in two groups based on how smoothly continuous MCC performance was across all models. A cluster suggests particular corpus:model combinations are collectively detecting unique features. The larger the cluster the more likely these features are to be meaningful. The smaller the cluster the more more likely these models are detecting features that are noise or anomalies. The same rule also applies to the degree with which the cluster's MCC metrics vary from surrounding models.

Fixed, fuzzy or dynamically tuned thresholds can automate this clustering process. However, the gold standard measure of significance ultimately depends upon collaborative interpretations between both ML and topic domain experts. Even in a fully automated form, clustering in these MCC plots provides guidance on how to further select, optimize and interpret key features of every corpus:model combination.

\begin{figure}[!ht]
\centering
\includegraphics[width=0.6\linewidth]{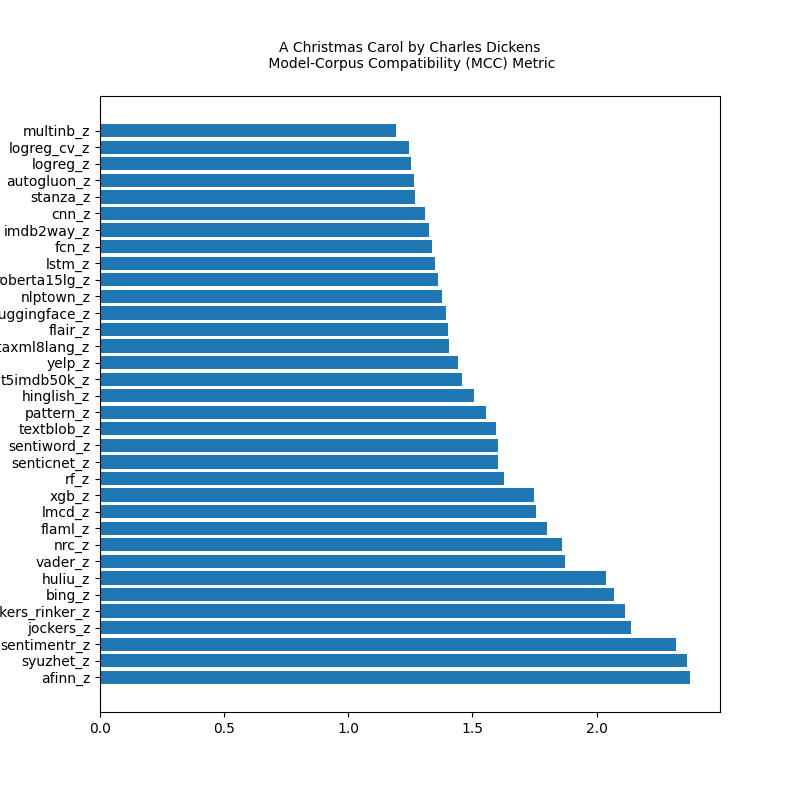}
\caption{ \textit{A Christmas Carol} by Charles Dickens}
\label{appfig:metric_mcc_cdickens_acc}
\end{figure}

\begin{figure}[!ht]
\centering
\includegraphics[width=0.6\linewidth]{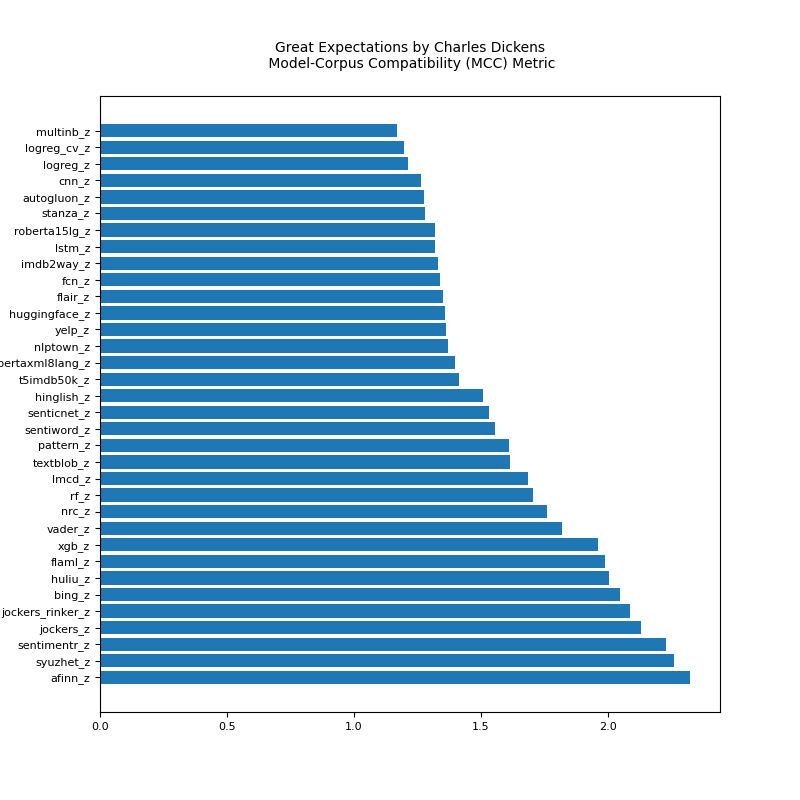}
\caption{ \textit{Great Expectations} by Charles Dickens}
\label{appfig:metric_mcc_cdickens_ge}
\end{figure}

\begin{figure}[!ht]
\centering
\includegraphics[width=0.6\linewidth]{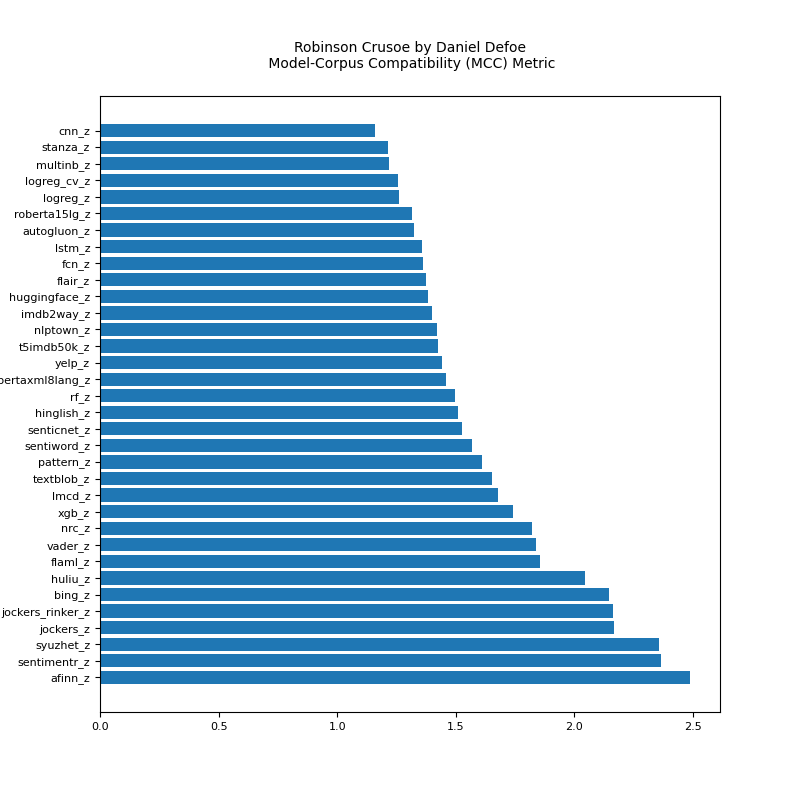}
\caption{ \textit{Robinson Crusoe} by Daniel Defoe}
\label{appfig:metric_mcc_ddefoe_rc}
\end{figure}

\begin{figure}[!ht]
\centering
\includegraphics[width=0.6\linewidth]{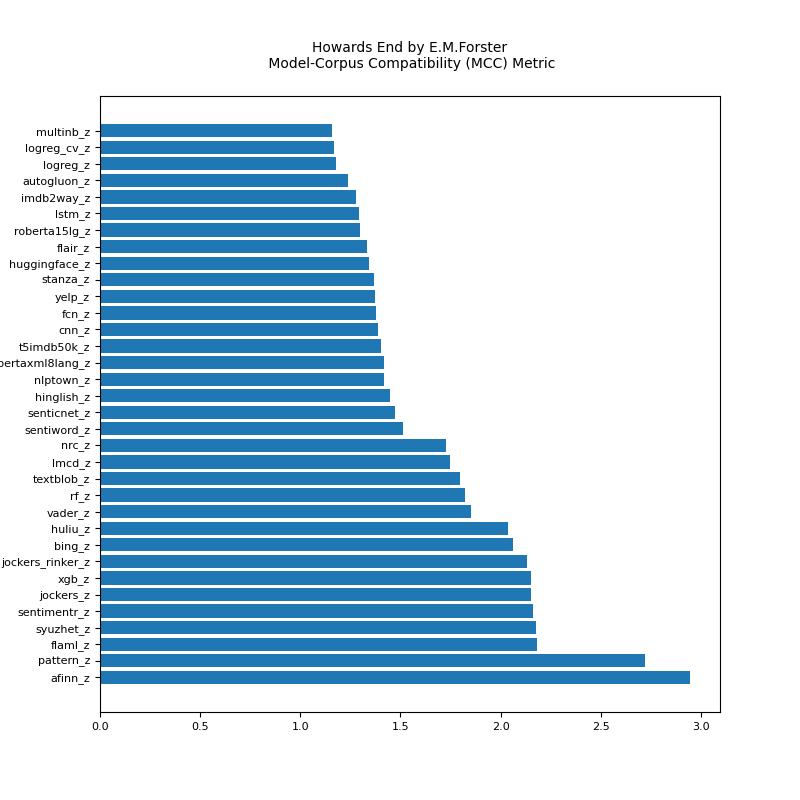}
\caption{ \textit{Howards End} by E. M. Forster}
\label{appfig:metric_mcc_emforster_he}
\end{figure}

\begin{figure}[!ht]
\centering
\includegraphics[width=0.6\linewidth]{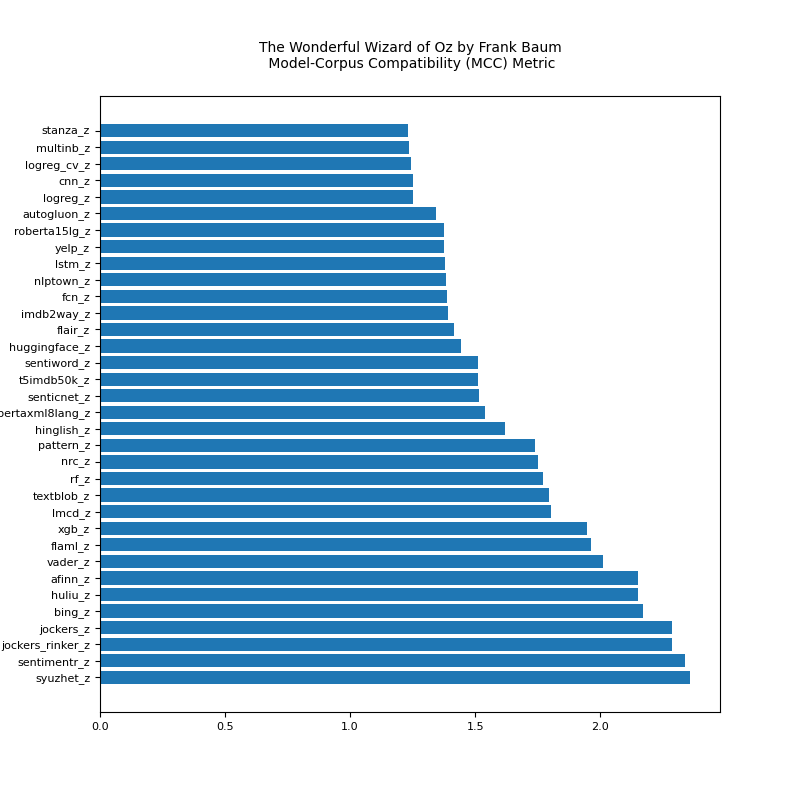}
\caption{ \textit{The Wonderful Wizard of Oz} by Frank Baum}
\label{appfig:metric_mcc_fbaum_twwoo}
\end{figure}

\begin{figure}[!ht]
\centering
\includegraphics[width=0.6\linewidth]{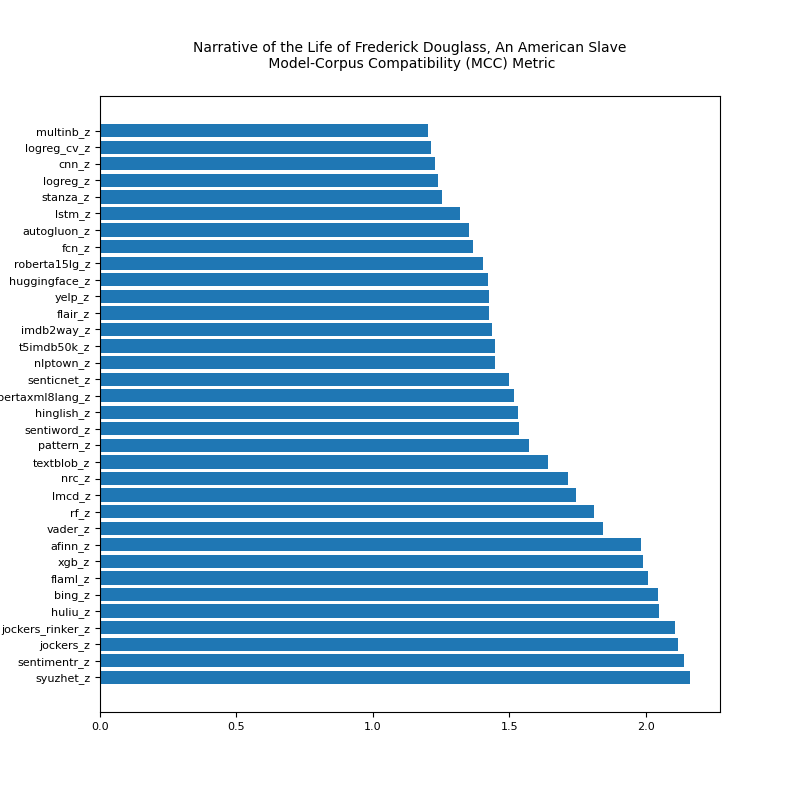}
\caption{ \textit{The Narrative of the Life of Frederick Douglass, an American Slave} by Frederick Douglass}
\label{appfig:metric_mcc_fdouglass_tnotlofdaas}
\end{figure}

\begin{figure}[!ht]
\centering
\includegraphics[width=0.6\linewidth]{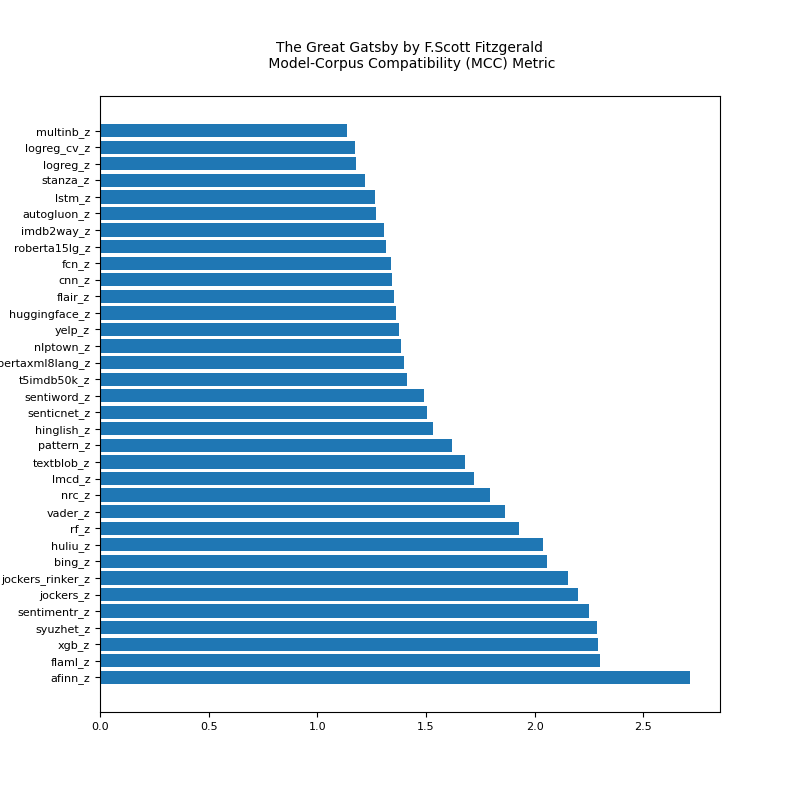}
\caption{ \textit{The Great Gatsby} by F. Scott Fitzgerald}
\label{appfig:metric_mcc_fsfitzgerald_gg}
\end{figure}

\begin{figure}[!ht]
\centering
\includegraphics[width=0.6\linewidth]{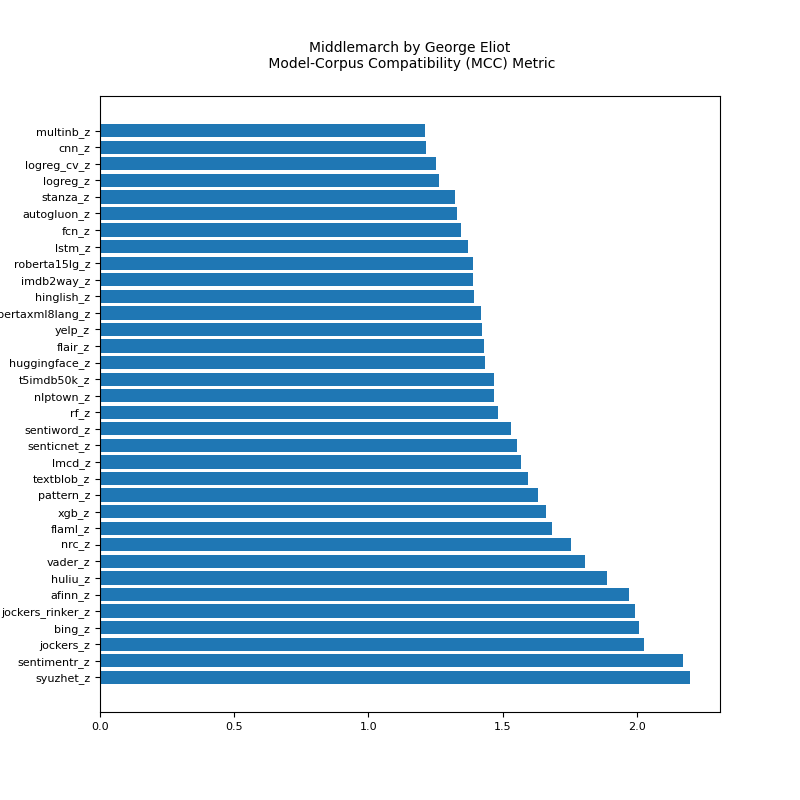}
\caption{ \textit{Middlemarch} by George Eliot}
\label{appfig:metric_mcc_geliot_mm}
\end{figure}

\begin{figure}[!ht]
\centering
\includegraphics[width=0.6\linewidth]{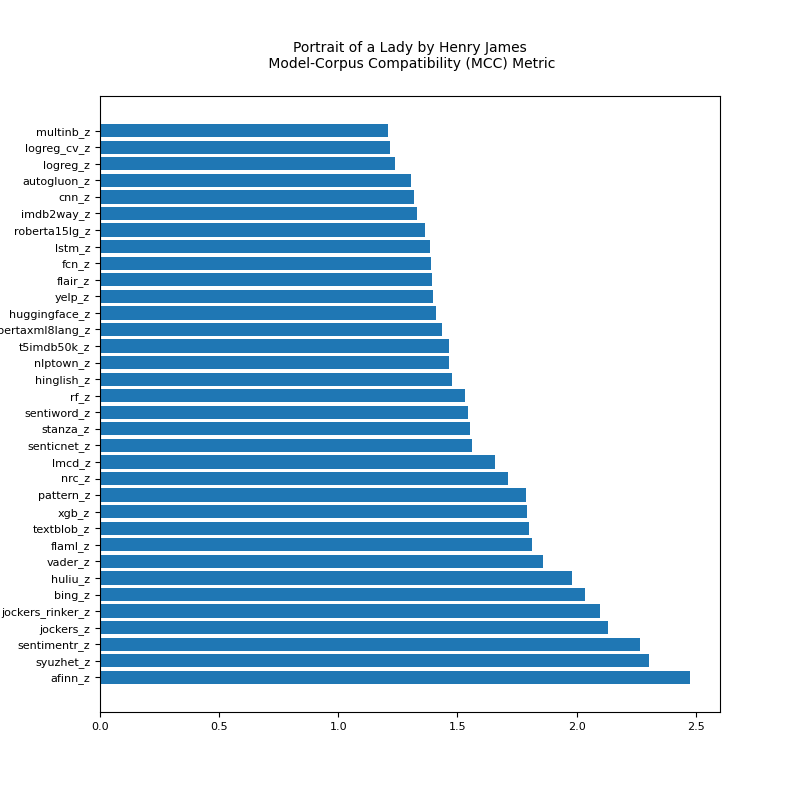}
\caption{ \textit{Portrait of a Lady} by Henry James}
\label{appfig:metric_mcc_hjames_poal}
\end{figure}

\begin{figure}[!ht]
\centering
\includegraphics[width=0.6\linewidth]{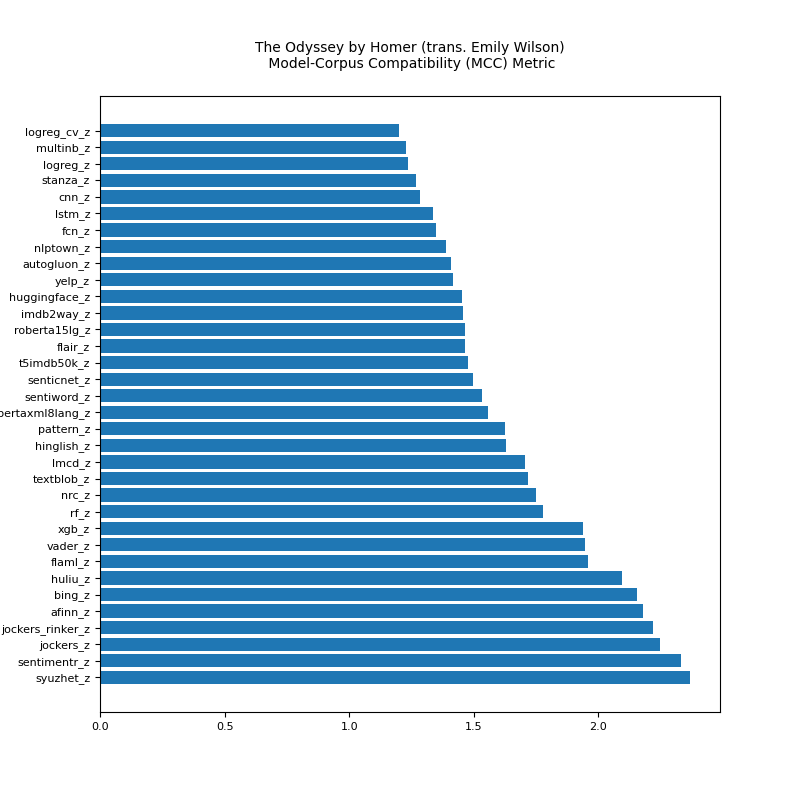}
\caption{ \textit{Odyssey} by Homer (trans. Emily Wilson) }
\label{appfig:metric_mcc_homerwilson_o}
\end{figure}

\begin{figure}[!ht]
\centering
\includegraphics[width=0.6\linewidth]{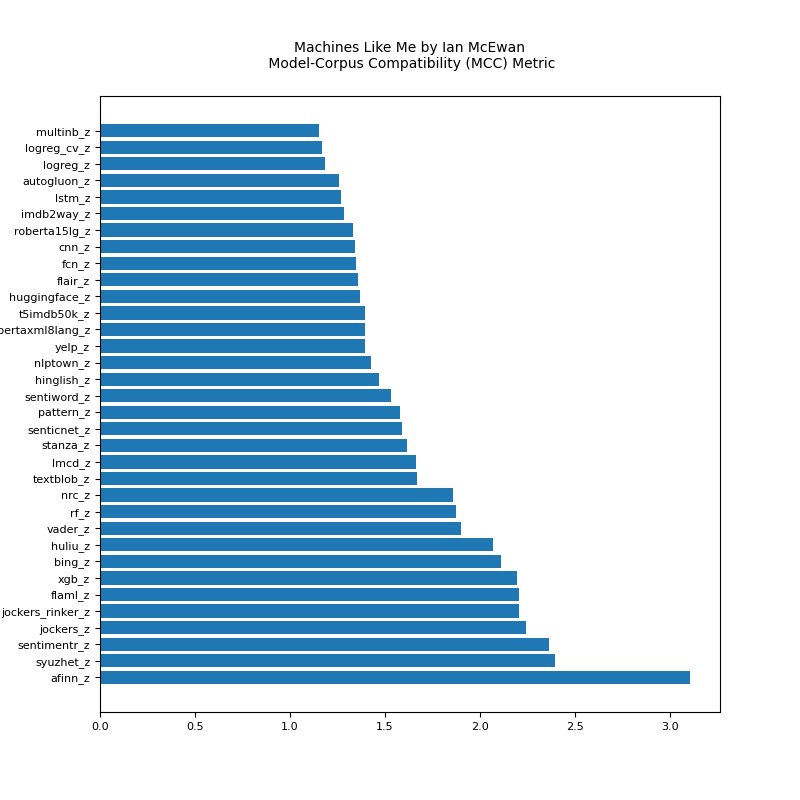}
\caption{ \textit{Machines Like Me} by Ian McEwan}
\label{appfig:metric_mcc_imcewan_mlm}
\end{figure}

\begin{figure}[!ht]
\centering
\includegraphics[width=0.6\linewidth]{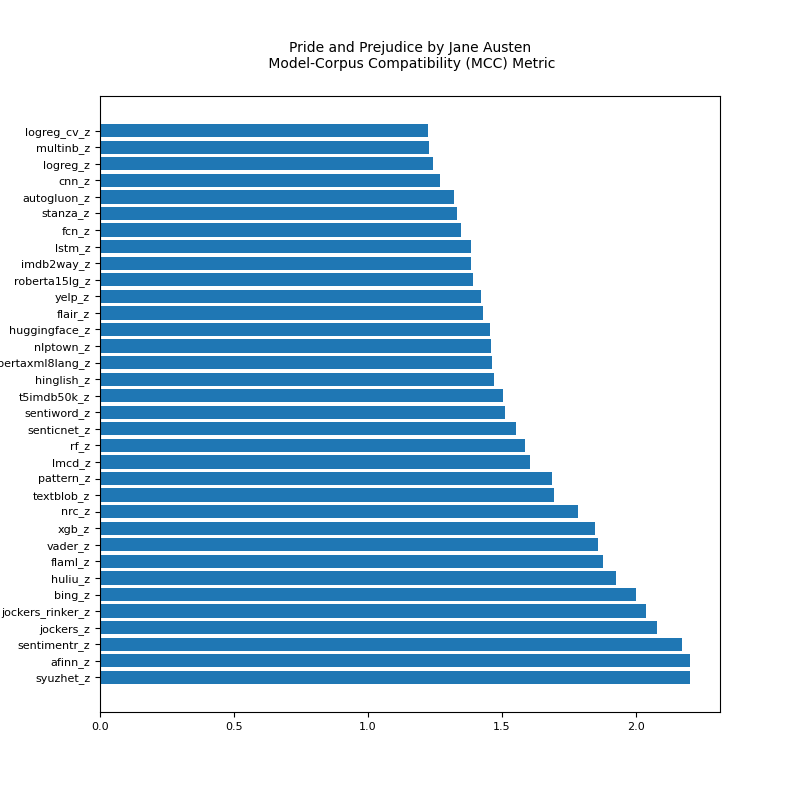}
\caption{ \textit{Pride and Prejudice} by Jane Austen }
\label{appfig:metric_mcc_jausten_pap}
\end{figure}

\begin{figure}[!ht]
\centering
\includegraphics[width=0.6\linewidth]{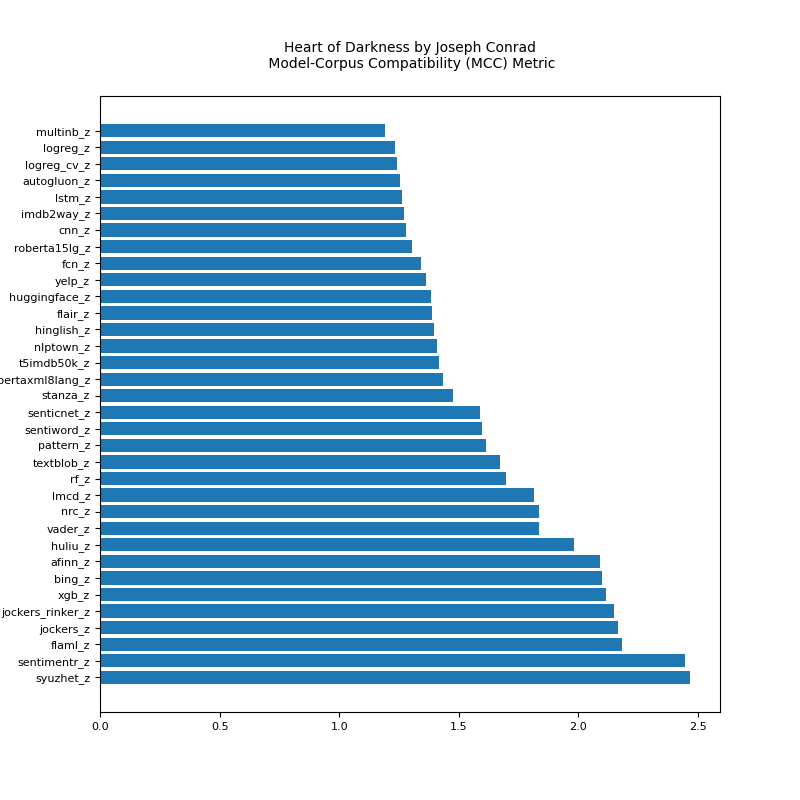}
\caption{ \textit{Heart of Darkness} by Joseph Conrad}
\label{appfig:metric_mcc_jconrad_hod}
\end{figure}

\begin{figure}[!ht]
\centering
\includegraphics[width=0.6\linewidth]{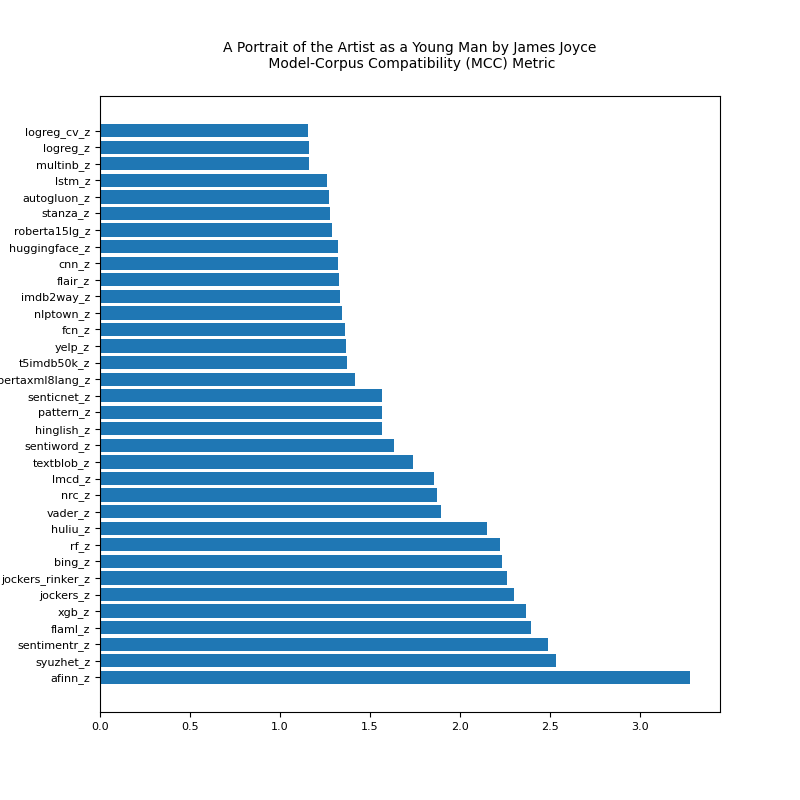}
\caption{ \textit{A Portrait of the Artists as a Young Man} by James Joyce }
\label{appfig:metric_mcc_jjoyce_apotaaaym}
\end{figure}

\begin{figure}[!ht]
\centering
\includegraphics[width=0.6\linewidth]{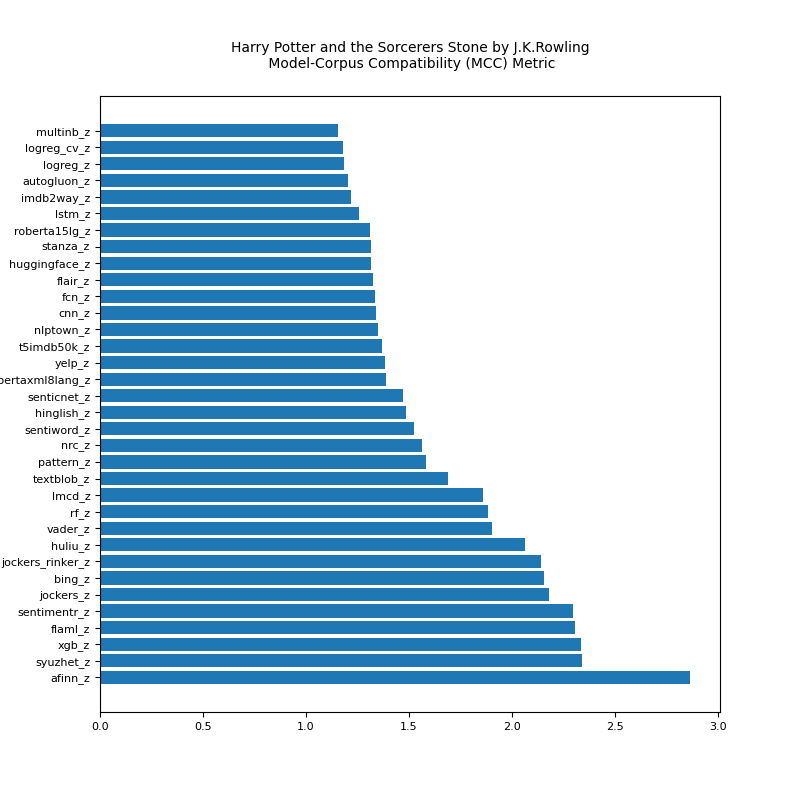}
\caption{ \textit{Harry Potter and the Sorcerer's Stone} by J. K. Rowling}
\label{appfig:metric_mcc_jkrowling_hpatss}
\end{figure}

\begin{figure}[!ht]
\centering
\includegraphics[width=0.6\linewidth]{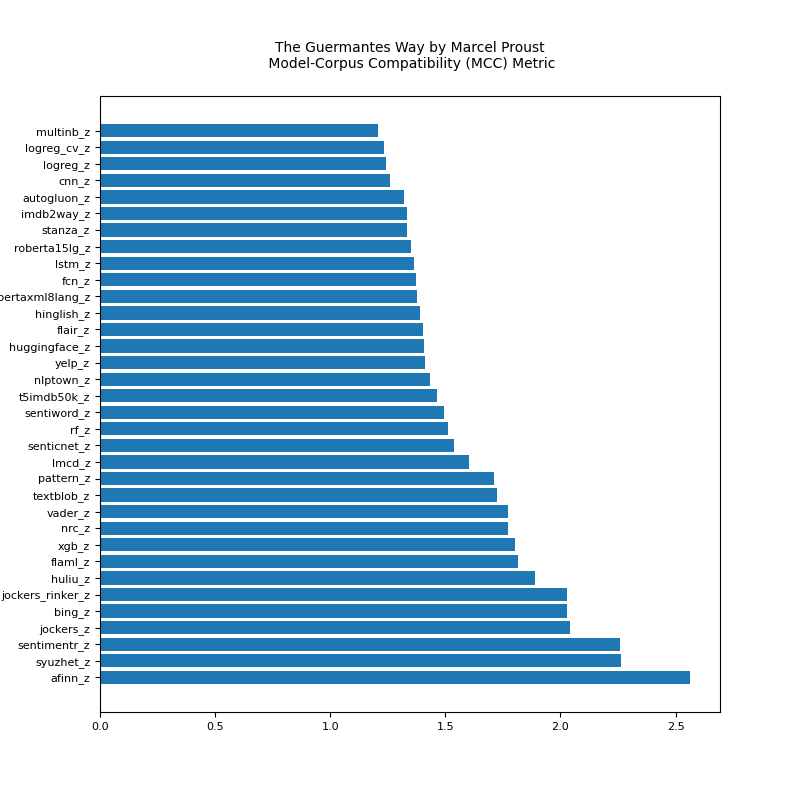}
\caption{ \textit{The Guermantes Way} by Marcel Proust (trans. Mark Treharne) }
\label{appfig:metric_mcc_mprousttreharne_tgw}
\end{figure}

\begin{figure}[!ht]
\centering
\includegraphics[width=0.6\linewidth]{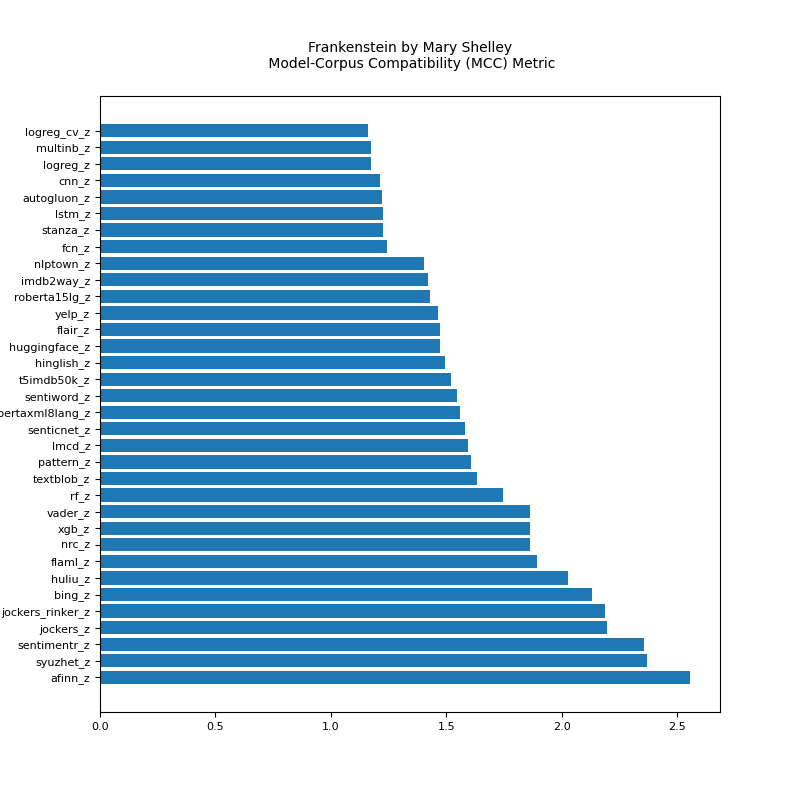}
\caption{ \textit{Frankenstein} by Mary Shelley}
\label{appfig:metric_mcc_mshelley_f}
\end{figure}

\begin{figure}[!ht]
\centering
\includegraphics[width=0.6\linewidth]{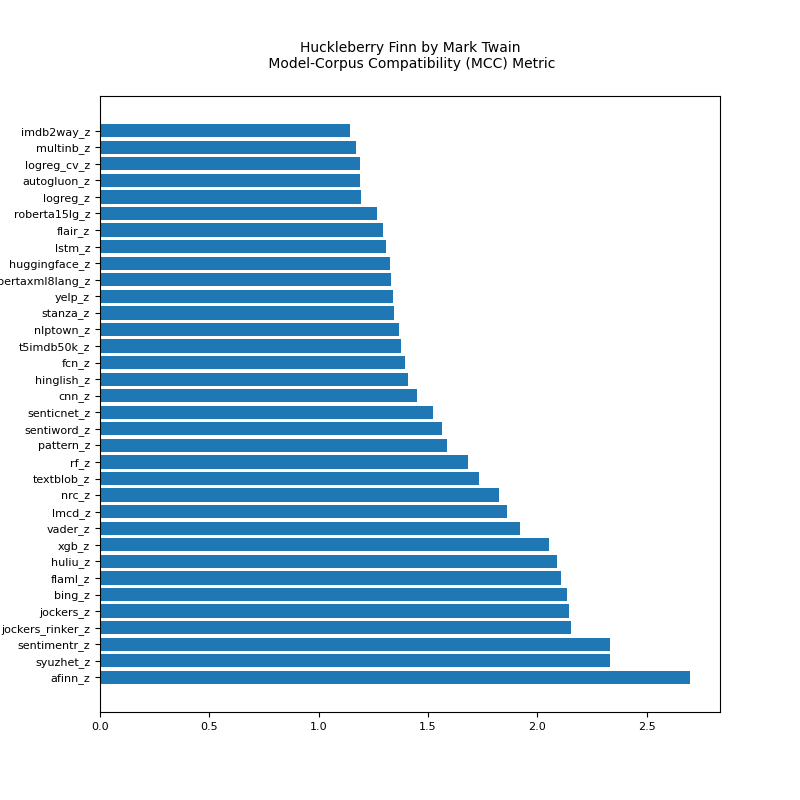}
\caption{ \textit{Huckleberry Finn} by Mark Twain}
\label{appfig:metric_mcc_mtwain_hf}
\end{figure}

\begin{figure}[!ht]
\centering
\includegraphics[width=0.6\linewidth]{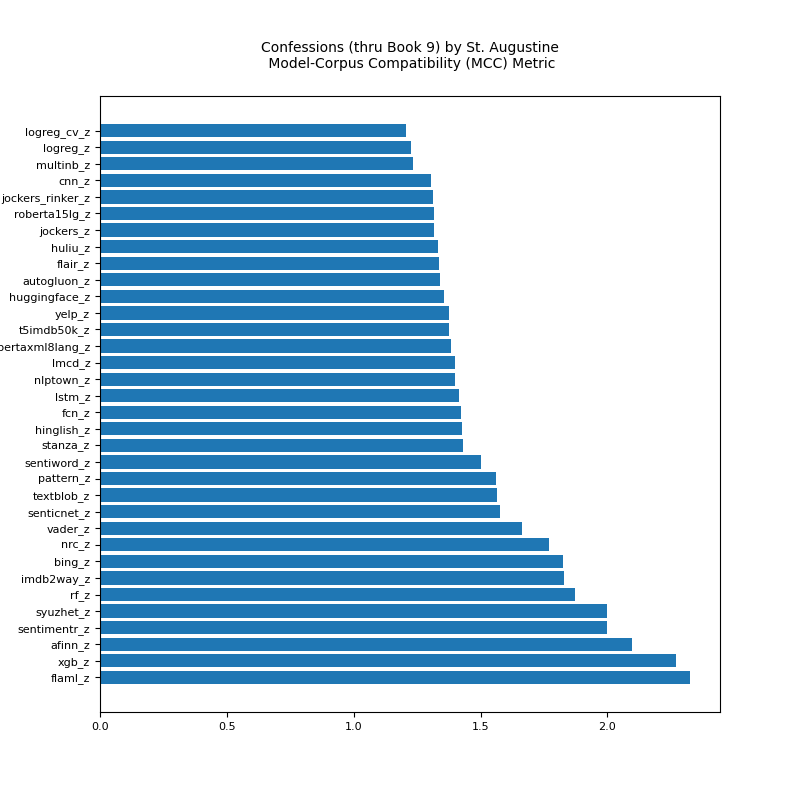}
\caption{ \textit{Confessions (thru Book  9)} by St. Augustine}
\label{appfig:metric_mcc_staugustine_c}
\end{figure}

\begin{figure}[!ht]
\centering
\includegraphics[width=0.6\linewidth]{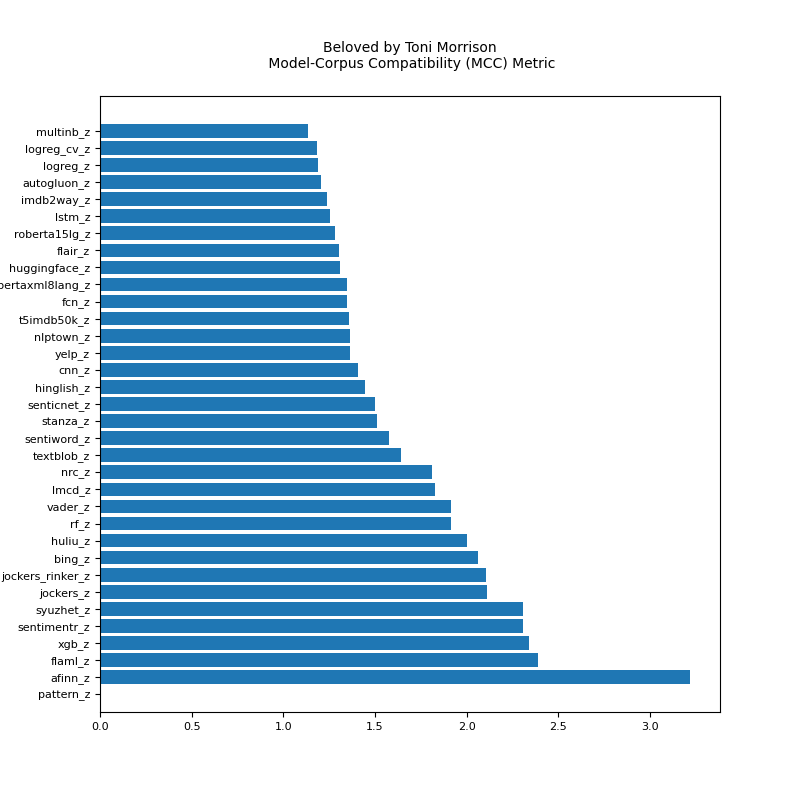}
\caption{ \textit{Beloved} by Toni Morrison}
\label{appfig:metric_mcc_tmorrison_b}
\end{figure}

\begin{figure}[!ht]
\centering
\includegraphics[width=0.6\linewidth]{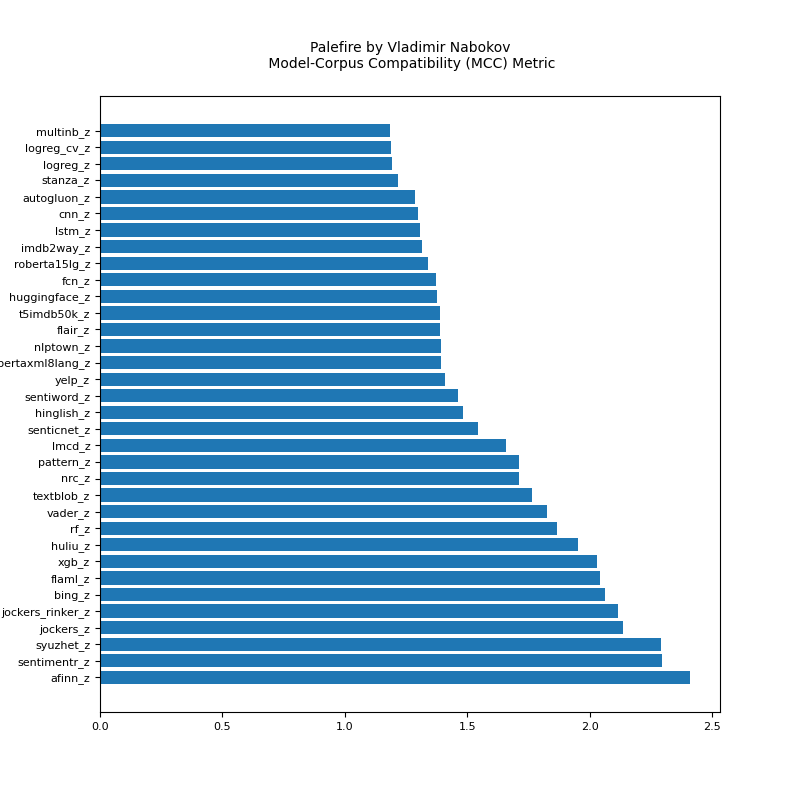}
\caption{ \textit{Pale Fire} by Vladimir Nabokov}
\label{appfig:metric_mcc_vnabokov_pf}
\end{figure}

\begin{figure}[!ht]
\centering
\includegraphics[width=0.6\linewidth]{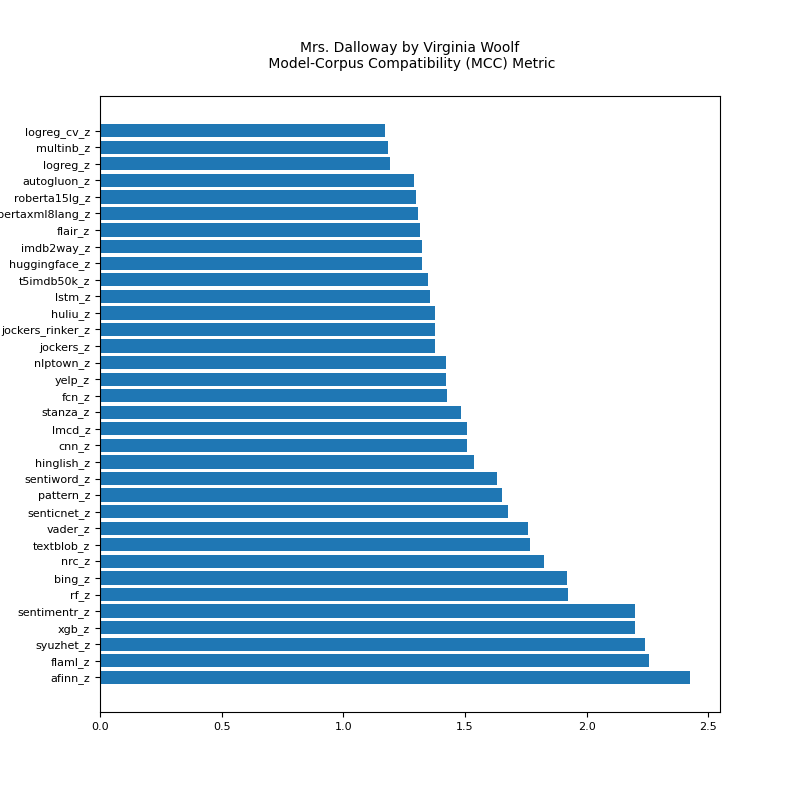}
\caption{ \textit{Mrs. Dalloway} by Virginia Woolf}
\label{appfig:metric_mcc_vwoolf_md}
\end{figure}

\begin{figure}[!ht]
\centering
\includegraphics[width=0.6\linewidth]{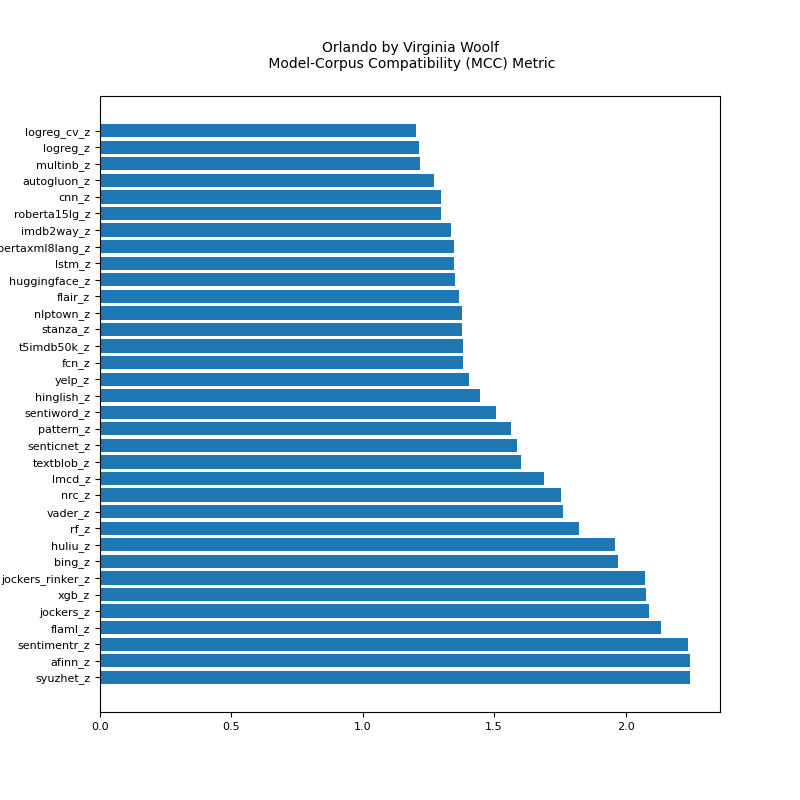}
\caption{ \textit{Orlando} by Virginia Woolf}
\label{appfig:metric_mcc_vwoolf_o}
\end{figure}

\begin{figure}[!ht]
\centering
\includegraphics[width=0.6\linewidth]{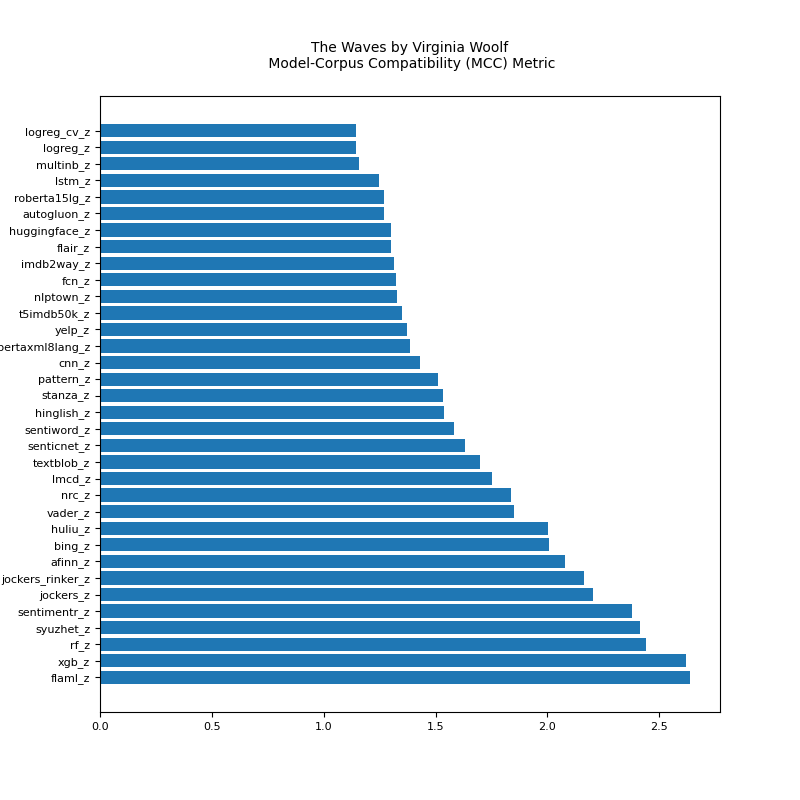}
\caption{ \textit{The Waves} by Virginia Woolf}
\label{appfig:metric_mcc_vwoolf_tw}
\end{figure}

\begin{figure}[!ht]
\centering
\includegraphics[width=0.6\linewidth]{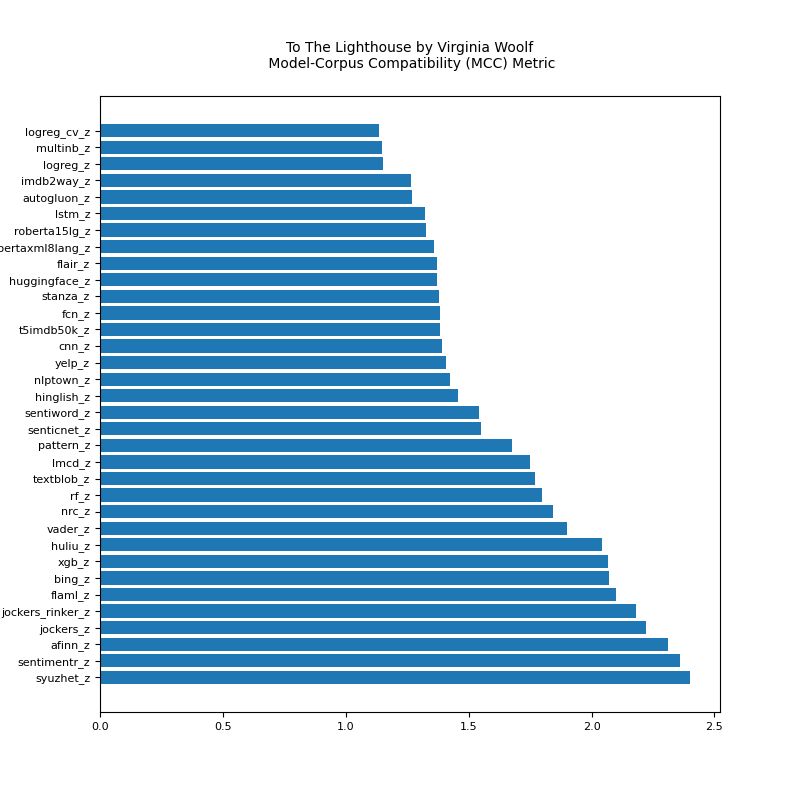}
\caption{ \textit{To The Lighthouse} by Virginia Woolf}
\label{appfig:metric_mcc_vwoolf_ttl}
\end{figure}

\clearpage
\section{Appendix C. Agglomerative Hierarchical Clustering of Sentiment Arcs}
\label{app_c}

For each novel in the reference corpora, an agglomerative hierarchical dendrogram shows the similarity between sentiment arcs generated by each model in the ensemble. As described in Section \ref{sec:methodology} Methodology, these sentiment arcs are created from a time series processing pipeline with the steps: (a) conversion to a floating point number where possible/necessary, (b) standardization with z-scores, (c) smoothing with 10\% simple moving averages and (c) reducing and normalizing dimensions down to 25 data points with the LTTB algorithm. LTTB downsampling creates irregularly spaced time series while preserving the key features. Dynamic Type Warping generates the distance matrix used for creating the agglomerative clustering dendrograms.

The clustering in these dendrograms shows how similar models perform on a particular corpus in terms of the DTW distances between sentiment arcs. Simple models (e.g. lexical and heuristic) tend to cluster together while more advanced models (e.g. Transformers, DNNs and AutoML variants) are more randomly distributed in the dendrogram leaves.

\begin{figure}[!ht]
\centering
\includegraphics[width=0.9\linewidth]{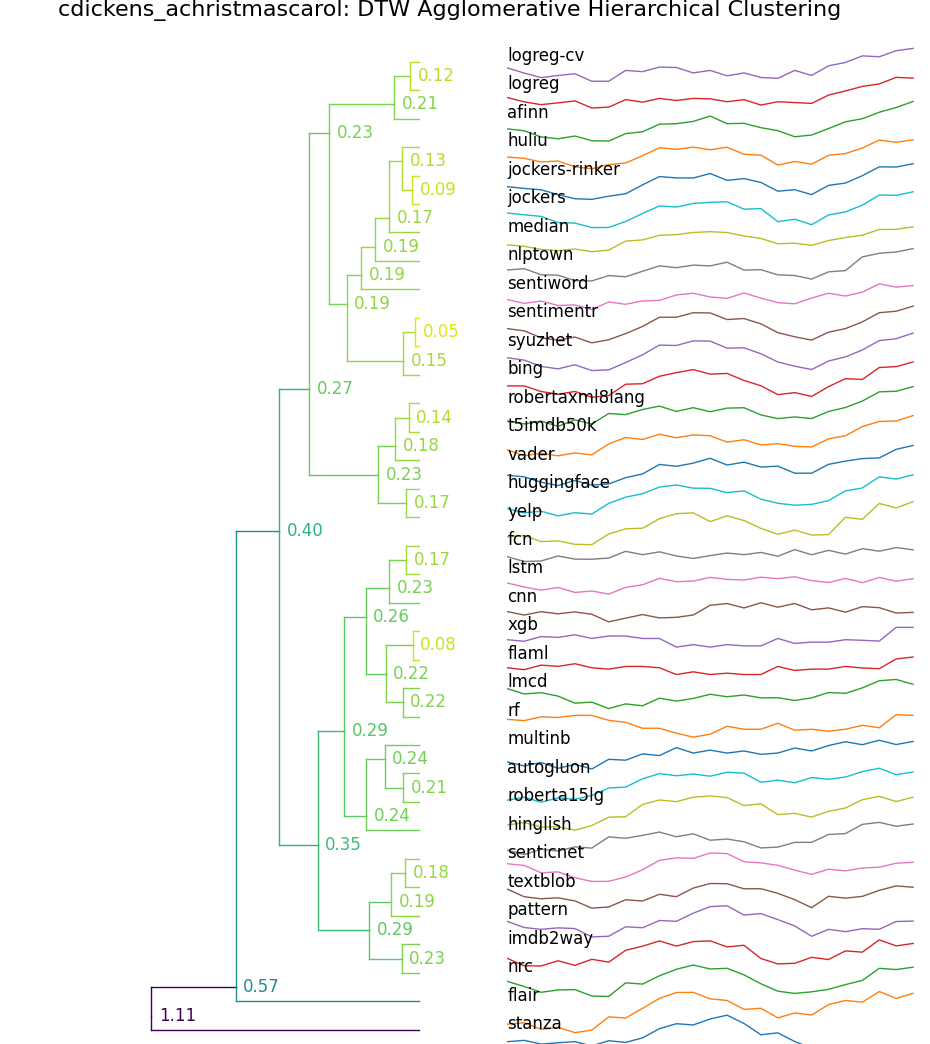}
\caption{\textit{A Christmas Carol} by Charles Dickens}
\label{appfig:metric_hcd_cdickens_acc}
\end{figure}

\begin{figure}[!ht]
\centering
\includegraphics[width=0.9\linewidth]{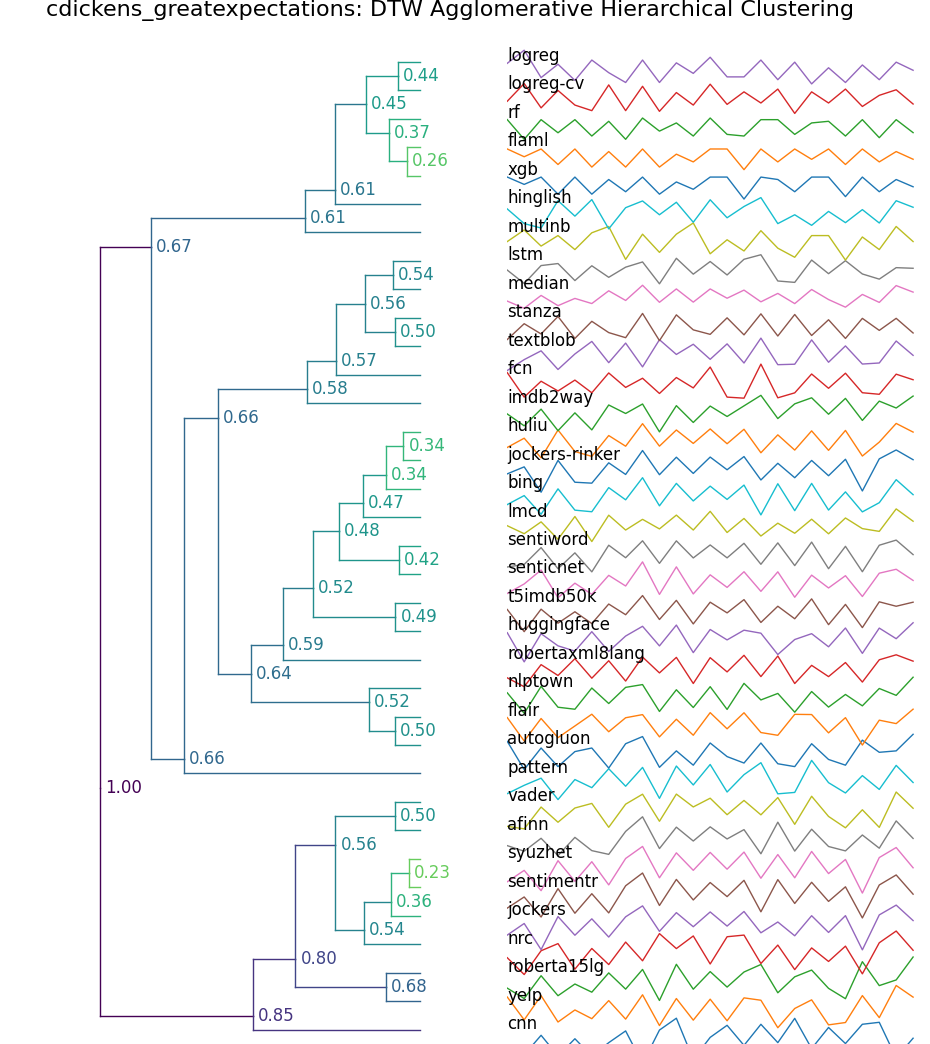}
\caption{ \textit{Great Expectations} by Charles Dickens}
\label{appfig:metric_hcd_cdickens_ge}
\end{figure}

\begin{figure}[!ht]
\centering
\includegraphics[width=0.9\linewidth]{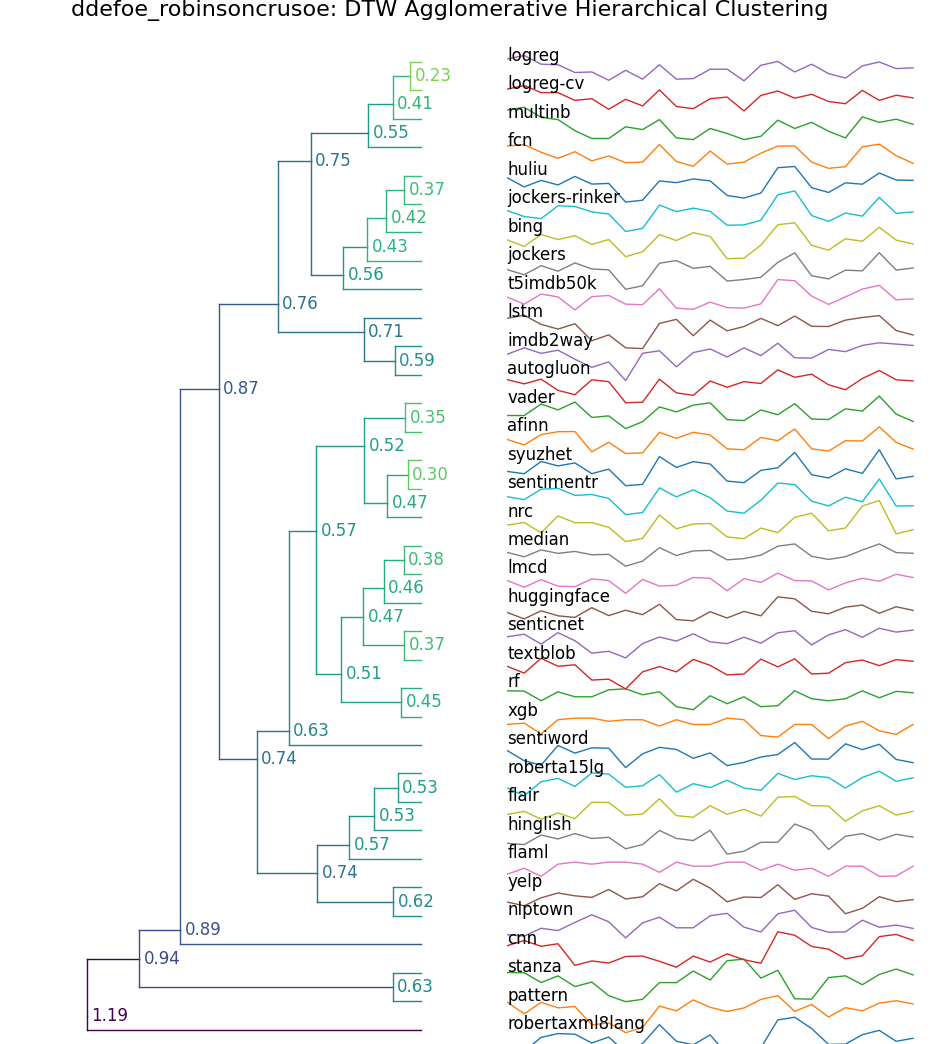}
\caption{ \textit{Robinson Crusoe} by Daniel Defoe}
\label{appfig:metric_hcd_ddefoe_rc}
\end{figure}

\begin{figure}[!ht]
\centering
\includegraphics[width=0.9\linewidth]{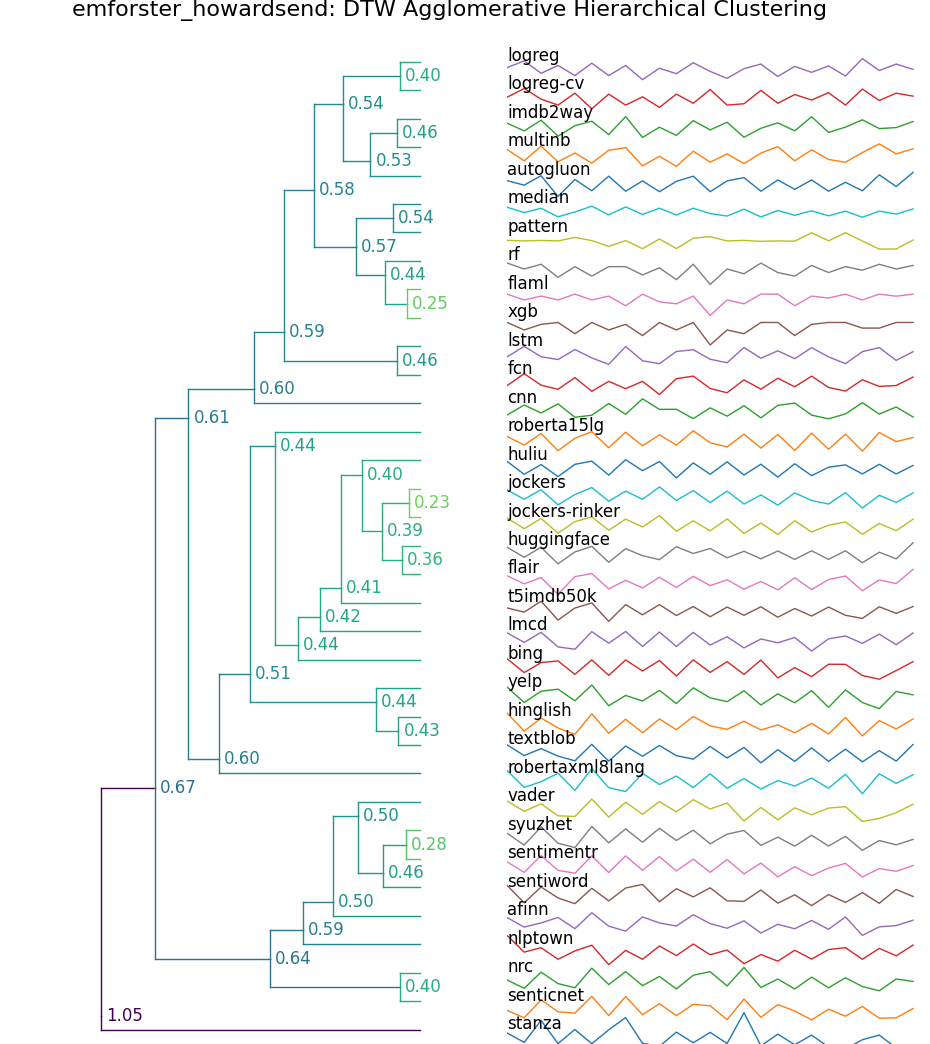}
\caption{ \textit{Howards End} by E. M. Forster}
\label{appfig:metric_hcd_emforster_he}
\end{figure}

\begin{figure}[!ht]
\centering
\includegraphics[width=0.9\linewidth]{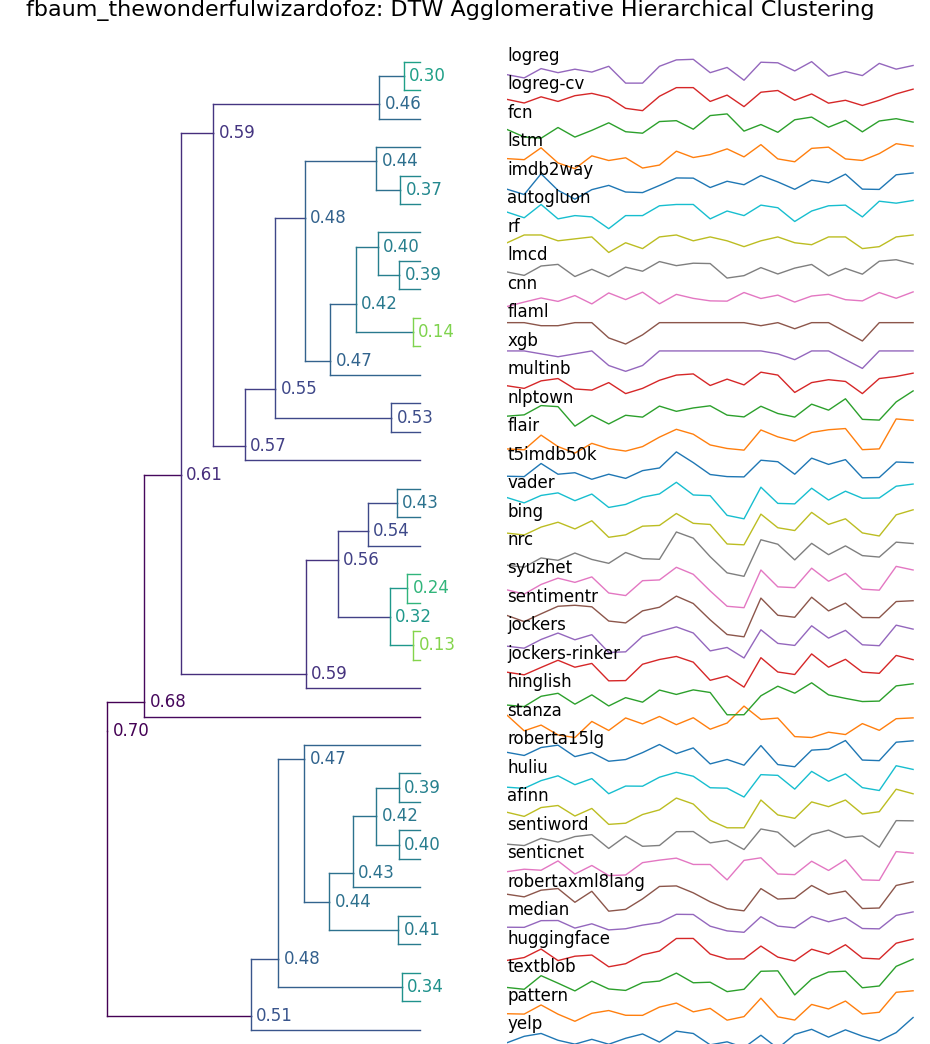}
\caption{ \textit{The Wonderful Wizard of Oz} by Frank Baum}
\label{appfig:metric_hcd_fbaum_twwoo}
\end{figure}

\begin{figure}[!ht]
\centering
\includegraphics[width=0.9\linewidth]{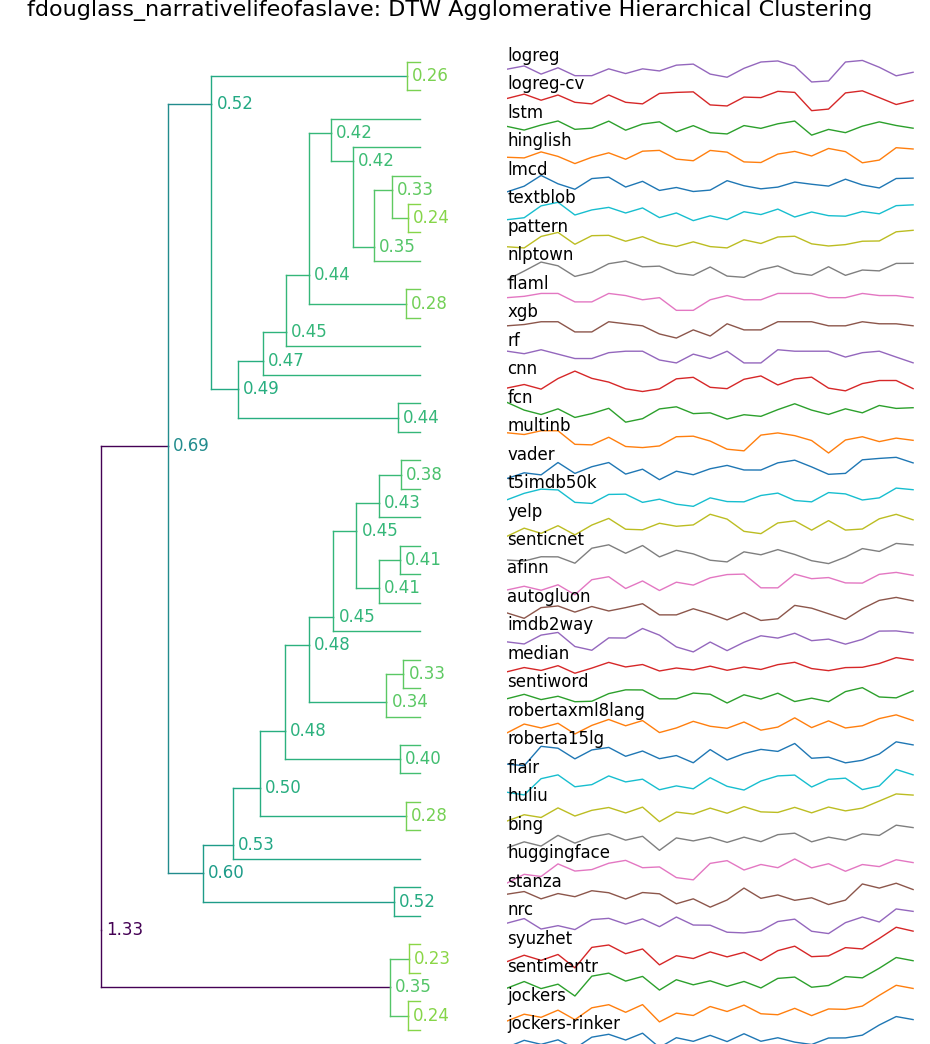}
\caption{ \textit{The Narrative of the Life of Frederick Douglass, an American Slave} by Frederick Douglass}
\label{appfig:metric_hcd_fdouglass_tnotlofdaas}
\end{figure}

\begin{figure}[!ht]
\centering
\includegraphics[width=0.9\linewidth]{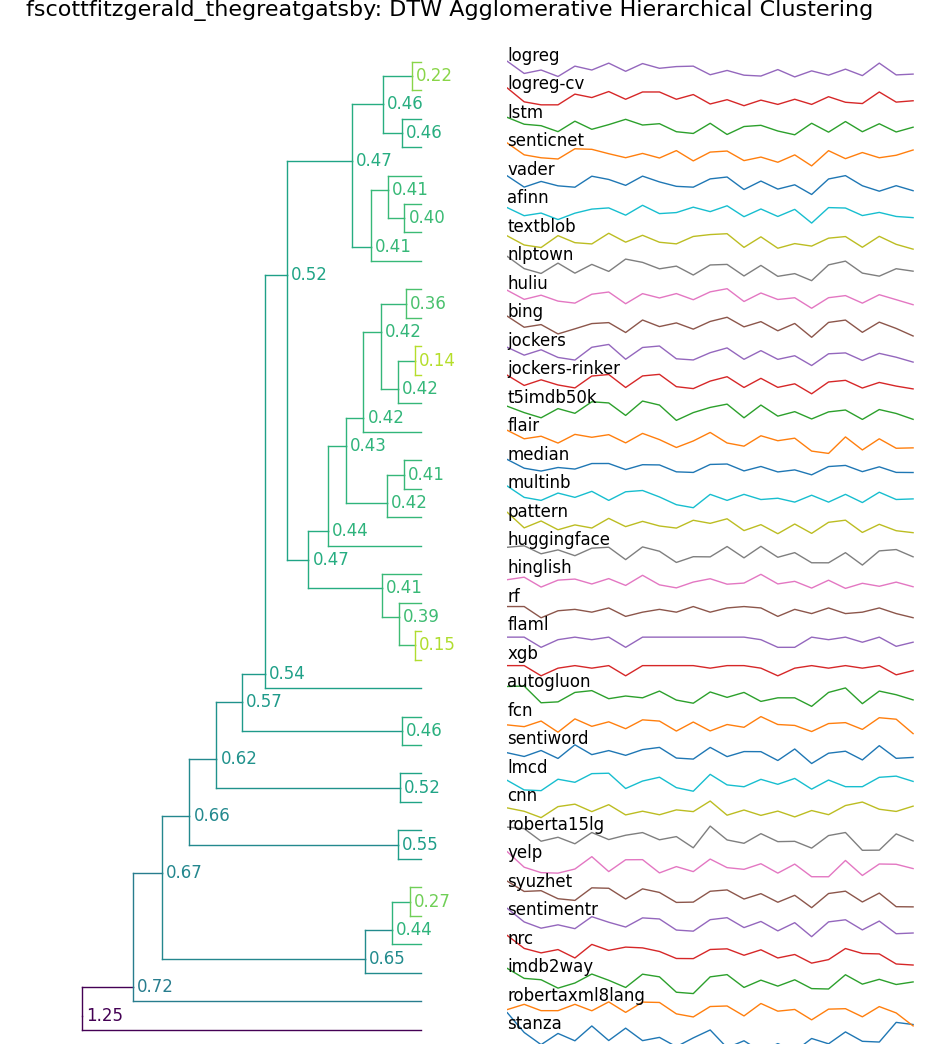}
\caption{ \textit{The Great Gatsby} by F. Scott Fitzgerald}
\label{appfig:metric_hcd_fsfitzgerald_gg}
\end{figure}

\begin{figure}[!ht]
\centering
\includegraphics[width=0.9\linewidth]{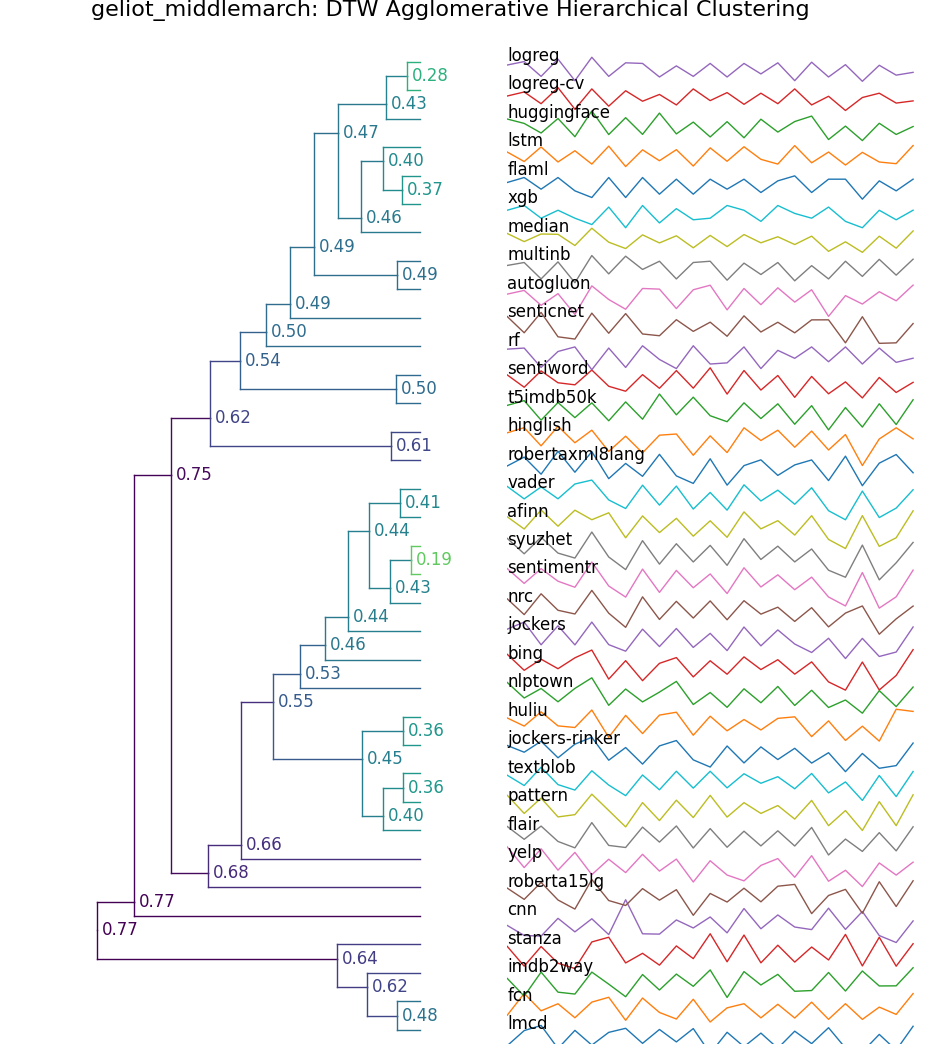}
\caption{ \textit{Middlemarch} by George Eliot}
\label{appfig:metric_hcd_geliot_mm}
\end{figure}

\begin{figure}[!ht]
\centering
\includegraphics[width=0.9\linewidth]{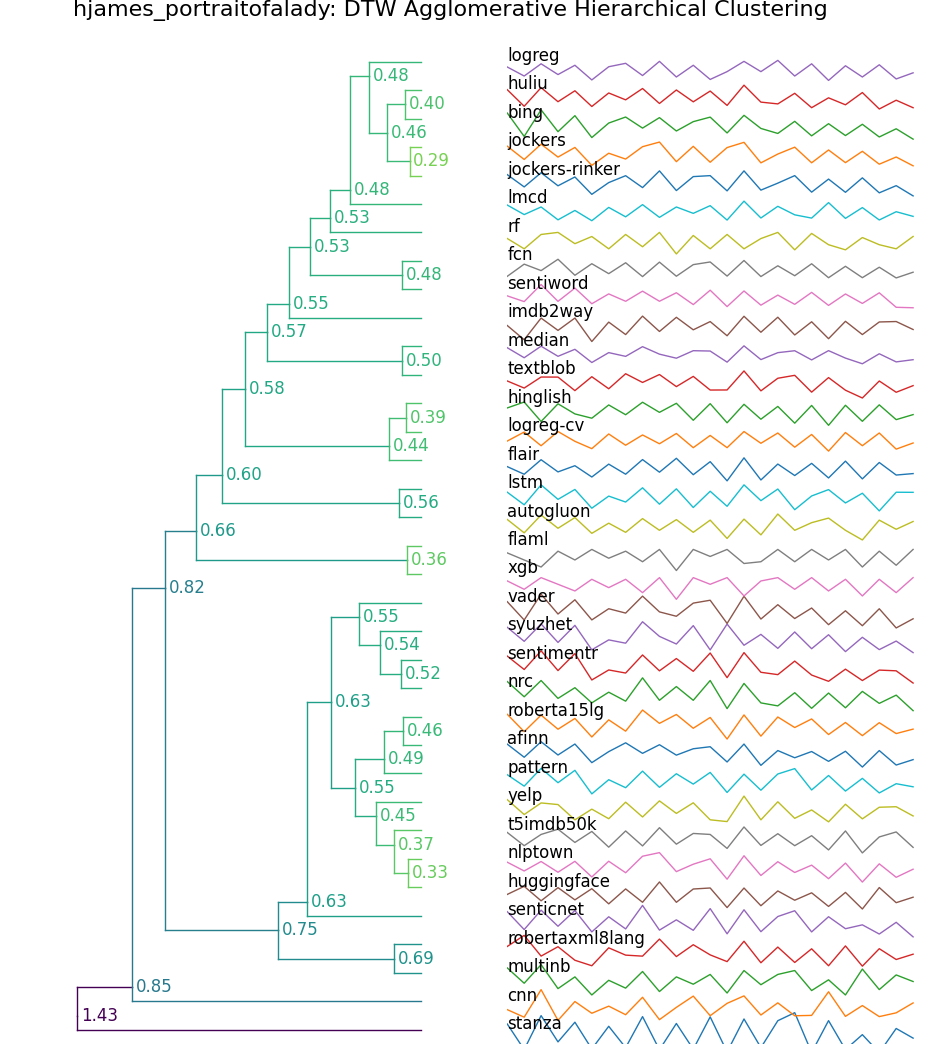}
\caption{ \textit{Portrait of a Lady} by Henry James}
\label{appfig:metric_hcd_hjames_poal}
\end{figure}

\begin{figure}[!ht]
\centering
\includegraphics[width=0.9\linewidth]{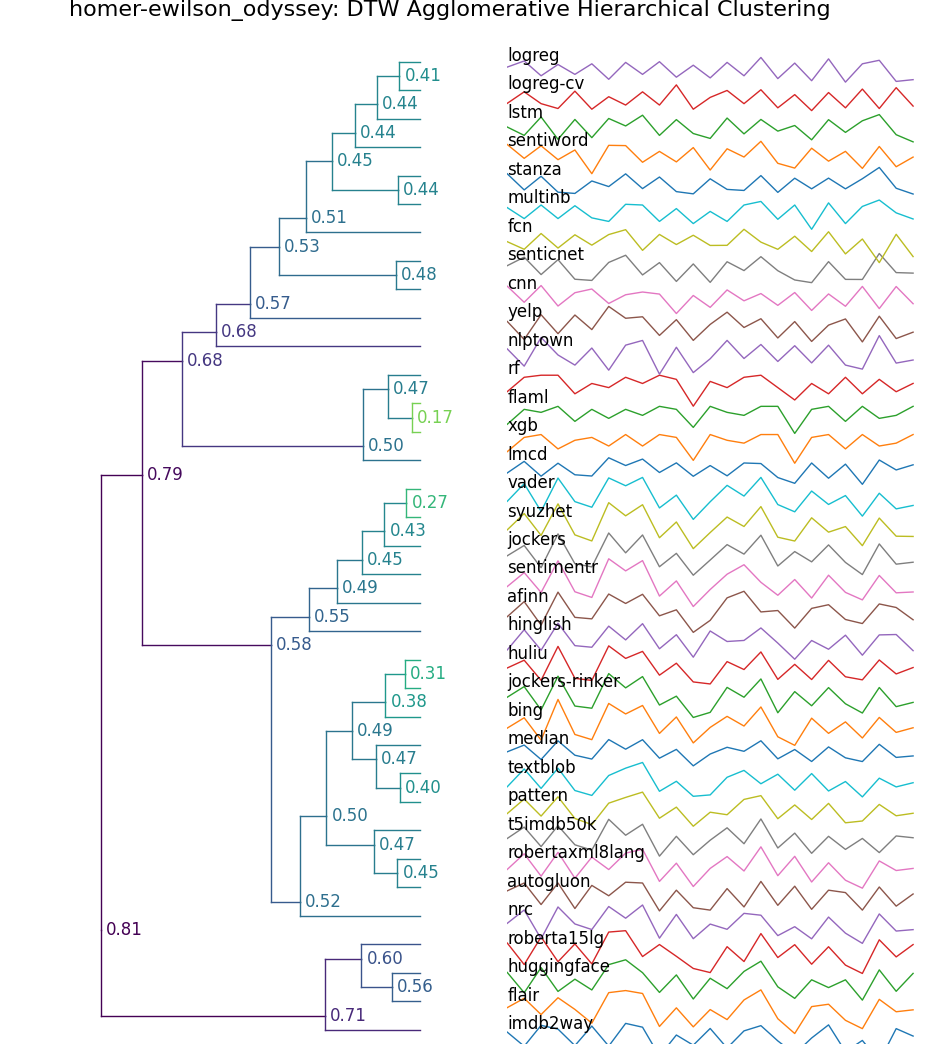}
\caption{ \textit{Odyssey} by Homer (trans. Emily Wilson) }
\label{appfig:metric_hcd_homerwilson_o}
\end{figure}

\begin{figure}[!ht]
\centering
\includegraphics[width=0.9\linewidth]{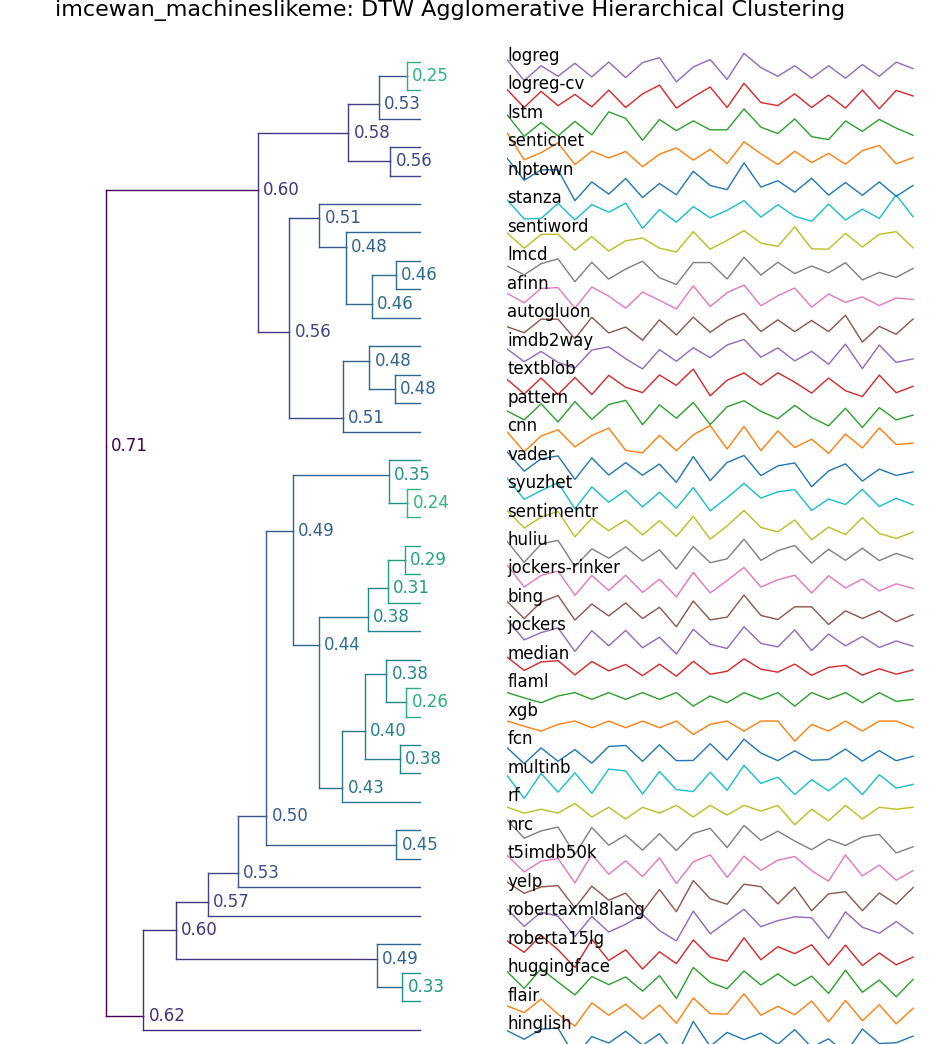}
\caption{ \textit{Machines Like Me} by Ian McEwan}
\label{appfig:metric_hcd_imcewan_mlm}
\end{figure}

\begin{figure}[!ht]
\centering
\includegraphics[width=0.9\linewidth]{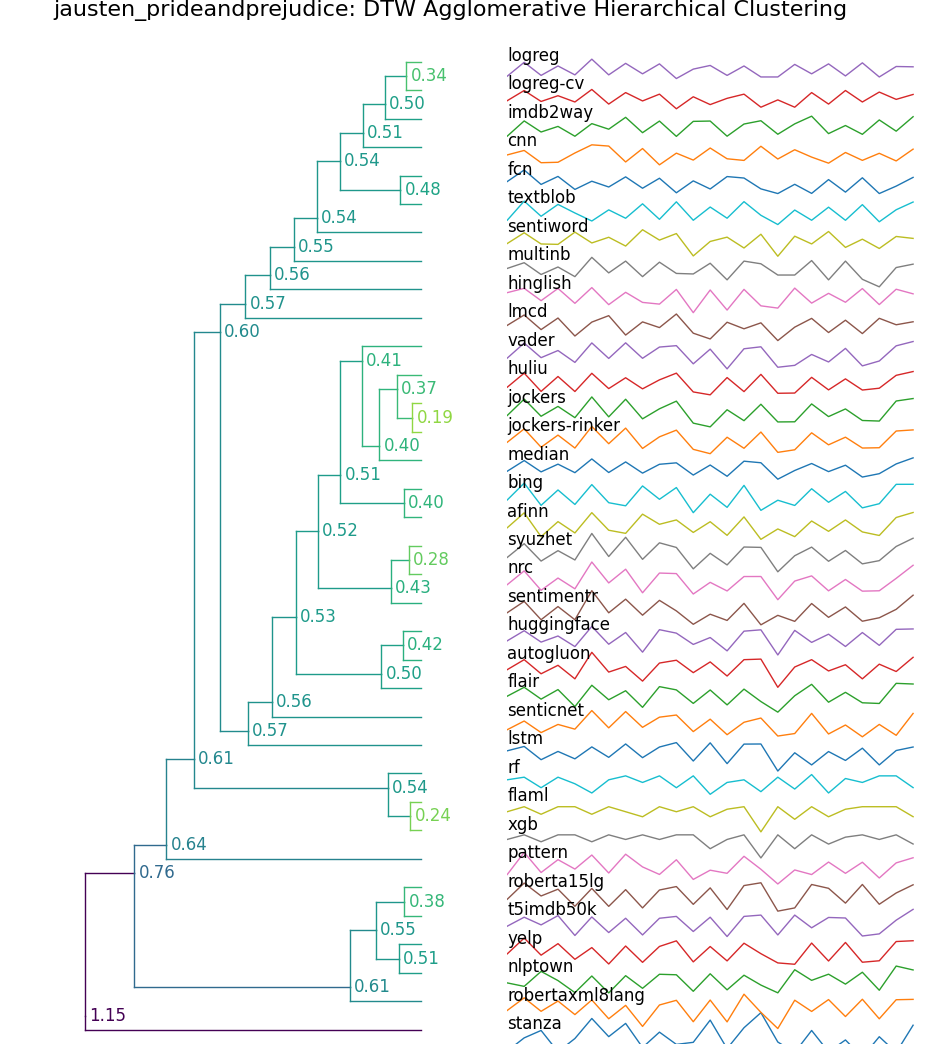}
\caption{Model-Corpus Compatibility on \textit{Pride and Prejudice} by Jane Austen }
\label{appfig:metric_hcd_jausten_pap}
\end{figure}

\begin{figure}[!ht]
\centering
\includegraphics[width=0.9\linewidth]{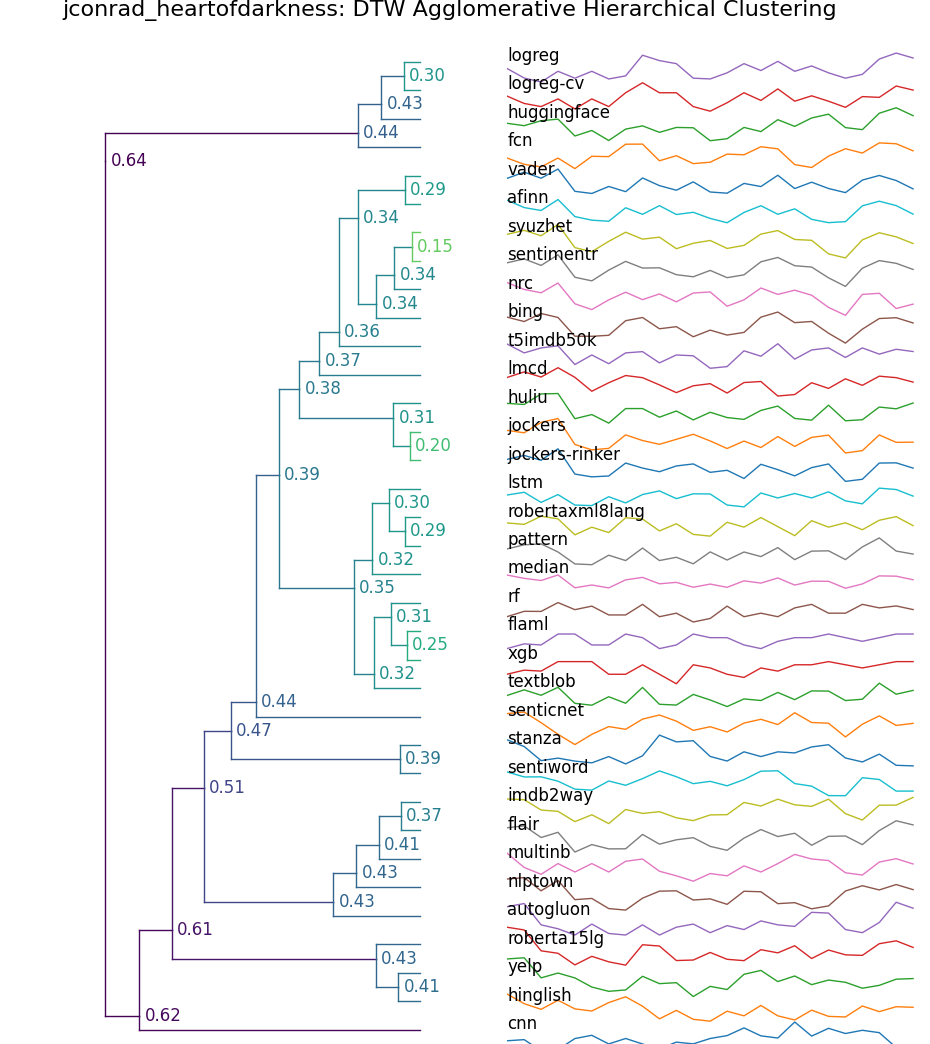}
\caption{ \textit{Heart of Darkness} by Joseph Conrad}
\label{appfig:metric_hcd_jconrad_hod}
\end{figure}

\begin{figure}[!ht]
\centering
\includegraphics[width=0.9\linewidth]{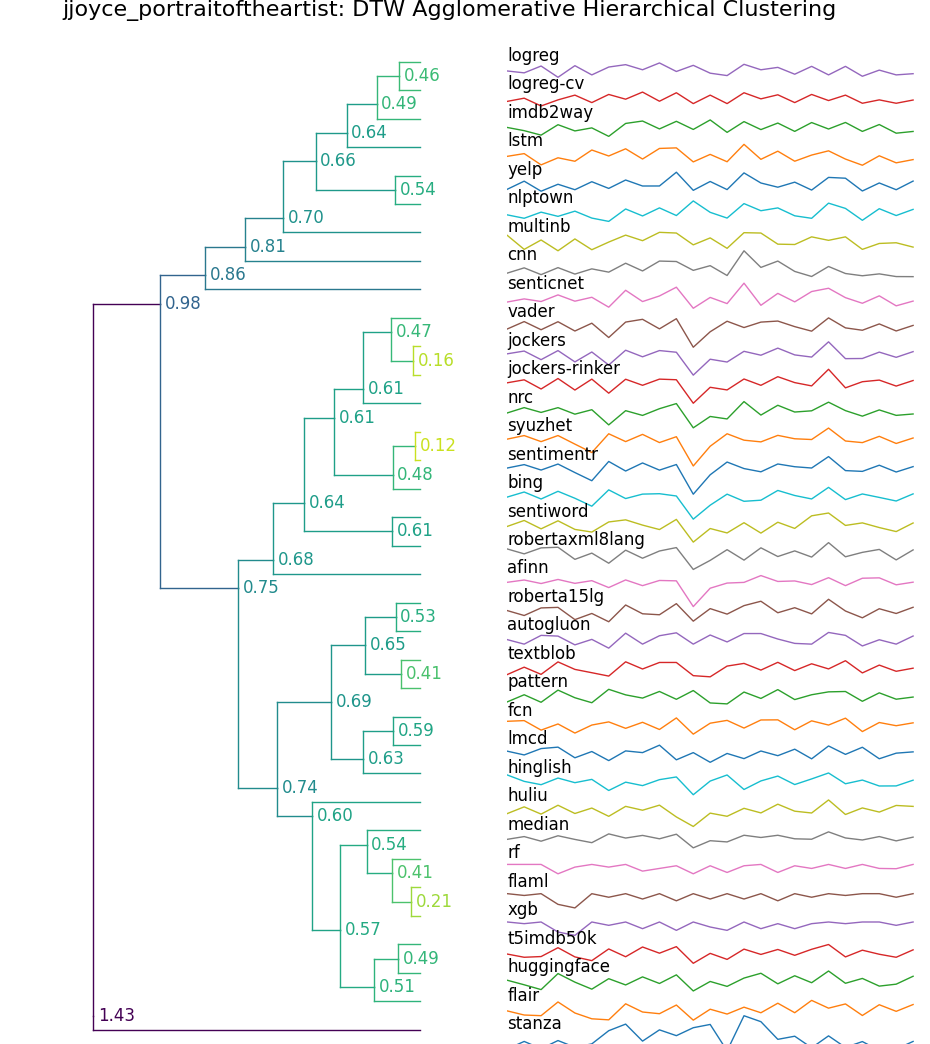}
\caption{ \textit{A Portrait of the Artists as a Young Man} by James Joyce }
\label{appfig:metric_hcd_jjoyce_apotaaaym}
\end{figure}

\begin{figure}[!ht]
\centering
\includegraphics[width=0.9\linewidth]{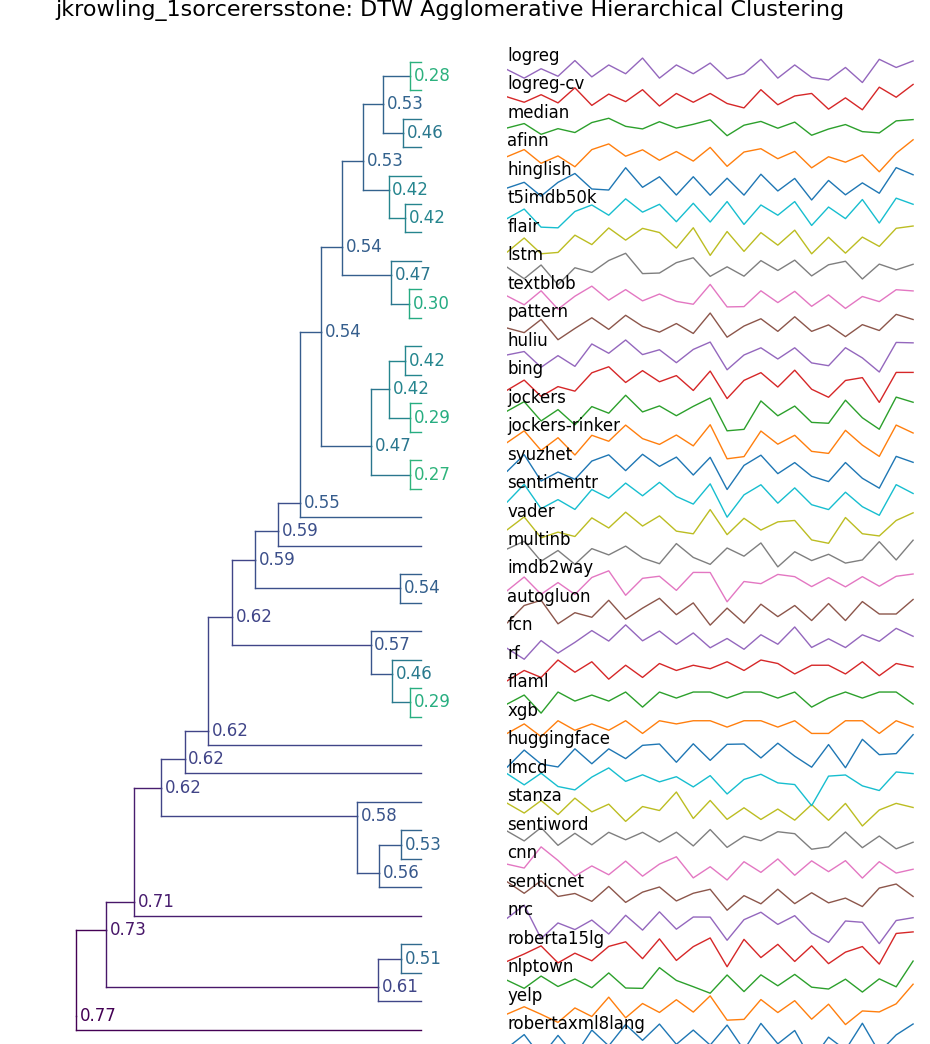}
\caption{ \textit{Harry Potter and the Sorcerer's Stone} by J. K. Rowling}
\label{appfig:metric_hcd_jkrowling_hpatss}
\end{figure}

\begin{figure}[!ht]
\centering
\includegraphics[width=0.9\linewidth]{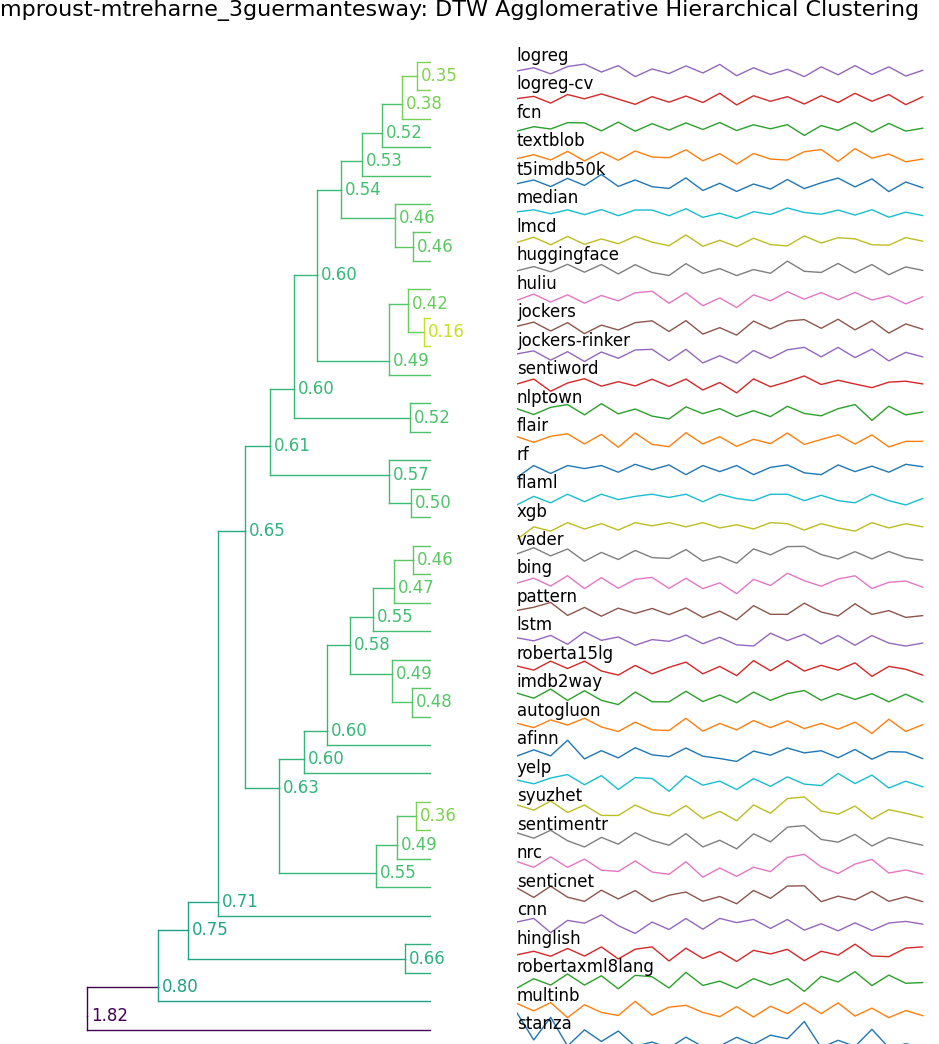}
\caption{ \textit{The Guermantes Way} by Marcel Proust (trans. Mark Treharne) }
\label{appfig:metric_hcd_mprousttreharne_tgw}
\end{figure}

\begin{figure}[!ht]
\centering
\includegraphics[width=0.9\linewidth]{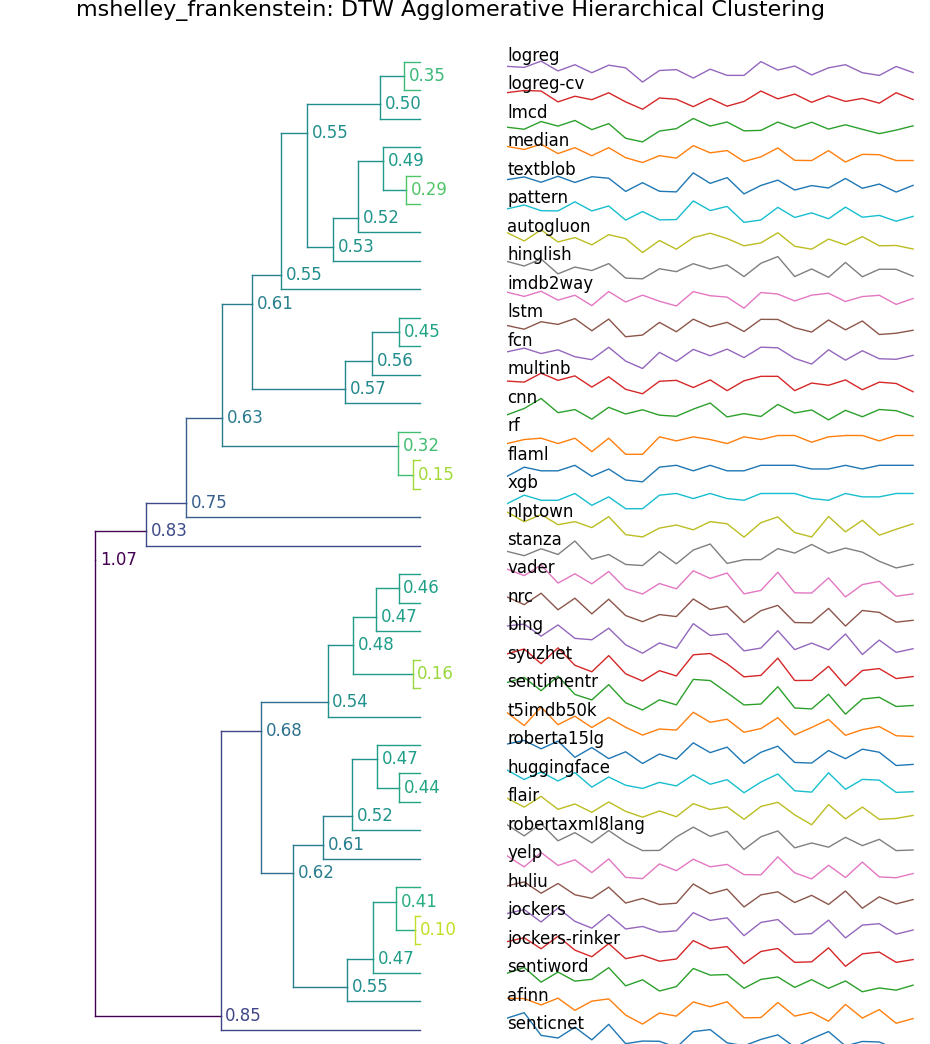}
\caption{ \textit{Frankenstein} by Mary Shelley}
\label{appfig:metric_hcd_mshelley_f}
\end{figure}

\begin{figure}[!ht]
\centering
\includegraphics[width=0.9\linewidth]{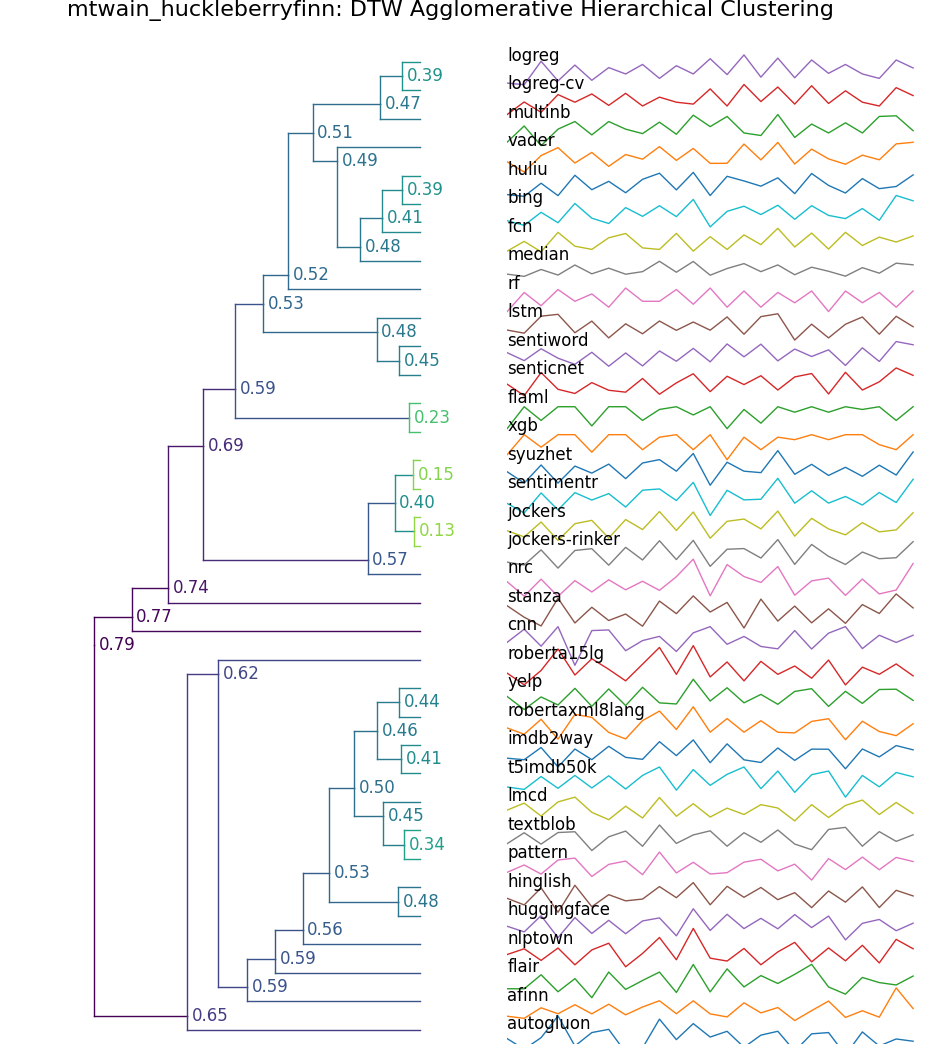}
\caption{ \textit{Huckleberry Finn} by Mark Twain}
\label{appfig:metric_hcd_mtwain_hf}
\end{figure}

\begin{figure}[!ht]
\centering
\includegraphics[width=0.9\linewidth]{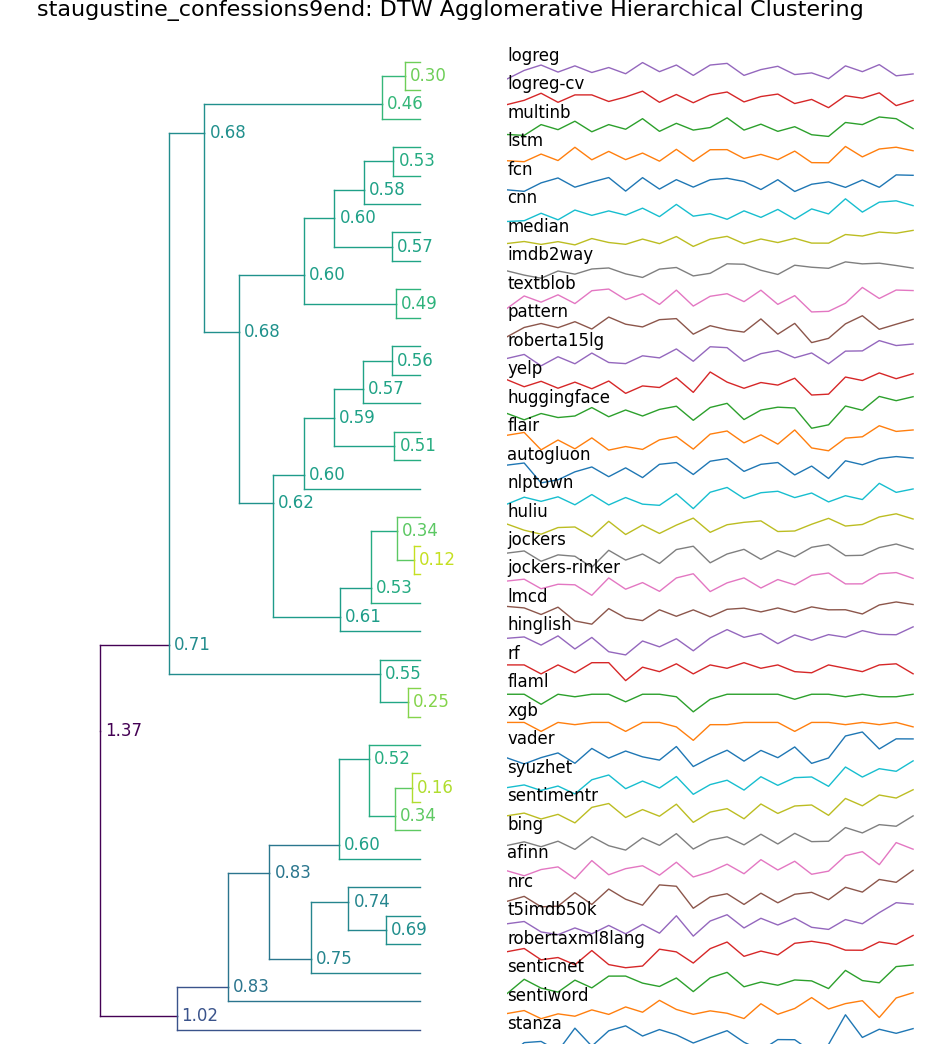}
\caption{ \textit{Confessions (thru Book  9)} by St. Augustine}
\label{appfig:metric_hcd_staugustine_c}
\end{figure}

\begin{figure}[!ht]
\centering
\includegraphics[width=0.9\linewidth]{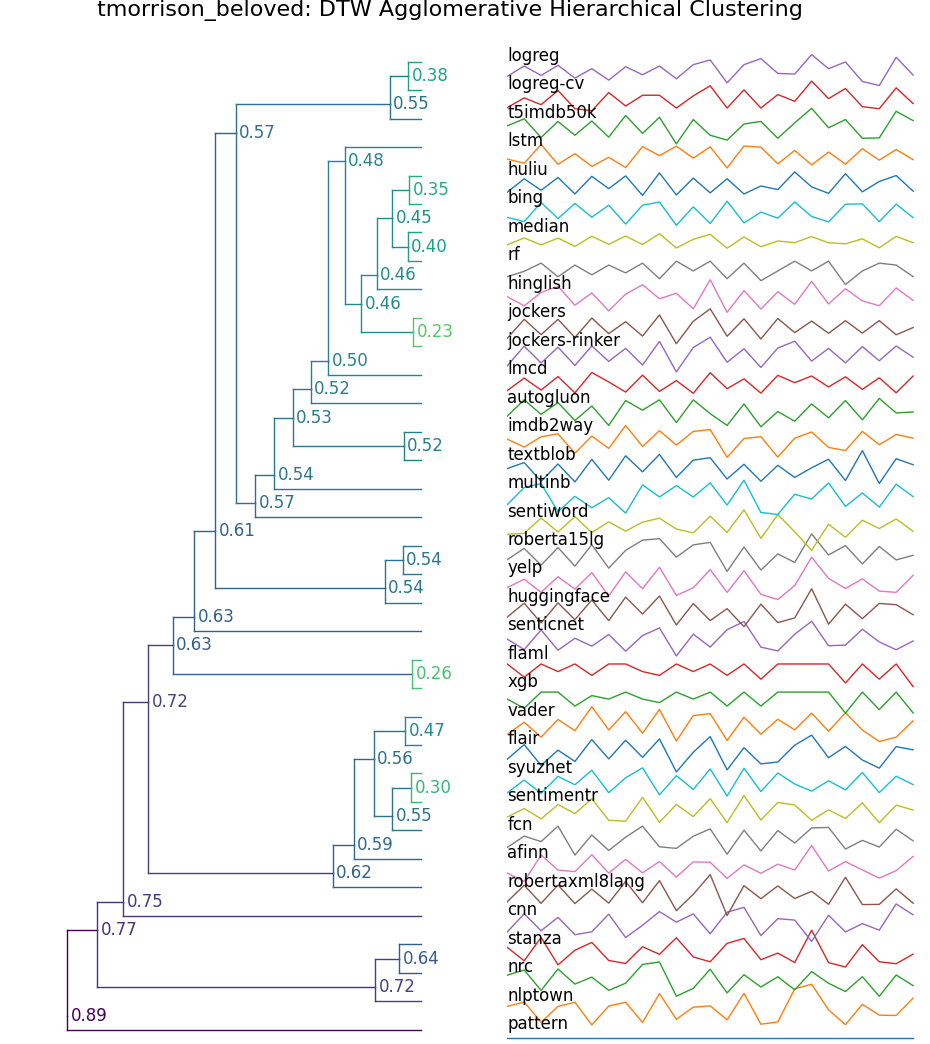}
\caption{ \textit{Beloved} by Toni Morrison}
\label{appfig:metric_hcd_tmorrison_b}
\end{figure}

\begin{figure}[!ht]
\centering
\includegraphics[width=0.9\linewidth]{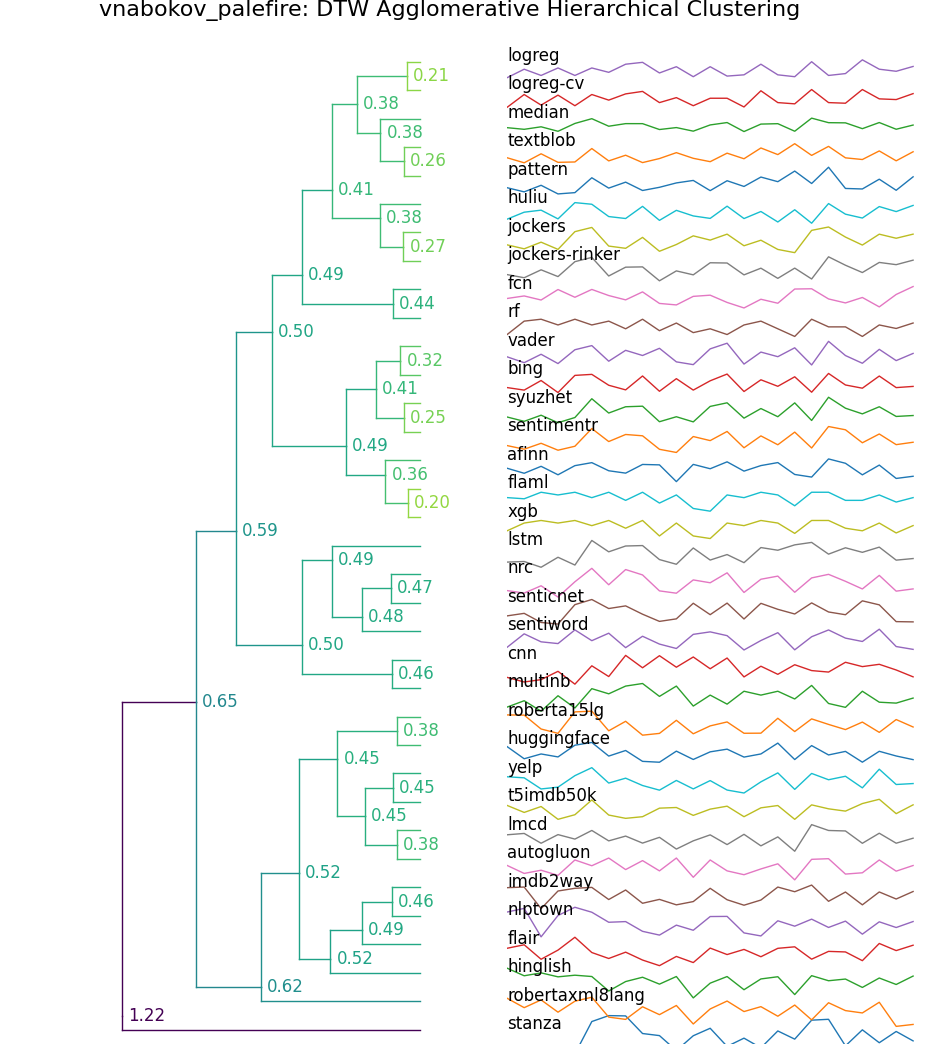}
\caption{ \textit{Pale Fire} by Vladimir Nabokov}
\label{appfig:metric_hcd_vnabokov_pf}
\end{figure}

\begin{figure}[!ht]
\centering
\includegraphics[width=0.9\linewidth]{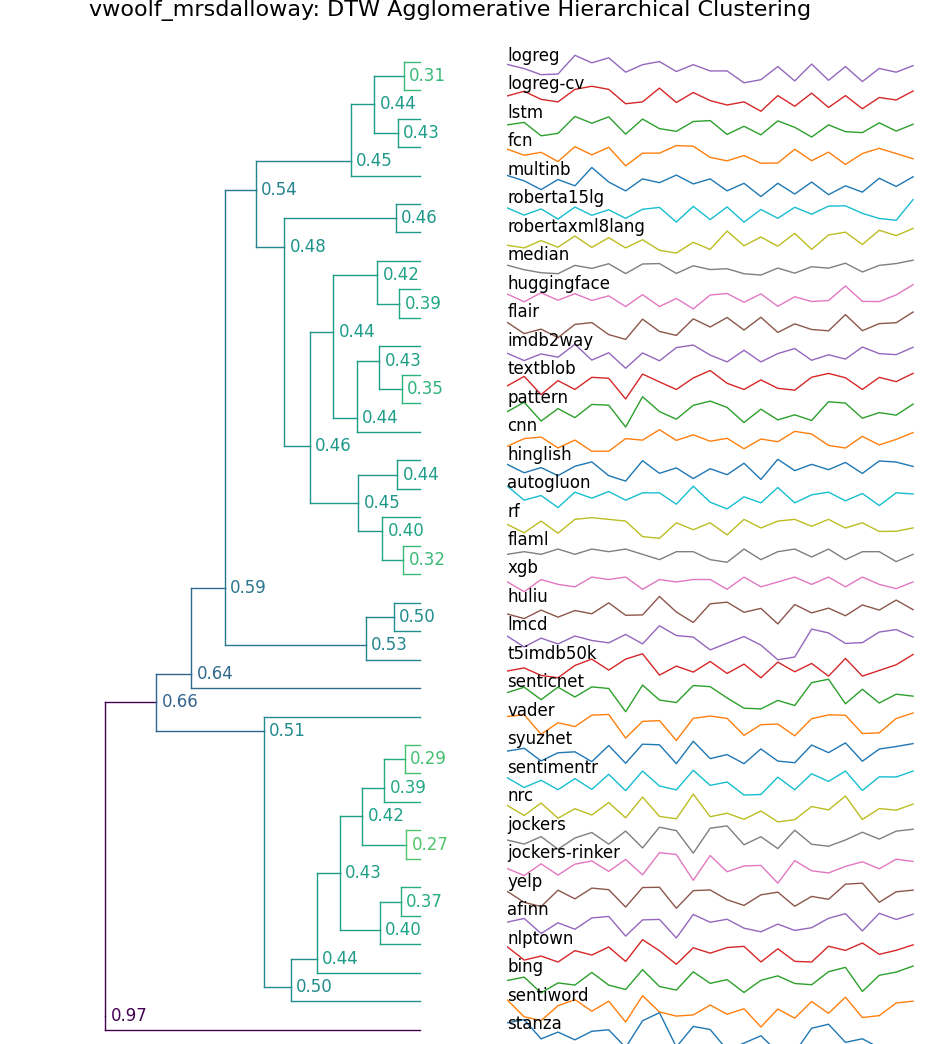}
\caption{ \textit{Mrs. Dalloway} by Virginia Woolf}
\label{appfig:metric_hcd_vwoolf_md}
\end{figure}

\begin{figure}[!ht]
\centering
\includegraphics[width=0.9\linewidth]{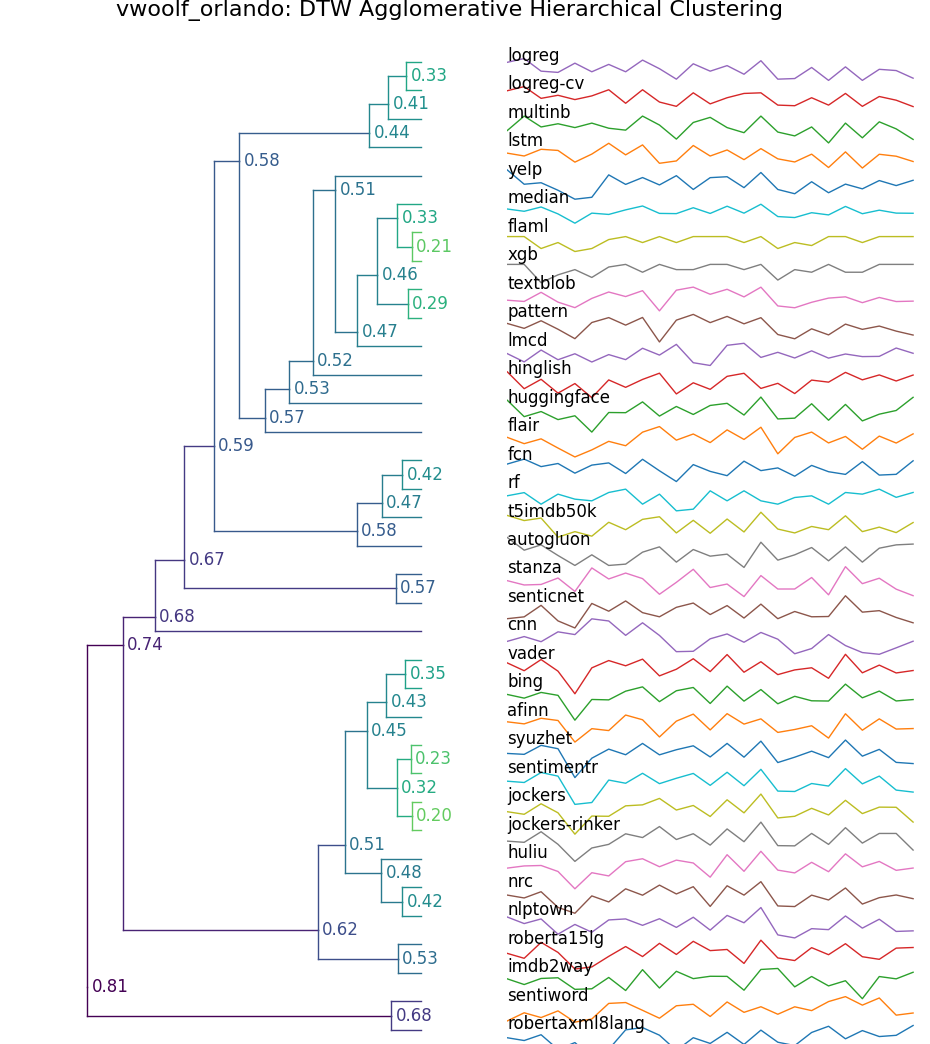}
\caption{ \textit{Orlando} by Virginia Woolf}
\label{appfig:metric_hcd_vwoolf_o}
\end{figure}

\begin{figure}[!ht]
\centering
\includegraphics[width=0.9\linewidth]{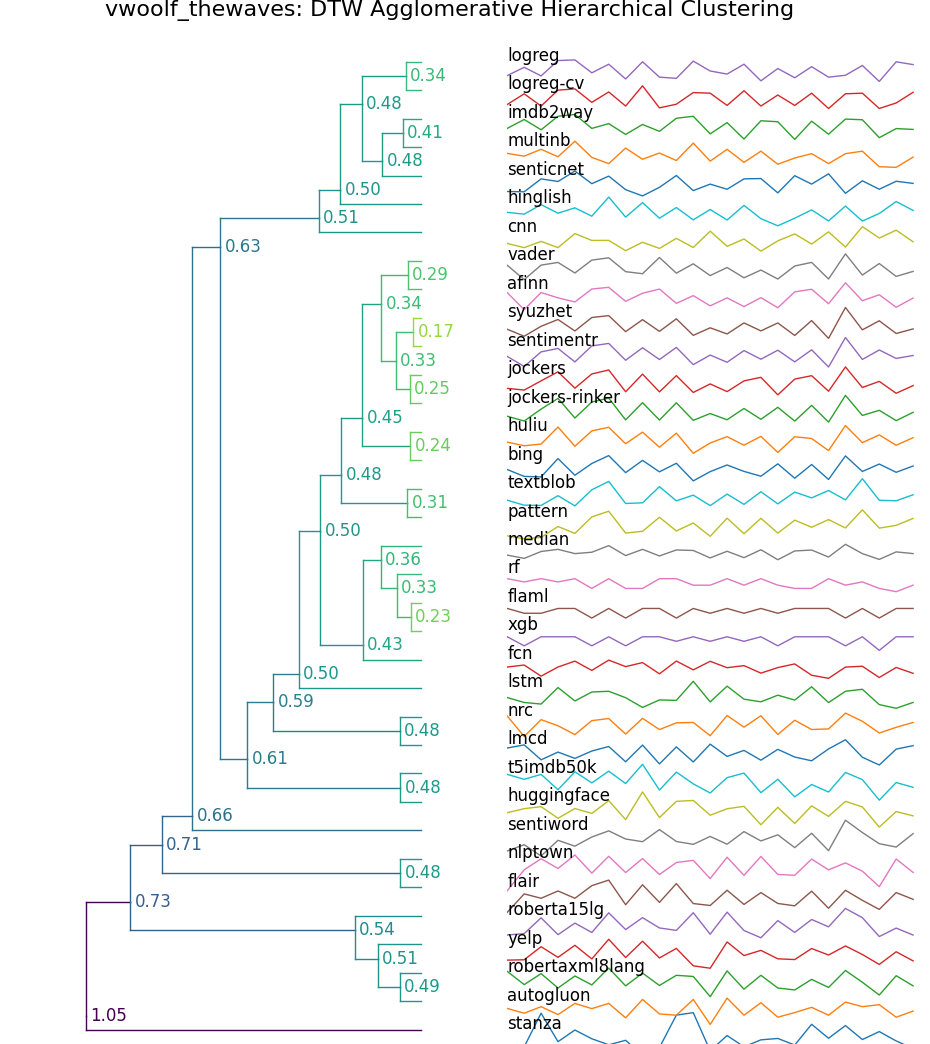}
\caption{ \textit{The Waves} by Virginia Woolf}
\label{appfig:metric_hcd_vwoolf_tw}
\end{figure}

\begin{figure}[!ht]
\centering
\includegraphics[width=0.9\linewidth]{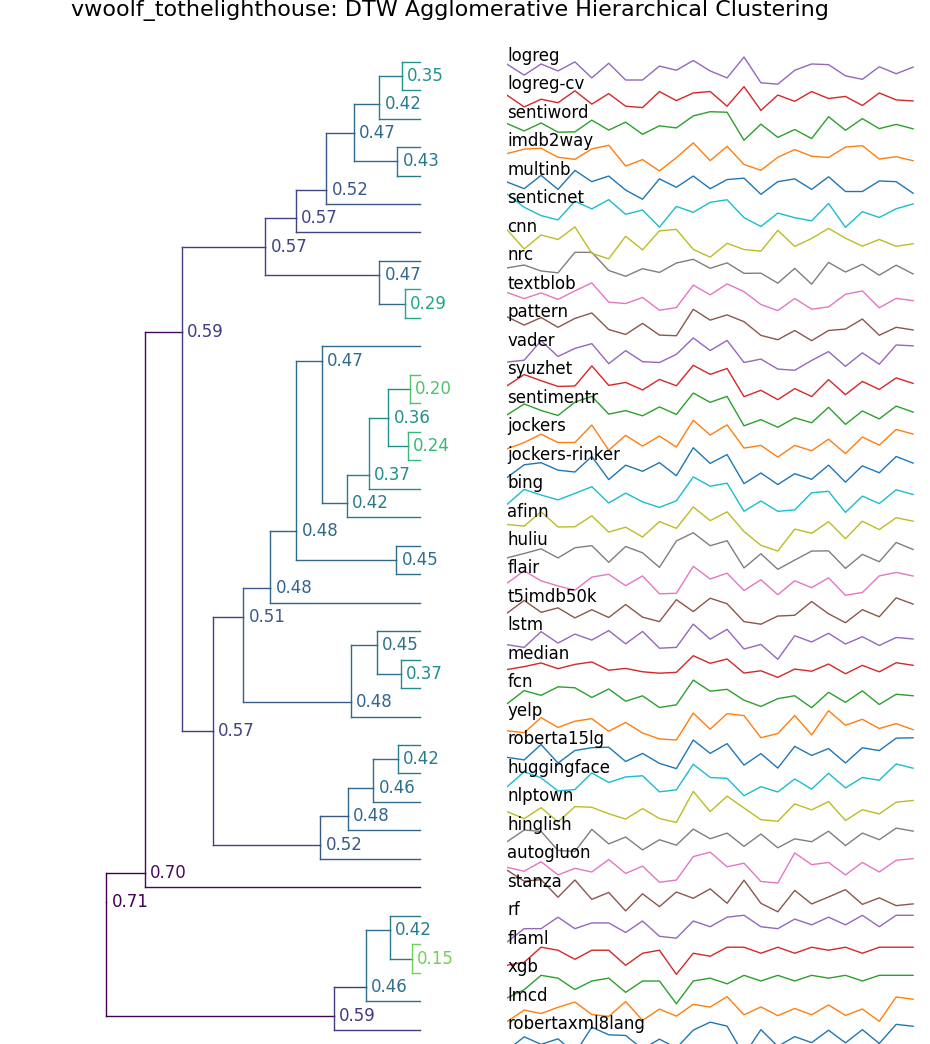}
\caption{ \textit{To The Lighthouse} by Virginia Woolf}
\label{appfig:metric_hcd_vwoolf_ttl}
\end{figure}

\end{document}